\documentclass[preprint,12pt]{elsarticle}

\usepackage{amsmath,amssymb,amsfonts}
\usepackage{algorithmic}
\usepackage{graphicx}
\usepackage{textcomp}
\usepackage[dvipsnames]{xcolor}
\usepackage{hyperref}
\usepackage{subcaption}
\usepackage{diagbox}
\usepackage{cleveref}
\usepackage{natbib}
\usepackage{cleveref}

\journal{
}

\begin{document}


\begin{frontmatter}



\title{
Deconstructing Recurrence, Attention, and Gating: Investigating the transferability of Transformers and Gated Recurrent Neural Networks in forecasting of dynamical systems
}


\author[inst1]{Hunter Heidenreich}
\author[inst2]{Pantelis R. Vlachas}
\author[inst1]{Petros Koumoutsakos}

\affiliation[inst1]{
			organization={School of Engineering and Applied Sciences, Harvard University},
            addressline={29 Oxford Street}, 
            city={Cambridge},
            postcode={MA 02138}, 
            state={Massachusetts},
            country={USA}}
\affiliation[inst2]{organization={Institute of Structural Engineering, ETH Zurich},
            postcode={CH 8093}, 
            state={Zurich},
            country={Switzerland}
            }

\begin{abstract}
Machine learning architectures, including transformers and recurrent neural networks (RNNs) have revolutionized forecasting in applications ranging from text processing to extreme weather.
Notably, advanced network architectures, tuned for applications such as natural language processing, are transferable to other tasks such as spatiotemporal forecasting tasks.
However, there is a scarcity of ablation studies to illustrate the key components that enable this forecasting accuracy.
The absence of such studies, although explainable due to the associated computational cost, intensifies the belief that these models ought to be considered as black boxes.
In this work, we decompose the key architectural components of the most powerful neural architectures, namely gating and recurrence in RNNs, and attention mechanisms in transformers.
Then, we synthesize and build novel hybrid architectures from the standard blocks, performing ablation studies to identify which mechanisms are effective for each task.
The importance of considering these components as hyper-parameters that can augment the standard architectures is exhibited on various forecasting datasets, from the spatiotemporal chaotic dynamics of the multiscale Lorenz 96 system, the Kuramoto-Sivashinsky equation, as well as standard real world time-series benchmarks.
A key finding is that neural gating and attention improves the performance of all standard RNNs in most tasks, while the addition of a notion of recurrence in transformers is detrimental.
Furthermore, our study reveals that a novel, sparsely used, architecture which integrates Recurrent Highway Networks with neural gating and attention mechanisms, emerges as the best performing architecture in high-dimensional spatiotemporal forecasting of dynamical systems.
\end{abstract}




\end{frontmatter}


\section{Introduction}

Accurate prediction of complex dynamical systems is crucial for scientific investigations and decision-making processes, spanning a multitude of fields such as atmosphere and climate science, economics, public health, and energy generation.
Accurate forecasts allow for the detection of future trends, which informs better resource allocation, economic planning, and decision-making, and has social benefits by reducing the impact of natural disasters, improving public health outcomes, and enhancing the overall quality of life. 

Machine learning (ML) has become integral in the analysis and prediction of complex dynamical systems and time-series data \cite{gamboa2017deep,lim2021time}.
Much of the recent success of ML is due to deep learning (DL) models, i.e., neural networks with many layers that learn data patterns of increasing complexity.
ML is especially appealing in cases where the use of other forecasting techniques that assume knowledge of the equations governing underlying dynamics is prohibitive (first principles, physics-based approach), either because the equations are not available, or because their evaluation is computationally expensive or intractable.
Instead, Data-driven ML algorithms can identify patterns, trends, and other important features necessary for accurately forecasting complex dynamical systems, without the need for physics-based modeling (although incorporating priors on physics can be beneficial and can assist the ML model) \cite{wan2018data,beck2019deep,pan2018data,san2018neural}.
ML-based forecasting has already found numerous applications in a range of fields, including weather forecasting~\cite{prudden2020review,rasp2020weatherbench,espeholt2022deep}, financial forecasting~\cite{werge2022adavol,xing2018natural,kumbure2022machine}, energy production~\cite{magazzino2021machine,hijazi2016machine}, and health monitoring~\cite{sujith2022systematic,hijazi2016machine,garcia2018mental}. 

Recurrent Neural Networks (RNNs)~\cite{sutskever2013training} are a special type of neural network that are particularly well-suited for analyzing and predicting dynamical systems~\cite{vlachas2020backpropagation}.
Unlike feedforward neural networks (FNN), which take a fixed input and produce a fixed output, RNNs are designed to work with sequential data.
In dynamical systems, where the state of the system evolves over time in a complex and often nonlinear way, RNNs can take into account the history of the state when making predictions about its future behavior.
In this way, RNNs can capture complex temporal interdependencies between the different states, allowing them to model and predict more accurately the system's dynamics.

Early applications of FNNs and RNNs~\cite{vermaak1998recurrent} to time series data and complex dynamical systems have been hindered due to scalability issues, limited computational resources, as well as numerical problems in training, i.e., the vanishing and exploding gradients (VEGs) problem~\cite{hochreiter1998vanishing,hanin2018neural,sutskever2013training}.
The use of neural networks for modeling and predicting dynamical systems dates back to Lapedes and Farber's work in 1987~\cite{lapedes1987nonlinear}, where they demonstrated the effectiveness of FNNs in modeling deterministic chaos.
The potential of RNNs for capturing temporal dynamics in physical systems was explored first using low dimensional RNNs in 1997~\cite{elman1990finding}.

Recent advances in computational hardware and software~\cite{paszke2019pytorch}, as well as the introduction of novel RNN architectures that employ gating mechanisms to cope with VEGs~\cite{hochreiter1997long,graves2012long,chung2014empirical,zilly2017recurrent,tallec2018can}, paved the way for accurate forecasting of high-dimensional complex dynamical systems~\cite{vlachas2018data,vlachas2020backpropagation}.
Long Short-Term Memory networks (LSTMs)~\cite{hochreiter1997long,graves2012long}, Gated Recurrent Units (GRUs)~\cite{chung2014empirical}, and the more recent Recurrent Highway Networks (RHNs)~\cite{zilly2017recurrent} extend RNNs by introducing gating mechanisms that allow a network to learn to control the flow of information through the network over time, thereby alleviating the VEG problem.

In~\cite{bianchi2017recurrent}, the authors have benchmarked RNN architectures in short-term load forecasting for demand and resource consumption in supply networks. 
In~\cite{laptev2017time}, on the other hand, the authors have utilized RNNs for extreme event detection in low-dimensional time series.
In~\cite{vlachas2018data}, LSTMs have been employed to forecast the state evolution of high-dimensional dynamical systems that exhibit spatiotemporal chaos demonstrating significantly better scalability and accuracy compared to Gaussian Processes.
In~\cite{vlachas2020backpropagation}, LSTMs and GRUs are benchmarked along with Reservoir Computers (RCs) in forecasting the high-dimensional dynamics of the multiscale Lorenz 96 system, and the Kuramoto-Sivashinsky equation.
The trained networks are then used to accurately reproduce important properties of the underlying dynamical systems, e.g., the Lyapunov spectrum.
In~\cite{vlachas2022multiscale}, gated RNNs are employed to forecast multiscale dynamical systems, i.e., the Navier Stokes flow past a cylinder at high Reynolds numbers.

RHNs have been shown to be effective for a range of natural language processing (NLP) tasks~\cite{pundak2017highway,suarez2017language,srivastava2015training}, but their applications to dynamical systems forecasting remain under-investigated.
Furthermore, recent broad-spectrum investigations \cite{gilpin2chaos,gilpin2023large} into neural forecasting of chaotic attractors have clearly demonstrated the efficacy of neural forecasting over traditional methods, but have left open questions as to the right inductive biases and mechanisms for forecasting.

Despite the fact that gated architectures have played a critical role for RNNs and their achievements in NLP and dynamical system forecasting \cite{tallec2018can}, training recurrent networks remains inherently slow due to the sequential nature of their data processing.
To address this issue, Transformer models~\cite{vaswani2017attention} were introduced to process input sequences in parallel during training rather than sequentially like RNNs. 
This results in more efficient processing of long sequences, such as entire documents or transcriptions of conversations, and has revolutionized the field of NLP.
Transfomers have achieved state-of-the-art performance on a wide range of language processing tasks, such as translation, question answering, and text summarization.
The recently published ChatGPT language model developed by OpenAI generates impressive human-like responses to a wide range of questions and prompts.
The backbone of ChatGPT is the Generative Pre-trained Transformer~\cite{radford2018improving,brown2020language}, one of the largest and most powerful neural models to date.
At the same time, the Vision Transformer (ViT)~\cite{dosovitskiy2020image} has achieved state-of-the-art performance in computer vision tasks, while DALL-E~\cite{ramesh2021zero,ramesh2022hierarchical}, a text-to-image generation network that employs the GPT, has revolutionized text-to-image models and multi-modal models more broadly.

A central aspect of the Transformer model is the attention mechanism~\cite{niu2021review}.
An attention mechanism allows a model to focus on different parts of an input sequence, conditionally dependent on the contents of that sequence, allowing for the mechanism to selectively rely on the most relevant information for a given timestep prediction.
Despite the widespread success of transformers in NLP, their application to dynamical systems is in its infancy.
In~\cite{geneva2022transformers,pannatier2022accurate} Transformers are employed for forecasting fluid flows and other physical systems.
Transformers are also used in~\cite{gao2022earthformer} for precipitation nowcasting and El Niño/Southern Oscillation (ENSO) forecasting.
In~\cite{zhou2022fedformer} Transformers are coupled with a seasonal-trend decomposition method for long-term time series forecasting, while in~\cite{xu2020spatial} they are used for traffic flow prediction.
In~\cite{wu2020deep} Transformers are employed to forecast influenza illness time-series data, while in~\cite{chattopadhyay2020deep} they are used to forecast geophysical turbulence.
Transformers are also one of many models considered in a broad benchmark of chaotic attractors \cite{gilpin2chaos,gilpin2023large}.


Although Transformers and gated RNNs have been shown to be effective in forecasting dynamical systems, they are often applied blindly without further examination of their internal components.
In simpler terms, neural architectures that perform well in one application, such as neural translation, are often directly applied to other tasks, like dynamical systems forecasting, without modification. 
This study highlights the importance of a more thoughtful application of sequential models to novel domains by breaking down neural architectures into their core components, forming novel architectures by augmenting these components and conducting ablation studies to determine which mechanisms are effective for each task.
Without this analysis, the lack of transparency surrounding DL architectures is reinforced, perpetuating the idea of these models as black boxes.
Furthermore, a practitioner leaves easy-to-attain performance gains on the table by not doing so.

An influential study in \cite{dey2017gate} explores the effectiveness of three variants of the gating mechanism in GRUs, with a focus on simplifying the structure and reducing the parameters in the gates. 
The authors demonstrate that even gating mechanisms with fewer parameters can be effective for specific applications. 
Similarly, \cite{greff2016lstm} performs an extensive architectural search over LSTMs, analyzing eight variants on three representative tasks: speech recognition, handwriting recognition, and polyphonic music modeling, although no variant significantly outperforms the standard LSTM cell.
Another interesting consideration is that of \cite{tallec2018can} which illustrates that gating mechanisms equip an RNN with a general quasi-invariance to time transformations. 
In fact, it is suggested that this type of gating is a large part of what makes gated RNNs so potent at temporal forecasting.
In the context of Transformers, gating has been attempted in a few situations that increase the performance (and complexity) of a standard Transformer: 
gated recurrence or feedback \cite{didolkar2022temporal,hutchins2022block}, self-gating \cite{shazeer2020glu,chai2020highway,lim2021temporal}, or multi-Transformer gating \cite{liu2021gated}. 
Additionally, as discussed further below, attention can similarly be viewed as a form of conditional gating \cite{baldi2023quarks}.
Despite the promising results of LSTMs and Transformers in forecasting time-series data and complex high-dimensional systems, such focused, mechanistic research is currently absent for dynamical systems.

In this work we
\begin{enumerate}
    \item identify and decompose the core mechanisms of Transformers and gated RNNs, namely gating, attention, and recurrence,
    \item propose novel enhancements to the core mechanisms,
    \item and, create hybrid novel architectures by combining the core mechanisms, and benchmark their performance in standard prototypical complex dynamical systems and open source time-series datasets.
\end{enumerate}
We differentiate between stateless (non-recurrent) Transformer models and stateful RNNs.
Regarding RNNs, we study three architectures, LSTMs, GRUs, and RHNs.
For Transformers, we study two variants that depend on the position of the normalization layer, specifically pre-layer normalization, and post-layer normalization.
In both transformers and RNNs, we test four different gating mechanisms i.e., additive, learned rate, input-dependent, and coupled input-dependent, augmenting the functional form of the gating used.
For attention, we consider multi-headed scaled dot production attention without positional bias as well as data-dependent and data-independent relative positional bias for both Transformers and RNNs.
Finally, we attempt to investigate augmenting Transformers with an additional recurrent memory matrix to understand the forecasting implications on chaotic systems.
A more elaborate explanation of these mechanisms is provided in~\Cref{sec:methods}.

We demonstrate that neural gating, attention mechanisms, and recurrence can be treated as architectural hyper-parameters, designing novel hybrids like gated Transformers with recurrence, or deep-in-time RNNs with gated attention.
We find that the default architectures of Transformers or gated RNNs were the least effective in the forecasting benchmark tasks considered in this work.
On the contrary, the highest forecasting accuracy is achieved when tuning these architectural core mechanisms.
Our findings emphasize that the architectures should not be treated as fixed oracles and should not be transferred as-is in different tasks, as this leads to suboptimal performance.
In order to exploit the full potential of DL architectures, the core components must be tuned for optimal performance.

In~\Cref{sec:methods}, we introduce the gated RNN and Transformer architectures, the basic neural mechanisms they are based on, as well as the general framework for training and forecasting.
In~\Cref{sec:metrics}, we introduce the comparison metrics used to measure the forecasting performance of the methods.
In~\Cref{sec:lorenz}, we study the contribution of the different mechanisms for each model in forecasting the dynamics of the multiscale Lorenz 96 dynamical system.
In~\Cref{sec:KS}, we study the effects of different mechanisms on each model in forecasting the spatiotemporal dynamics of the Kuramoto-Sivashinsky equation.
In~\Cref{sec:realworld}, we benchmark the methods on open-source, real-world, time-series data, and~\Cref{sec:conclusion} concludes the paper summarizing the key findings.

\section{Methods}  
\label{sec:methods}

This work considers machine learning algorithms for temporal forecasting of an observable vector $\mathbf{o} \in \mathbb{R}^{d_o}$ sampled at a fixed rate $(\Delta t)^{-1}$, $\{\mathbf{o}_1, ..., \mathbf{o}_T\}$, where $\Delta t$ is omitted for concision. 
We compare two classes of models for capturing temporal dependencies: stateless Transformers \cite{vaswani2017attention} and stateful Recurrent Neural Networks (RNNs) \cite{elman1990finding}.


Given the current observable $\mathbf{o}_t$, the output of each model is a prediction $\mathbf{\hat{o}}_{t+1}$ of the observation at the next time instant $\mathbf{o}_{t+1}$. 
Let a model's history $\mathbf{H}_t \in \mathbb{R}^{n \times d}$ be the additional temporal information available beyond the current observable $\mathbf{o}_t$, where $\mathbf{H}_t = \mathbf{h}_{t-1} \in \mathbb{R}^{1 \times d_h}$ is for stateful models and $\mathbf{H}_t = \mathbf{o}_{t-S:t-1} \in \mathbb{R}^{\left(S-1\right) \times d_o}$ (the $S - 1$ last observations) is for stateless models. 
We capture both classes of temporal dependency in a general functional form, 
\begin{equation}
    \mathbf{\hat{o}}_{t+1}, \mathbf{H}_{t+1} = f_{\theta}(\mathbf{o}_t, \mathbf{H}_t),
\end{equation}
where $f_\theta$ takes the current observable and model history and produces an updated history and prediction for the next time step based on a learned set of parameters $\theta$. 
While all models in this work share this high-level description, they differ in their specific realizations of $f_\theta$ and in how their weights are learned from the data to forecast the dynamics of the observable. 

\begin{figure*}[htb!]
    \centering
    \begin{subfigure}{0.5\textwidth}
        \centering
        \includegraphics[width=.9\linewidth]{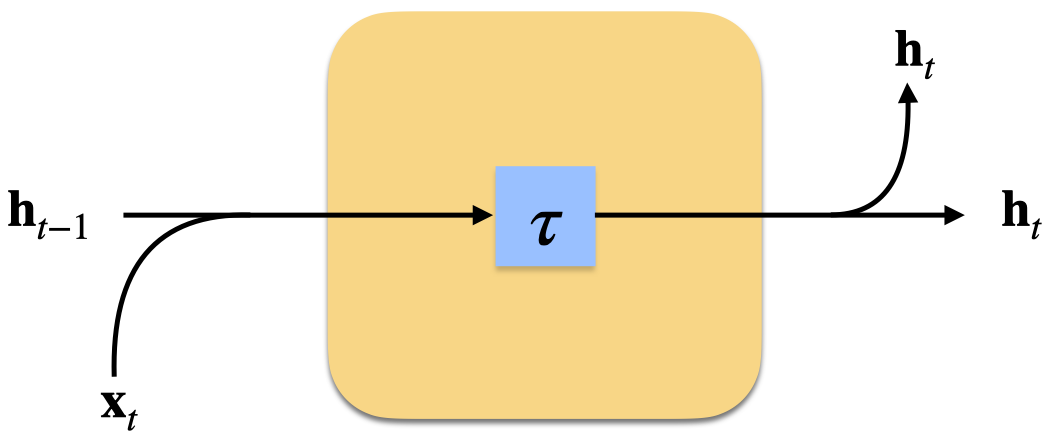}
        \caption{Elman Cell}
        \label{fig:rnn_elman_cell}
    \end{subfigure}%
    \begin{subfigure}{0.5\textwidth}
        \centering
        \includegraphics[width=.9\linewidth]{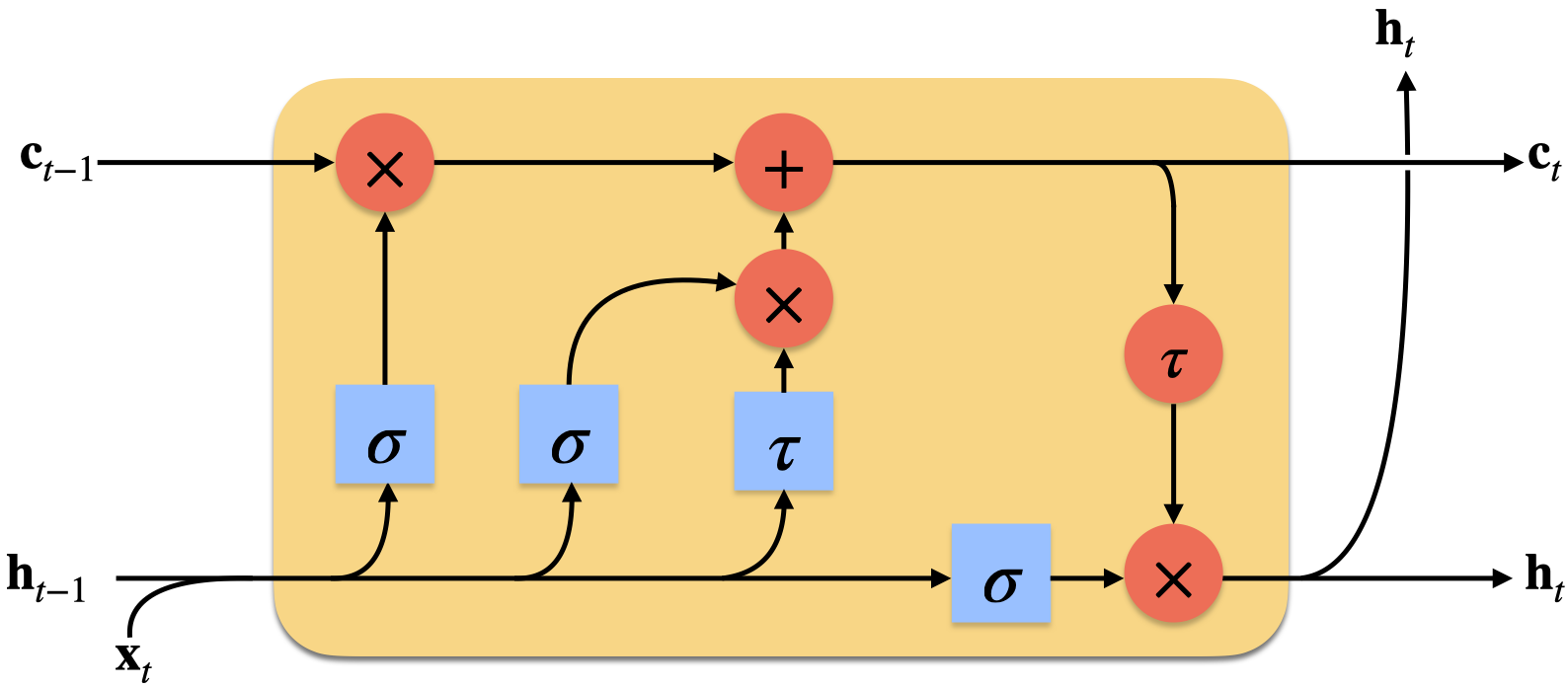}
        \caption{LSTM Cell}
        \label{fig:rnn_lstm_cell}
    \end{subfigure}
    
    \begin{subfigure}{0.5\textwidth}
        \centering
        \includegraphics[width=.9\linewidth]{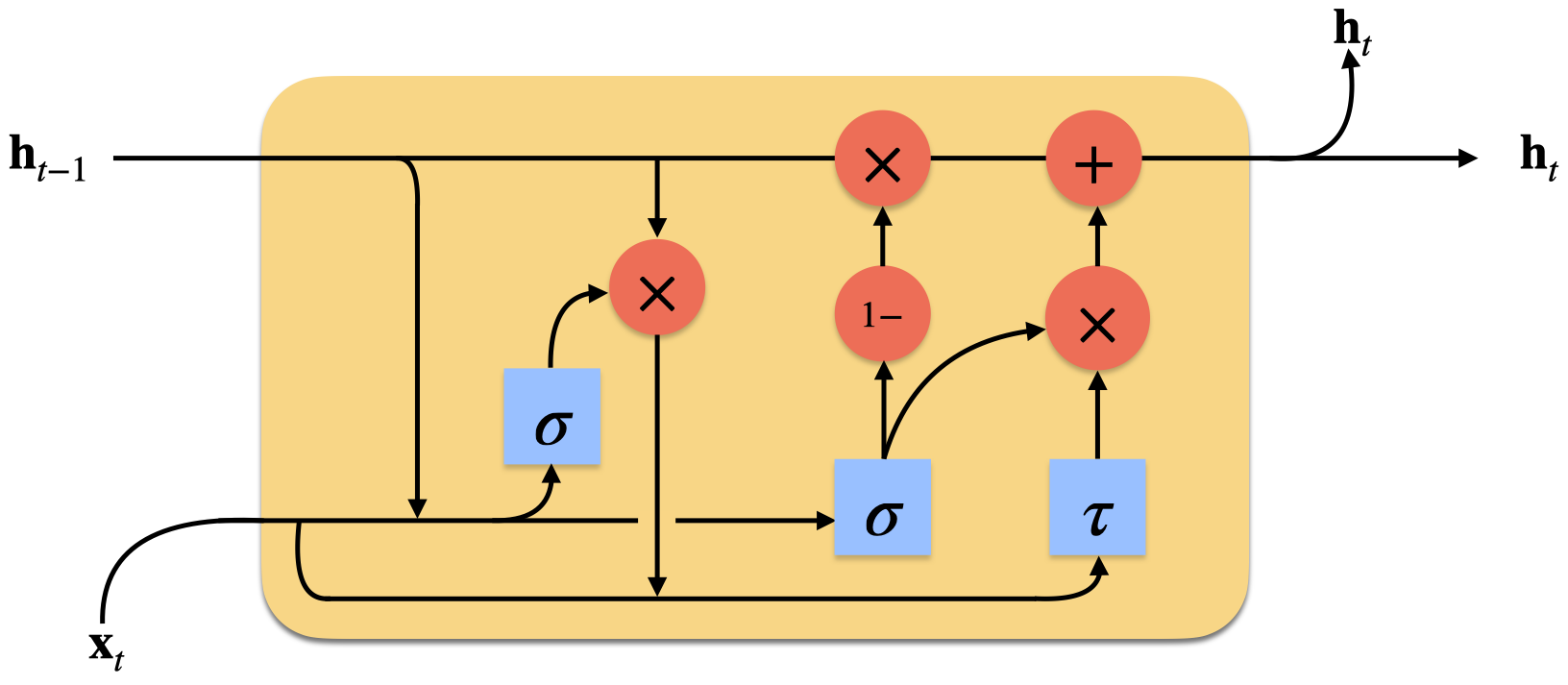}
        \caption{GRU Cell}
        \label{fig:rnn_gru_cell}
    \end{subfigure}%
    \begin{subfigure}{0.5\textwidth}
        \centering
        \includegraphics[width=.9\linewidth]{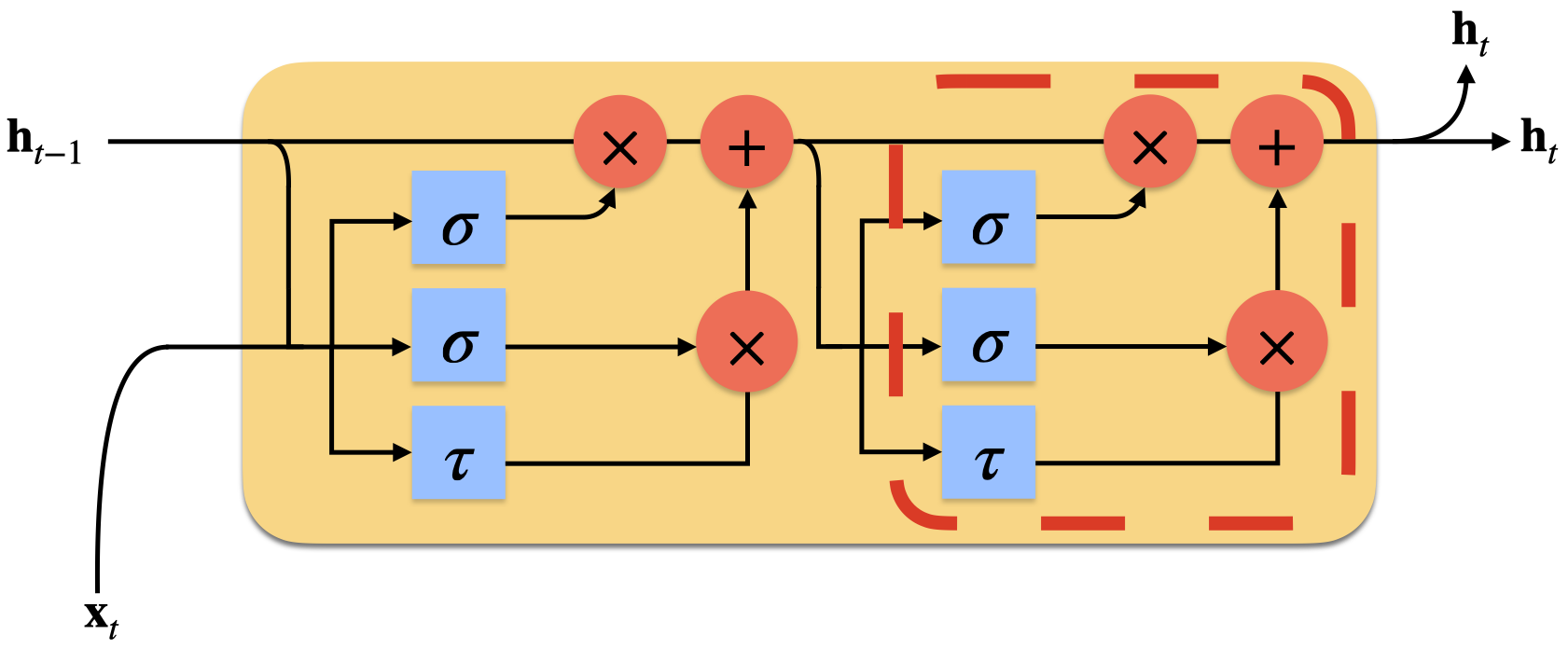}
        \caption{RHN Cell ($L = 2$)}
        \label{fig:rnn_rhn_cell}
    \end{subfigure}%
    \caption{
        Information flows through recurrent cells.
        Layers are denoted with squares and element-wise operations with circles.
        $\sigma$ is the sigmoid function and $\tau$ is the hyperbolic tangent.
        For the RHN, we depict a $L=2$ cell
        where the unit highlighted in red is repeated twice.
        Concatenation is denoted by the intersection of two directed lines, and copying is denoted by their forking.
    }
    \label{fig:rnn_cell_types}
\end{figure*}

\subsection{Recurrent Neural Networks (RNNs)}

We consider three classes of RNNs in this work: the Long Short-Term Memory (LSTM), the Gated Recurrent Unit (GRU), and the Recurrent Highway Network (RHN). These are depicted, alongside the Vanilla RNN for comparison, in~\Cref{fig:rnn_cell_types}.

\subsubsection{Long Short-Term Memory (LSTM)}
The LSTM \cite{hochreiter1997long,hochreiter1998vanishing} was the first architecture to alleviate the issue of VEGs in RNNs through the introduction of neural gating mechanisms. Let $\mathbf{z}_t = \left(\mathbf{h}_{t-1}, \mathbf{o}_t \right)$. The LSTM evolves $\mathbf{h}_t$ in time through
\begin{align}
    \mathbf{h}_t & = \mathbf{g}_t^o \odot \mathrm{tanh}(\mathbf{c}_t), & \\ 
    \label{eq:lstm_gate}
    \mathbf{c}_t & = \mathbf{g}_t^f \odot \mathbf{c}_{t-1} + \mathbf{g}_t^i \odot \mathbf{\Tilde{c}}_t, & \\
    \label{eq:lstm_gate_vec}
    \mathbf{g}_t^{k} & = \sigma \left(\mathbf{W}_k\mathbf{z}_t + \mathbf{b}_k \right),& k \in \{f, i, o\} \\
    \mathbf{\Tilde{c}}_t & = \mathrm{tanh}\left( \mathbf{W}_c\mathbf{z}_t + \mathbf{b}_c \right), &
\end{align}
where $\mathbf{g}_t^f,\mathbf{g}_t^i,\mathbf{g}_t^o \in \mathbb{R}^{d_h}$ are gating vectors (forget, input, and output gates, respectively), and $\mathbf{c}_t \in \mathbb{R}^{d_h}$ is an internal cell state that is propagated forward in time. The learnable parameters consist of the weight matrices $\mathbf{W}_k \in \mathbb{R}^{d_h \times (d_h + d_o)}$ and bias vectors $\mathbf{b}_k \in \mathbb{R}^{d_h}$ for $k \in \{f, i, o, c\}$. $\odot$ denotes the Hadamard (element-wise) product and $\sigma$ is the sigmoid function.  

\subsubsection{Gated Recurrent Units (GRU)}

The GRU \cite{cho2014learning,chung2014gru} is a variation of the LSTM that reduces its parameter count by combining the input and forget gates into a singular update gate. 
Let $\mathbf{z}_{t} = \left(\mathbf{h}_{t-1}, \mathbf{o}_t \right)$. The GRU updates its memory in time through
\begin{align}
    \label{eq:gru_hidden}
    \mathbf{h}_t & = \mathbf{g}^z_t \odot \mathbf{\Tilde{h}}_t  + (1 - \mathbf{g}^z_t) \odot \mathbf{h}_{t-1}, & \\ 
    \label{eq:gru_reset}
    \mathbf{\Tilde{h}}_t & = \mathrm{tanh}\left(\mathbf{W}_h \left(\mathbf{g}^r_t \odot \mathbf{h}_{t-1}, \mathbf{o}_t\right) + \mathbf{b}_h \right), &
    \\
    \label{eq:gru_gate_vec}
    \mathbf{g}_t^k  & = \sigma  \left( \mathbf{W}_k\mathbf{z}_{t} + \mathbf{b}_k \right), & k \in \{z, r\}
\end{align}
where $\mathbf{g}^z_t, \mathbf{g}^r_t \in \mathbb{R}^{d_h}$ are gate vectors (update and reset gate, respectively). 
The learnable parameters are the weight matrices $\mathbf{W}_k \in \mathbb{R}^{d_h \times (d_h + d_o)}$ and the bias vectors $\mathbf{b}_k \in \mathbb{R}^{d_h}$ for $k \in \{z, r, h\}$. 
The update gate $\mathbf{g}^z_t$ functions similarly to input and forget gates in the LSTM by determining which information is used to update the hidden state.
By tying the gate outputs together with the constraint that they add to one element-wise, the GRU reduces its parameter count. 
Additionally, the GRU neglects an output gate, instead introducing a reset gate $\mathbf{g}^r_t$ as a mechanism to reset its memory directly prior to computing the memory update.

\subsubsection{Recurrent Highway Network (RHN)}

Recurrent Highway Networks (RHN) \cite{zilly2017recurrent} expand RNN cells to enable deep transition functions in time. 
While LSTMs and GRUs only allow step-to-step transition depths of $L=1$, RHN cells can have arbitrary transition depths.
Let $\mathbf{z}_t^\ell = \left(\mathbf{o}_t \mathbb{I}_{\{\ell = 1\}}, \mathbf{h}_t^{\ell - 1}\right)$ such that $\mathbf{z}_t^\ell = \mathbf{h}_t^{\ell -1}$ for $\ell > 1$. 
The RHN defines a step-to-step transition depth $L$ ($\ell \in \{1, 2, ..., L\}$) through
\begin{align}
    \label{eq:rhn_highway}
    \mathbf{h}^\ell_{t} & = \mathbf{g}_t^{r, \ell} \odot \mathbf{s}_t^\ell + \mathbf{g}_t^{c, \ell} \odot \mathbf{h}_t^{\ell - 1}, & \\
    \mathbf{s}_t^\ell & = \mathrm{tanh}\left( \mathbf{W}_{s_\ell} \mathbf{z}_t^\ell + \mathbf{b}_{s_\ell} \right), & \\
    \label{eq:rhn_gate_vec}
    \mathbf{g}_t^{k, \ell}
    & = 
    \sigma\left( \mathbf{W}_{k_\ell} \mathbf{z}_t^\ell + \mathbf{b}_{k_\ell} \right), & k \in \{r, c\}
\end{align}
where 
\begin{align}
    \mathbf{h}_t^0 = \mathrm{tanh}\left( \mathbf{W}_0 \left(\mathbf{o}_t, \mathbf{h}_{t-1}^L\right) + \mathbf{b}_0\right)
\end{align}
and $\mathbf{h}_t^\ell \in \mathbb{R}^{d_h}$ are intermediate outputs, $\mathbf{g}_t^{r,\ell}, \mathbf{g}_t^{c, \ell} \in \mathbb{R}^{d_h}$ are transfer and carry gates respectively, and $\mathbf{s}_t^\ell \in \mathbb{R}^{d_h}$ is a hidden state. 
Critically, the RHN achieves deeper transition depths through the usage of highway connections controlled by learnable gating mechanisms, as seen in~\Cref{eq:rhn_highway}. 
Mirroring the gated structure of a GRU, \cite{zilly2017recurrent} suggests a coupled gating structure for the transfer and carry gates, $\mathbf{g}_t^{r, \ell} = 1 - \mathbf{g}_t^{c, \ell}$, which we assume as the canonical RHN gating style. 
Without this simplification, the RHN has learnable weight matrices $\mathbf{W}_0, \mathbf{W}_{k_1} \in \mathbb{R}^{d_h \times (d_o + d_h)},\ \mathbf{W}_{k_\ell} \in \mathbb{R}^{d_h \times d_h}$ and learnable bias vectors $\mathbf{b}_{0}, \mathbf{b}_{k_1}, \mathbf{b}_{k_\ell} \in \mathbb{R}^{d_h}$ for $\ell \in \{1, ..., L\}$ and $k \in \{s, r, c\}$.

\subsubsection{Output Mapping}  

For all classes of RNNs, the output mapping is a simple, affine projection from the final hidden representation at time $t$, $\mathbf{h}_t$, to the predicted observable $\hat{\mathbf{o}}_{t+1}$.
\begin{equation}
    \mathbf{\hat{o}}_{t+1} = \mathbf{W}_o \mathbf{h}_t + \mathbf{b}_o,
\end{equation}
with $\mathbf{W}_o \in \mathbb{R}^{d_o \times d_h}$ and $\mathbf{b}_o \in \mathbb{R}^{d_o}$.

\subsection{Transformers}

The Transformer architecture \cite{vaswani2017attention}, first originating in natural language processing (NLP), is a non-recurrent architecture that has seen widespread adoption across a broad spectrum of domains.
Transformers were developed in response to the slow training speeds of RNNs, and as such, they sacrifice the recurrent data processing for training speed.
To remain effective at sequential processing tasks, Transformers employ architectural adjustments to help emulate sequential processing without sacrificing the speed afforded by non-recurrent training. 
While the Transformer architecture is generally defined as an encoder-decoder network, this work focuses on decoder-only Transformers similar to GPT-2 in NLP \cite{radford2019language} which have seen prior success in forecasting dynamical systems \cite{geneva2022transformers}, and still serve as the backbone of modern language models like ChatGPT and GPT-4 \cite{openai2023gpt4}.

As Transformers lack sequence-order a priori, typically a Transformer has an initial input stage where an observable is encoded as a higher-dimensional hidden vector and combined with a fixed absolute position vector \cite{vaswani2017attention}. 
However, in this work, we omit the absolute position embedding as we found absolute positional information to be detrimental to the convergence of models. 
Instead, we focus on endowing Transformers with positional sensitivity through a relative bias in their self-attention operations \cite{huang2020improve,raffel2020exploring,shaw2018self}. 
Thus, an initial observable vector is lifted to the latent dimension of the model with an affine transformation, pointwise non-linearity, and an (optional) application of dropout \cite{srivastava2014dropout},
\begin{equation}
    \mathbf{h}_t^0 = \mathrm{Dropout}\left(g\left(\mathbf{W}_i \mathbf{o}_t + \mathbf{b}_i \right)\right),
\end{equation}
with learnable $\mathbf{W}_i \in \mathbb{R}^{d_h \times d_o}$ and $\mathbf{b}_i \in \mathbb{R}^{d_h}$.

\subsubsection{Transformer Blocks}

After computing an initial hidden representation $\mathbf{h}_t^0 \in \mathbb{R}^{d_h}$ position-wise, the blocked sequence $\mathbf{h}_{t-S:t}^0$ is fed through a stack of transformer blocks, $\ell \in \{0, 1, ..., L - 1\}$. For the $\ell$-th block, the mapping from block $\ell$ to $\ell+1$ is defined as:
\begin{align}
    \label{eq:residual1}
    \mathbf{\Tilde{h}}_{t-S:t}^\ell & = 
    \mathbf{h}_{t-S:t}^\ell + \mathrm{MHA}(\mathrm{LN}(\mathbf{h}_{t-S:t}^\ell), \mathrm{LN}(\mathbf{h}_{t-S:t}^\ell)), \\ 
    \label{eq:residual2}
    \mathbf{h}_{t-S:t}^{\ell+1} & = 
    \mathbf{\Tilde{h}}_{t-S:t}^\ell + \mathrm{MLP}(\mathrm{LN}(\mathbf{\Tilde{h}}_{t-S:t}^\ell)),
\end{align}
where MHA is multi-head attention (specifically, self-attention), LN is layer normalization \cite{ba2016layer} (with learnable $\gamma, \beta \in \mathbb{R}^{d_h}$ for scaling and centering data), and MLP is a 2-layer multi-layer perceptron of the form:
\begin{equation}
    f(\mathbf{x}_t) = \mathrm{Dropout}(\mathbf{W}_{\text{out}} g(\mathbf{W}_{\text{in}} \mathbf{x}_t + \mathbf{b}_{\text{in}}) + \mathbf{b}_{\text{out}}).
\end{equation}

\begin{figure}[htb!]
    \centering
    \includegraphics[width=0.4\textwidth]{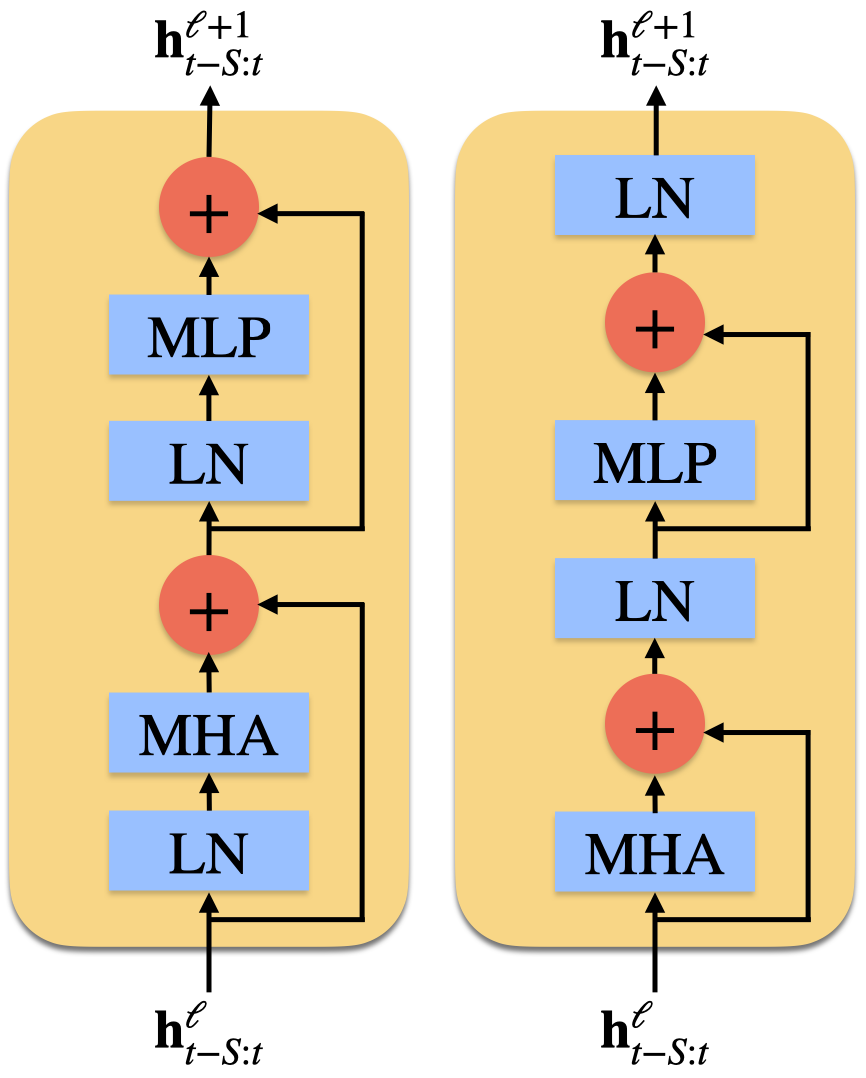}
    \caption{Visualization of a pre-layer normalization (left) or post-layer normalization (right) Transformer block. 
    }
    \label{fig:pre_post_ln}
\end{figure}

Note that \Cref{eq:residual1,eq:residual2} express a so-called \emph{pre-layer normalization} variant of the Transformer block, where layer normalization is performed prior to attention or the MLP. This differs from the original Transformer \cite{vaswani2017attention} and previous work on Transformers for dynamical systems \cite{geneva2022transformers}, both of which use a \emph{post-layer normalization} variant. 
There is theoretical and empirical evidence that the choice of pre- vs. post-layer normalization has significant effects on the stability of gradient updates and smoothness of the overall loss landscape \cite{nguyen2019transformers,liu2020understanding,shleifer2021normformer,xiong2020layer, wang2022deepnet}. 
This work considers the position of layer normalization as an additional hyperparameter for the Transformer architecture.
These two normalization organizations are depicted in~\Cref{fig:pre_post_ln}.

\subsubsection{Output Mapping}  

The output mapping for the Transformer is similar to that of the RNNs, with the addition of a layer normalization operation at the final layer for pre-layer normalization Transformer variants,
\begin{align}
    \hat{\mathbf{o}}_{t+1} & = \mathbf{W}_o \text{LN}(\mathbf{h}^L_t) + \mathbf{b}_o,
\end{align}
with learnable $\mathbf{W}_o \in \mathbb{R}^{d_o \times d_h}$ and $\mathbf{b}_o \in \mathbb{R}^{d_o}$. Additionally, whereas RNNs output their latent, compressed state as the history $\mathbf{H}_t$, Transformers maintain a cache of the $S$ most recent time steps in a FIFO (first in, first out) stack or queue.

\subsection{Neural Mechanisms}
Both RNNs and Transformers are equipped to model sequences, but the neural mechanisms they use to do so differ. In this work, we consider these mechanistic differences and attempt to form novel blends of architectures by applying useful mechanisms more broadly. This work considers three key mechanisms: gating, attention, and recurrence.

\subsubsection{Neural Gating (NG)}

\begin{figure*}[htb!]
    \centering
    \begin{subfigure}{0.25\textwidth}
        \centering
        \includegraphics[width=.9\linewidth]{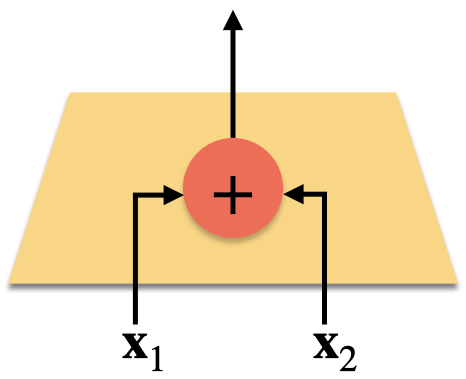}
        \caption{Additive}
        \label{fig:gates_add}
    \end{subfigure}%
    \begin{subfigure}{0.25\textwidth}
        \centering
        \includegraphics[width=.9\linewidth]{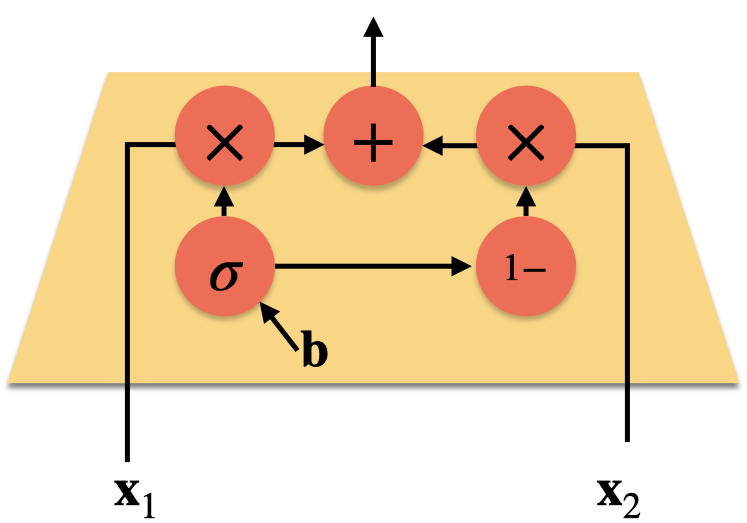}
        \caption{Learned Rate}
        \label{fig:gates_fixed}
    \end{subfigure}%
    \begin{subfigure}{0.25\textwidth}
        \centering
        \includegraphics[width=.9\linewidth]{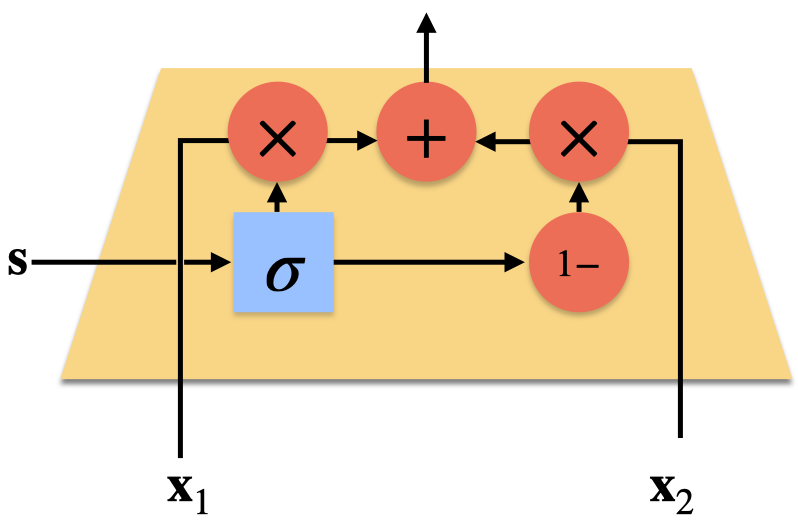}
        \caption{Dependent-Coupled}
        \label{fig:gates_coupled}
    \end{subfigure}%
    \begin{subfigure}{0.25\textwidth}
        \centering
        \includegraphics[width=.9\linewidth]{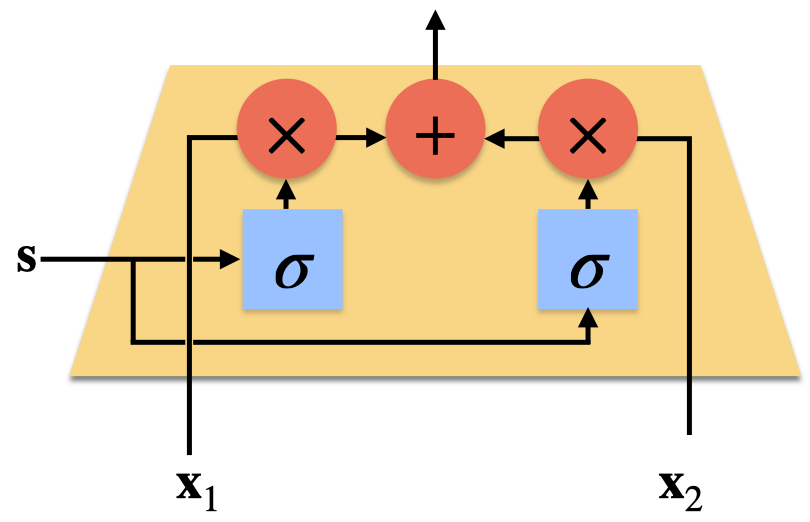}
        \caption{Dependent}
        \label{fig:gates_dual}
    \end{subfigure}%
    \caption{
        Visualized information flows through neural gates. 
        Gate types are of increasing complexity from left to right. 
        Orange circles denote element-wise operations without learnable parameters, yellow rectangles denote a full layer. 
    }
    \label{fig:gate_types}
\end{figure*}

In this work, we use \emph{gating mechanism} to refer to a grouping of operations that receives as input two vectors and (possibly) a selection vector, and outputs an aggregated vector of the inputs.
This may be viewed as a multiplexing of the information flow in the neural network \cite{baldi2023quarks}, introducing an efficient, sparse approximation of quadratic interactions between two input vectors. As this is a very general concept, we present 
\begin{equation}
    G_\theta(\mathbf{x}_1, \mathbf{x}_2, \mathbf{s}) = g_{1}( \mathbf{s}) \odot  \mathbf{x}_1 + g_{2}( \mathbf{s}) \odot  \mathbf{x}_2,
\end{equation}
as the form of the parameterized gating mechanisms considered here, where 
$\theta$ are any learnable parameters associated with the gate, 
$\mathbf{x}_i \in \mathbb{R}^{d_x}$ are the vectors to be multiplexed,
and $\mathbf{s} \in \mathbb{R}^{d_s}$ is the selection vector. 
Since we allow for the possibility of $d_s \neq d_x$ and require that a gate is element-wise bounded on $[0, 1]$, $g_{k}: \mathbb{R}^{d_s} \mapsto \left[0, 1\right]^{d_x}$.

Such NG is essential for the stabilization of RNNs \cite{greff2016lstm}. As an example, consider the LSTM's input and forget gate (\Cref{eq:lstm_gate_vec}), and their role in updating the LSTM's cell state (\Cref{eq:lstm_gate}).
An algebraic re-phrasing allows the expression of this structure as: 
\begin{align}
    g_{k}(\mathbf{z}_t) = \sigma \left(\mathbf{W}_k\mathbf{z}_t + \mathbf{b}_k \right)  = \mathbf{g}_t^{k}, \ k \in \{f, i\} \\ 
    G_{\theta}(\mathbf{c}_{t-1}, \mathbf{\Tilde{c}}_t, \mathbf{z}_t) = g_{f}(\mathbf{z}_t) \odot \mathbf{c}_{t-1}+ g_{i}(\mathbf{z}_t) \odot \mathbf{\Tilde{c}}_t,
\end{align}
where $\theta = \bigcup_{k \in \{f, i\}}\{ \mathbf{W}_k, \mathbf{b}_k\}$. 
Similar algebraic formulations are possible with both the update gate in the GRU (\Cref{eq:gru_hidden,eq:gru_gate_vec}) and the gates in the RHN (\Cref{eq:rhn_highway,eq:rhn_gate_vec}). Furthermore, gating implicitly appears within Transformers in their residual connections (\Cref{eq:residual1,eq:residual2}), functioning as a parameter-less additive gate where $g_k$ outputs a constant vector of ones for all inputs.

Note that this definition specifically excludes gates like the output gate $\mathbf{g}_t^o$ in the LSTM (\Cref{eq:lstm_gate}) and the reset gate $\mathbf{g}_t^r$ in the GRU (\Cref{eq:gru_reset}) which do not multiplex two input vectors but instead modulate the magnitude of one vector conditioned on another.
While these gates are still interesting, we restrict our study to multiplexing gates with two inputs.

We consider four types of gates in this work.
The information flows through the considered gates are depicted in~\Cref{fig:gate_types}.
Mathematically, we can describe these gates as:
\begin{itemize}
    \item \textbf{Additive (A)}: 
    \begin{align}
        g_{k }(\mathbf{s}) = \mathbf{1}, \quad k = 1, 2
    \end{align}
    This gate is static, operating independently of the input data. It has no learnable parameters $\theta$ and is the default gate in Transformer residual connections (\Cref{eq:residual1,eq:residual2}).
    
    \item \textbf{Learned Rate (L)}: 
    \begin{align}
        g_{1}(\mathbf{s}) = \sigma(\mathbf{b}), & \quad  g_{2}(\mathbf{s}) = \mathbf{1} - g_{1}(\mathbf{s}) ,
    \end{align} 
    where $\sigma$ is a sigmoid, $\mathbf{b} \in \mathbb{R}^{d_x}$, and $\theta = \{\mathbf{b}\}$. This gate style learns dimension-specific ratios for adding vectors, operating as a learned, weighted averaging \cite{hutchins2022block}.
    \item \textbf{Input-Dependent (Coupled, C)}:
    \begin{align}
        g_1(\mathbf{s}) & = \sigma(\mathbf{W} \mathbf{s} + \mathbf{b}), & \quad  g_2(\mathbf{s}) = \mathbf{1} - g_1(\mathbf{s}),
    \end{align}
    where $\sigma$ is a sigmoid, $\mathbf{W} \in \mathbb{R}^{d_x \times d_s}$, $\mathbf{b} \in \mathbb{R}^{d_x}$, and $\theta = \{\mathbf{W}, \mathbf{b}\}$. This gate is dynamic, modulating information conditionally on $\mathbf{s}$. It couples the two inputs, enforcing a constraint that their output elements must sum to 1. 
    \item \textbf{Input-Dependent (D)}: 
    \begin{align}
        g_k(\mathbf{s}) = \sigma(\mathbf{W}_k \mathbf{s} + \mathbf{b}_k), & \quad
        k = 1, 2 
    \end{align}
    where $\sigma$ is a sigmoid, $\mathbf{W}_k \in \mathbb{R}^{d_x \times d_s}$, $\mathbf{b}_k \in \mathbb{R}^{d_x}$, and $\theta = \bigcup_{k =1}^2\{ \mathbf{W}_k, \mathbf{b}_k \}$. Like the previous gate, it dynamically modulates its input conditionally based on $\mathbf{s}$. However, the element-wise sum of the gates is unconstrained.
\end{itemize}
The type of binary gating mechanism in each architecture is treated as a categorical hyperparameter. 
We keep the input structure of data-dependent RNN gates unaltered. 
For Transformers, we set $\mathbf{x}_1, \mathbf{x}_2$ to the residual and updated hidden vectors, respectively, with the conditional input set to their concatenation: $\mathbf{s} = \left(\mathbf{x}_1, \mathbf{x}_2\right)$.

\subsubsection{Attention (AT)}

\begin{figure}[htb!]
    \centering
    \includegraphics[width=0.5\textwidth]{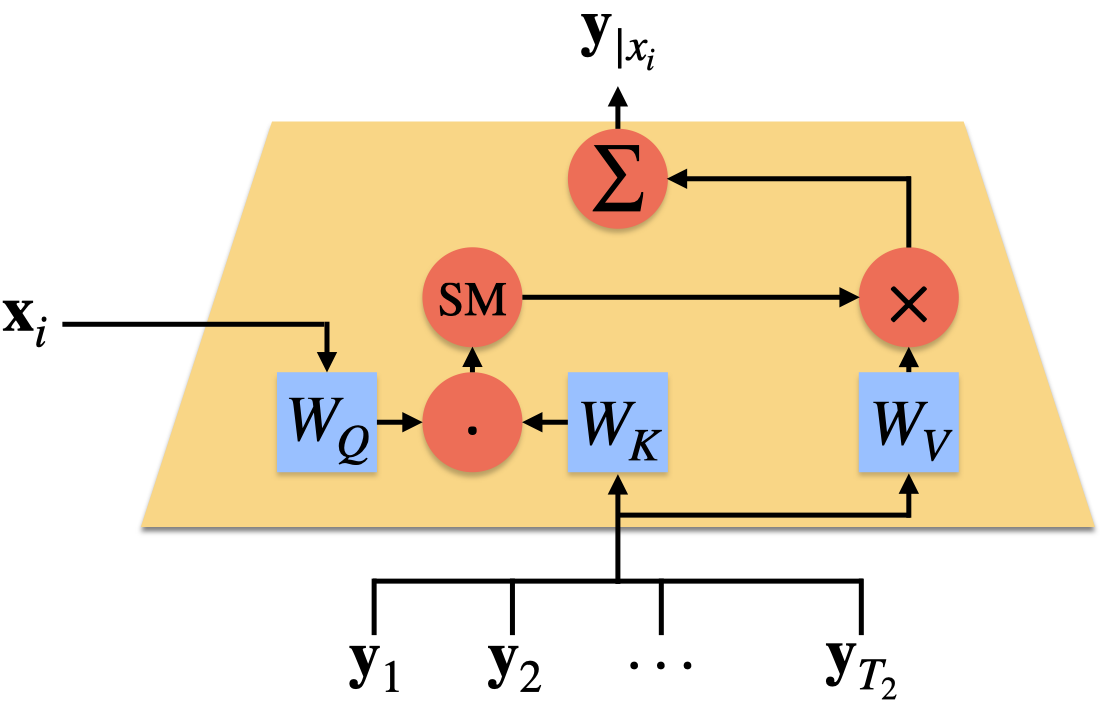}
    \caption{
        Scaled dot-product attention as a gate. 
        $\mathbf{x}_i$ is a single element in a sequence $\mathbf{X} \in \mathbb{R}^{T_1 \times d}$ gating sequence $\mathbf{Y} \in \mathbb{R}^{T_2 \times d}$.
        A full attention operation consists of $T_1$ circuits executed in parallel, one for each element $\mathbf{x}_i \in \mathbf{X}$.
        ${\rm SM}$ denotes Softmax.
    }
    \label{fig:attn_gate}
\end{figure}

A central mechanism in the Transformer architecture is the multi-headed, scaled dot-product attention operation (MHA). 
It is the mechanism Transformers use to propagate information in time.  
Consider two sequences represented by matrices $\mathbf{X} \in \mathbb{R}^{T_1 \times d}, \mathbf{Y} \in \mathbb{R}^{T_2 \times d}$. 
For simplicity, consider a single ``head'' of scaled-dot product attention. 
Here, $\mathbf{X}$ is used to dynamically compute a representation of $\mathbf{Y}$:
\begin{align}
    \label{eq:attn1}
    \mathrm{A}\left(\mathbf{X}, \mathbf{Y}\right) & = \mathrm{Softmax}\left(\frac{\left( \mathbf{X}  \mathbf{W}_Q \right)\left( \mathbf{Y}\mathbf{W}_K  \right)^\top }{\sqrt{d_k}} \right) \left( \mathbf{Y} \mathbf{W}_V\right), \\
    & = \mathrm{Softmax}\left(\frac{\mathbf{Q}_x \mathbf{K}_y^\top }{\sqrt{d_k}} \right) \mathbf{V}_y,
\end{align}
where $\mathbf{W}_q, \mathbf{W}_k \in \mathbb{R}^{d \times d_k}$, $\mathbf{W}_v \in \mathbb{R}^{d \times d_v}$ are learnable weight matrices. 
After this operation, $\mathrm{A}\left(\mathbf{X}, \mathbf{Y}\right) \in \mathbb{R}^{T_1 \times d_v}$ contains information from $\mathbf{Y}$ aligned temporally with $\mathbf{X}$. 
In other words, $\mathbf{Y}$ \emph{gated} by $\mathbf{X}$, emphasizing the connection between neural gating and attention \cite{baldi2023quarks}.
We depict this connection, visualizing a single element $\mathbf{x}_i$ gating a sequence $\mathbf{Y}$ in~\Cref{fig:attn_gate}.
Self-attention is a special case of this operation when $\mathbf{X} = \mathbf{Y}$.

To further increase throughput on modern hardware, Transformers leverage multiple attention operations (\Cref{eq:attn1}) in parallel.
Each attention operation is called a head and the collection is called multi-headed attention.
To ensure that the latent dimension of the sequence is invariant to the attention operation, the parallel results of each head are concatenated and passed through a final affine transformation: 
\begin{align}
    \mathrm{MHA}(\mathbf{X}, \mathbf{Y})
    & = 
    \mathbf{W}_o \left(\mathrm{A}_i(\mathbf{X}, \mathbf{Y})\right)_{i=1}^{H} + \mathbf{b}_o,
\end{align}
where $H$ is the number of parallel attention heads and $\mathbf{W}_o \in \mathbb{R}^{d \times H\cdot d_v}$, $\mathbf{b}_o \in \mathbb{R}^{d}$ are learnable parameters. A standard practice is to set $d_k = d_v = d / H$
so that the FLOP cost is constant no matter the number of heads used,
though this requires that $H$ is an integral divisor of $d$.

By default, such self-attention is a permutation invariant operation. Two key changes are necessary to use it for forecasting sequential data. The first is to introduce a triangular mask $\mathbf{M} \in \mathbb{R}^{T_1 \times T_1}$ to the attention operation
\begin{align}
    \label{eq:attn2}
    \mathrm{A}\left(\mathbf{X}, \mathbf{X}\right)
    & = \mathrm{Softmax}\left(\frac{\mathbf{Q} \mathbf{K}^\top \odot  \mathbf{M}}{\sqrt{d_k}} \right) \mathbf{V} \\ 
    \mathbf{M}_{ij} & = \begin{cases}
        1 & i \leq j  \\ 
        -\infty & i > j \\ 
    \end{cases}
\end{align}
to prevent information flow from the future to the past during batched training. Without the mask, attention is an all-to-all temporal computation, with each time step possibly attending to any other time step.

The second critical modification to the attention operation is to add a relative distance bias to the attention computation to instill a notion of sequential order in the Transformer. Consider the computation of one vector $\mathbf{x}_i \in \mathbf{X}$ attending to the full matrix $\mathbf{X}$
\begin{align}
    \mathbf{q}_i = \mathbf{x}_i  \mathbf{W}_Q, \quad & \mathbf{K} = \mathbf{X}  \mathbf{W}_K, \quad \mathbf{V} = \mathbf{X}  \mathbf{W}_V, \\
    \label{eq:attn_probs}
     \mathrm{A}(\mathbf{x}_i, \mathbf{X}) = \sum_{j = 1}^{i} \alpha_{ij} \mathbf{v}_j, \quad & \alpha_{ij} = \frac{\mathrm{exp} \left( \mathbf{q}_i \mathbf{k}_j^\top\right) }{\sum_{k=1}^i \mathrm{exp}\left( \mathbf{q}_i \mathbf{k}_k^\top \right) },
\end{align}
where division by $d_k^{-1/2}$ is omitted from~\Cref{eq:attn_probs} for clarity.
The simplest modification that adds a relative distance bias to attention is to adjust the computation of $\alpha_{i,j}$ (the attention weight) to include a scalar bias that is dependent on the relative distance between two timesteps. We consider two classes of relative attention biasing: 
\begin{itemize}
    \item \textbf{Data-Independent Bias} \cite{raffel2020exploring,dai2021coatnet}: A data-independent bias only considers the distance between two timesteps, not their latent representations, resulting in an attention weight computation of the form: 
\begin{align}
        \alpha_{ij} & = \frac{\mathrm{exp} \left( \mathbf{q}_i \mathbf{k}_j^\top + \omega_{|i-j|} \right) }{\sum_{k=1}^i \mathrm{exp}\left( \mathbf{q}_i \mathbf{k}_k^\top  + \omega_{|i-k|} \right) },
    \end{align}
	where $\mathbf{\omega} \in \mathbb{R}^{T_1}$ is a learned vector that biases attention weights based on the temporal distance between $\mathbf{q}_i, \mathbf{k}_j$. 
    \item \textbf{Data-Dependent Bias} \cite{dai2019transformer}: Alternatively, one can use a relative positional bias that is dependent on the current input data:
	\begin{align}
        \hat{\alpha}_{ij} = \mathbf{q}_i^\top \mathbf{k}_j & + \mathbf{q}_i^\top \mathbf{r}_{|i-j|} + \mathbf{u}^\top \mathbf{k}_j + \mathbf{v}^\top \mathbf{r}_{|i-j|}, \\
        \alpha_{ij} & = \frac{\mathrm{exp} \left( \hat{\alpha}_{ij} \right) }{\sum_{k=1}^i \mathrm{exp}\left( \hat{\alpha}_{ik} \right) } , 
    \end{align}
	where $\mathbf{u}, \mathbf{v} \in \mathbb{R}^{d_k}$ are learnable vectors abstracting interactions between positional bias and vector content, and $\mathbf{r} \in \mathbb{R}^{T_1 \times d_k}$ is a matrix of learned positional vectors. 
    Data interacts with the positional bias through the dot products of the query with $\mathbf{r}$ and the key with $\mathbf{u}$. 
    While strictly more expressive than a data-independent relative bias, a data-dependent bias has many more parameters to learn and can be susceptible to overfitting. 
\end{itemize}

Thus far, attention has been discussed in the context of the Transformer architecture, however, attention was initially invented for recurrent models that help align sequences and combat the information bottleneck that naturally occurs over long sequence lengths \cite{bahdanau2014neural}.
Multi-headed attention can be used to augment any recurrent architecture by allowing recurrent architectures to construct a new, contextualized hidden state based on the last $S$ hidden states.
This formulation is particularly useful, especially for stacked RNNs where multi-headed attention can be used to refine hidden states in-between recurrent layers, attending back to itself, the original input, or some other previous recurrent layer,
\begin{align}
    \mathbf{\Tilde{h}}^\ell_{t-S:t} = \mathrm{MHA}(\mathbf{h}^\ell_{t-S:t}, \mathbf{h}^k_{t-S:t}),
\end{align}
for $k \in \{0, ..., \ell\}$. For RNNs that use attention, we consider both types of relative position bias as well as MHA without any bias term since RNNs have an implicit representation of sequential order.

\subsubsection{Recurrence}

RNNs are recurrent and stateful, propagating temporal information forward in time with a compressed summary $\mathbf{h}_t$. 
Transformers are feed-forward and stateless models, lacking sequence compression and recurrence. 
This is by design as Transformers were invented to drop recurrence in favor of feed-forward, attention-only architectures. 
That said, it remains an open question as to when and in what capacity recurrence might be added to a Transformer to increase performance.
Several works have explored augmentations to Transformers to add recurrence \cite{didolkar2022temporal,hutchins2022block} or feedback \cite{fan2020addressing}, aligning Transformers more closely with the stateful behavior of RNNs.

This work follows that of the Temporal Latent Bottleneck \cite{didolkar2022temporal}, adding recurrence to the Transformer through the introduction of a memory matrix, $\mathbf{M} \in \mathbb{R}^{N \times d_M}$, propagated across sequence chunks. 
The addition of recurrence adds a horizontal pathway through the Transformer block which necessitates the introduction of two cross-attention operations that allow the memory matrix to query information from the current sequence block ${\rm MHA}(\mathbf{M}, \mathbf{X})$ and the current sequence block to query the memory ${\rm MHA}(\mathbf{X}, \mathbf{M})$. To minimize the parameter increase, we only consider one recurrent Transformer block and always place it last in the Transformer stack.

\subsection{Training}

All model evaluations are mapped to a single Nvidia Tesla P100 GPU and are executed on the XC50 compute nodes of the Piz Daint supercomputer at the Swiss national supercomputing center (CSCS).

\subsubsection{Batching Training Data}

All models are trained to forecast observables via the backpropagation-through-time (BPTT) algorithm \cite{werbos1990backpropagation,werbos1988generalization}. 
Models are trained with teacher forcing \cite{williams1989learning}. 
Considering models trained with free-running predictions or more advanced curricula \cite{bengio2015scheduled,lamb2016professor,vlachas2023learning} is beyond the scope of this work, but would be of interest in future studies.

All models have a training sequence length parameter $S$ and a prediction length parameter $1 \leq S' \leq S$.
Sequences are chunked into sub-sequences of length $S$.
For stateless models, this fixes the temporal horizon over which forecasting may depend. 
For stateful models, information still propagates across each sub-sequence via the hidden state. 
The actual gradient for BPTT is computed with respect to the error of the final $S'$ predictions of each sub-sequence. 
When $S' = S$, models must predict the temporal evolution of every observable, including the first observable in the sub-sequence which has no prior context (beyond the hidden state for stateful models). 
When $S'= 1$, models predict the final step of each sub-sequence using the full context.


\subsubsection{Optimizer and Scheduling}

All models are trained using the Adabelief optimizer \cite{zhuang2020adabelief}. 
Models are trained using validation-based early stopping with a step-wise learning rate schedule. 
The learning rate of a model is decreased by a multiplicative factor $\gamma<1$ after the validation loss fails to improve for $P$ epochs, denoting the end of a training round.
The models are trained for $R$ such rounds in total.
The values of the hyper-parameters $\gamma, P$ and $R$ are reported in the~\Cref{app:hyper}.

\subsubsection{Hyperparameter Optimization}

Each model considered here contains a large number of hyperparameters, determining both a model's architectural form as well as its training process. For each dataset, we discretize the hyperparameter search space and perform a grid search with Weights \& Biases \cite{wandb} to identify performant hyperparameterizations. 
We provide all hyperparameter search spaces in~\Cref{app:hyper}.

\section{Comparison Metrics}
\label{sec:metrics}

\subsection{NRMSE of Forecast Observable}

As a primary metric for forecasting performance,
we report the normalized root mean square error (NRMSE)
given by
\begin{align}
    \label{eq:NRMSE}
    \mathrm{NRMSE}(\mathbf{o}_t, \hat{\mathbf{o}}_t) & = 
    \sqrt{\left\langle \frac{\left(\hat{\mathbf{o}}_t - \mathbf{o}_t \right)}{\mathbf{\sigma}^2} \right\rangle},
\end{align}
where $\mathbf{o}_t \in \mathbb{R}^{d_o}$ is the true observable
at time $t$, 
$\hat{\mathbf{o}}_t \in \mathbb{R}^{d_o}$ is the model 
prediction, 
and $\mathbf{\sigma} \in \mathbb{R}^{d_o}$ 
is the standard deviation over time 
for each state component.
In~\Cref{eq:NRMSE},
$\langle \cdot \rangle$ denotes
the average over the state space. 
In order to alleviate 
the effects of initial conditions,
we report the time evolution of the NRMSE 
averaged over 100 randomly sampled initial conditions.

\subsection{Valid Prediction Time (VPT)}
\label{sec:vpt}

A fundamental way to characterize chaotic motion is through Lyapunov exponents \cite{ott2002chaos}. 
Consider two infinitesimally-close initial conditions $\mathbf{o}_0, \mathbf{o}_0 + \delta \mathbf{o}_0$. 
Their separation, $|\delta \mathbf{o}_t|$, diverges exponentially on average in time, i.e., $|\delta\mathbf{o}_t|/|\delta\mathbf{o}_0| \sim \mathrm{exp}\left(\Lambda t\right)$ as $t \rightarrow \infty$. 
Note that the displacement $\delta \mathbf{o}_t$ is a vector displacement in the state space, and in general, the Lyapunov exponent $\Lambda$ will depend on the orientation $(\delta\mathbf{o}_t/\|\delta\mathbf{o}_t\|)$ of this displacement. 
As infinitesimal differences can amplify (or contract) in various directions in the limit $t \rightarrow \infty$, there are multiple $\Lambda$ values that correspond to different divergent directions in the state space.
Ordering these values $\Lambda_1 \geq \Lambda_2 \geq \Lambda_3 \geq ...$ forms what is called the Lyapunov exponent spectrum (LS) of the chaotic system. 

The maximal Lyapunov exponent (MLE) $\Lambda_1$ plays an important role in the predictability of the system. 
Chaotic motion is defined by $\Lambda_1 > 0$. 
Note that if one randomly selects an orientation $\delta \mathbf{o}_0$, trajectories will diverge at an exponential rate $\Lambda_1$ with probability approaching 1. 
This is due to the fact that for lower exponents $(\Lambda_2, \Lambda_3, ...)$ to be realized, the orientation of $\delta \mathbf{o}_0$ would need to lie on a subspace of lower dimensionality than the original state-space.
Hence, the typical rate at which nearby trajectories will diverge is $\Lambda_1$, and $T^{\Lambda_1} = \Lambda_1^{-1}$, the \emph{Lyapunov time}, provides a characteristic scale for analyzing the forecasts of a model.

Given a system with a known MLE $\Lambda_1$, we may compute a singular, normalized metric called the Valid Prediction Time (VPT) \cite{vlachas2020backpropagation}.
The VPT takes an error threshold $\epsilon$ and returns the maximal time step $t_f$ such that the preceding forecast exhibits an error beneath the target threshold:
\begin{align}
    t^* = \underset{t_f}{\mathrm{argmax}} \{t_f & | \mathrm{NRMSE}(\mathbf{o}_{t_f}, \hat{\mathbf{o}}_{t_f}) < \epsilon, \forall t \leq t_f \}, 
    \; 
    \mathrm{VPT}_\epsilon(\mathbf{o}_t, \hat{\mathbf{o}}_t) = \frac{t^*}{\Lambda_1}
\end{align}
Alternatively phrased, the VPT gives a sense of how long a model can accurately forecast the observable based on the underlying chaotic nature of the system.
We assume $\epsilon = 0.5$ unless otherwise stated. 

\subsection{Power Spectrum (PSD)}

To quantify how well different models capture the long-term statistics of a system, we evaluate the mean power spectral density (PSD, power spectrum) over all elements $i \in \{1, ..., d_o\}$ in the state vector $\mathbf{o}_t^i$.
The power spectrum of $\mathbf{o}_t^i$ over time is computed by
\begin{align}
    \text{PSD}(f) & = 
    20 \log_{10}\left(2|U(f)|\right)
\end{align}
where $U(f) = \text{FFT}(\mathbf{o}_t^i)$ is the complex Fourier spectrum of the state evolution.

\section{Forecasting Dynamics of Multiscale Lorenz-96}
\label{sec:lorenz}


\subsection{Multiscale Lorenz-96 Model}

The Lorenz-96 system \cite{lorenz1996predictability} is a prototypical model of the atmosphere around a latitude circle. 
The model features a slow-varying, large-scale set of quantities coupled with many fast-varying, small-scale variables. 
Despite its simplicity, Lorenz-96 captures realistic aspects of the Earth's atmosphere, like chaoticity and nonlinear interactions between separate spatial scales.
Because of this, the Lorenz-96 system has a history of use for testing data-driven, multiscale methods \cite{pathak2017using,pathak2018model,vlachas2020backpropagation}.
In this work, we use a recent extension of the Lorenz-96 system, which introduces a smaller-scale set of terms that vary even more rapidly \cite{thornes2017use}.
This multiscale Lorenz-96 is more difficult to model, with macro-, meso-, and micro-scale dynamics evolving according to the following equations:
\begin{align}
    \label{eq:lorenz_x}
    \frac{d X_k}{dt} & = X_{k-1} \left( X_{k+1} - X_{k-2} \right) - X_k + F - \frac{hc}{b} \sum_{j=1}^J Y_{j,k} \\
    \label{eq:lorenz_y}
    \frac{d Y_{j,k}}{dt} & = -cb Y_{j+1, k}(Y_{j+2, k} - Y_{j-1, k}) - c Y_{j,k} + \frac{hc}{b} X_k - \frac{he}{d}\sum_{i=1}^I Z_{i,j,k} \\ 
    \label{eq:lorenz_z}
    \frac{d Z_{i, j , k}}{dt} & = ed Z_{i-1, j, k}(Z_{i+1, j, k} - Z_{i-2, j, k}) - g_Z eZ_{i, j, k} + \frac{he}{d} Y_{j, k}
\end{align}
where $X, Y, Z$ are the macro-, meso-, and micro-scales, $F$ is the large-scale forcing applied to the system, $h=1$ is the coupling coefficient across spatial scales, $g_Z = 1$ is a damping parameter, and $b = c = d = e = 10$ are parameters prescribing the relative magnitudes and speeds between each scale. 
We let $K = J = I = 8$ for a full system dimensionality $D = K + J^2 + K^3 = 584$. 
All boundary conditions are assumed to be periodic.

In this work, we investigate whether models can learn to forecast the macro-scale dynamics of the multiscale Lorenz-96 system when only exposed directly to the macro-scale data $X_k$ during training. 
We consider two forcing values $F=10$ and $F = 20$, where the forcing parameter determines the overall chaoticity of the system.
We solve~\Cref{eq:lorenz_x,eq:lorenz_z} starting from a random initial condition with a Fourth Order Runge-Kutta scheme and a time-step of $\delta t = 0.005$. 
We run the solver up to $T = 2000$ after discarding transient effects ($T_\text{trans} = 1000$). 
The first $2 \cdot 10^5$ samples form the training set, with the rest held for testing.

As the multiscale Lorenz-96 system exhibits chaotic dynamics, it has a known Lyapunov time (the inverse of its MLE), $T^{\Lambda_1} = \Lambda_1^{-1}$. 
We use $\Lambda_1 = 2.2$ and $\Lambda_1 = 4.5$ for  $F=10$ and $F=20$, as reported in \cite{thornes2017use}. 
For both model classes, we run a large-scale hyperparameter sweep for $F=10$. 
Then, to test the sensitivity of hyperparameter selection to the forcing parameter, we consider a smaller search space for $F=20$ informed by the best parameters for $F=10$.
We describe the hyperparameter search space in~\Cref{app:hyper}.

\subsection{Results on the Multiscale Lorenz-96 Model}

\begin{figure*}[htb!]
    \centering
    Multiscale Lorenz-96 (F=10) 
    \begin{subfigure}{0.33\textwidth}
        \centering
        \includegraphics[width=\linewidth]{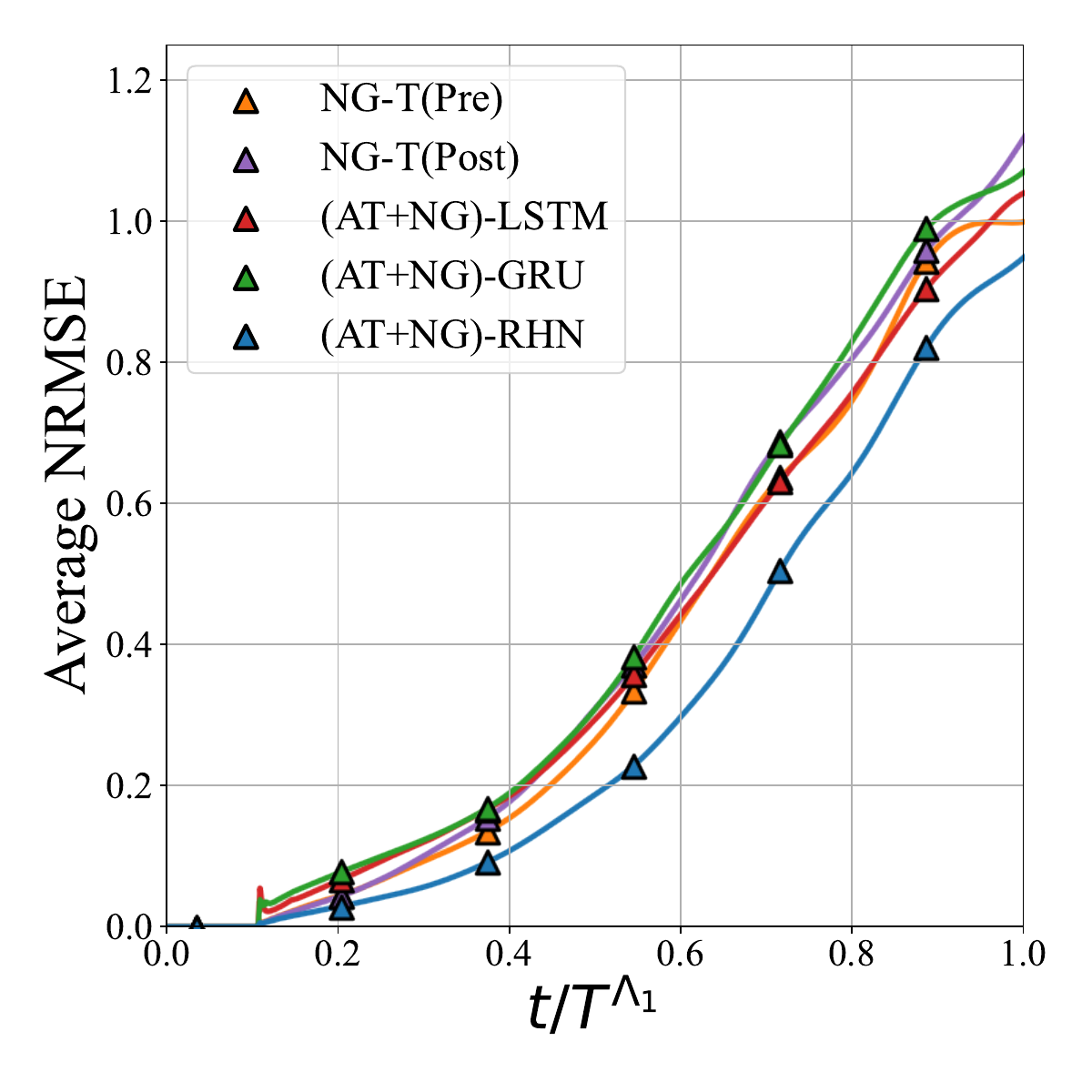}
        \caption{Best Models}
        \label{fig:l96_nrmse_all}
    \end{subfigure}%
    \begin{subfigure}{0.33\textwidth}
        \centering
        \includegraphics[width=\linewidth]{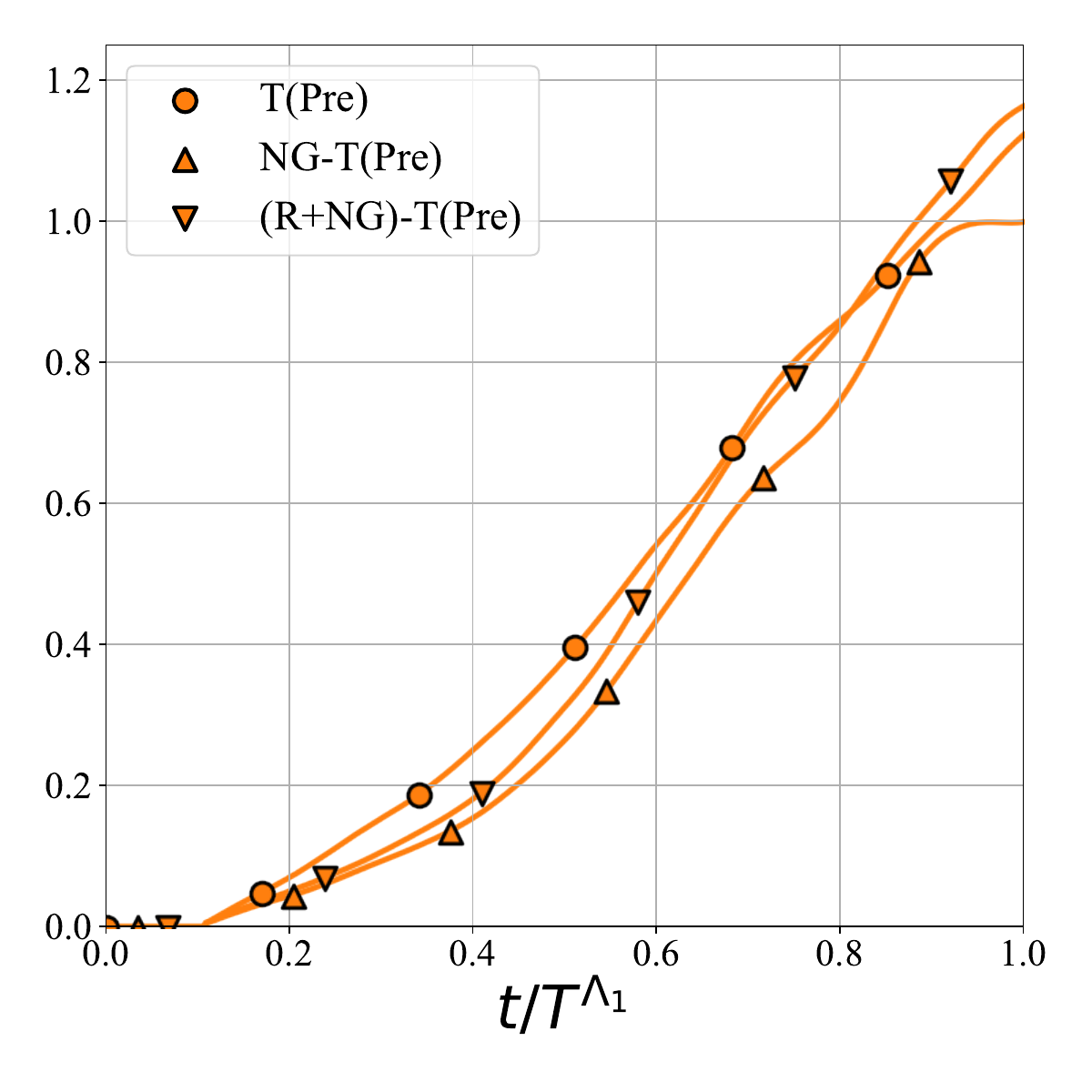}
        \caption{PreLN Transformer}
        \label{fig:l96_nrmse_pre}
    \end{subfigure}%
    \begin{subfigure}{0.33\textwidth}
        \centering
        \includegraphics[width=\linewidth]{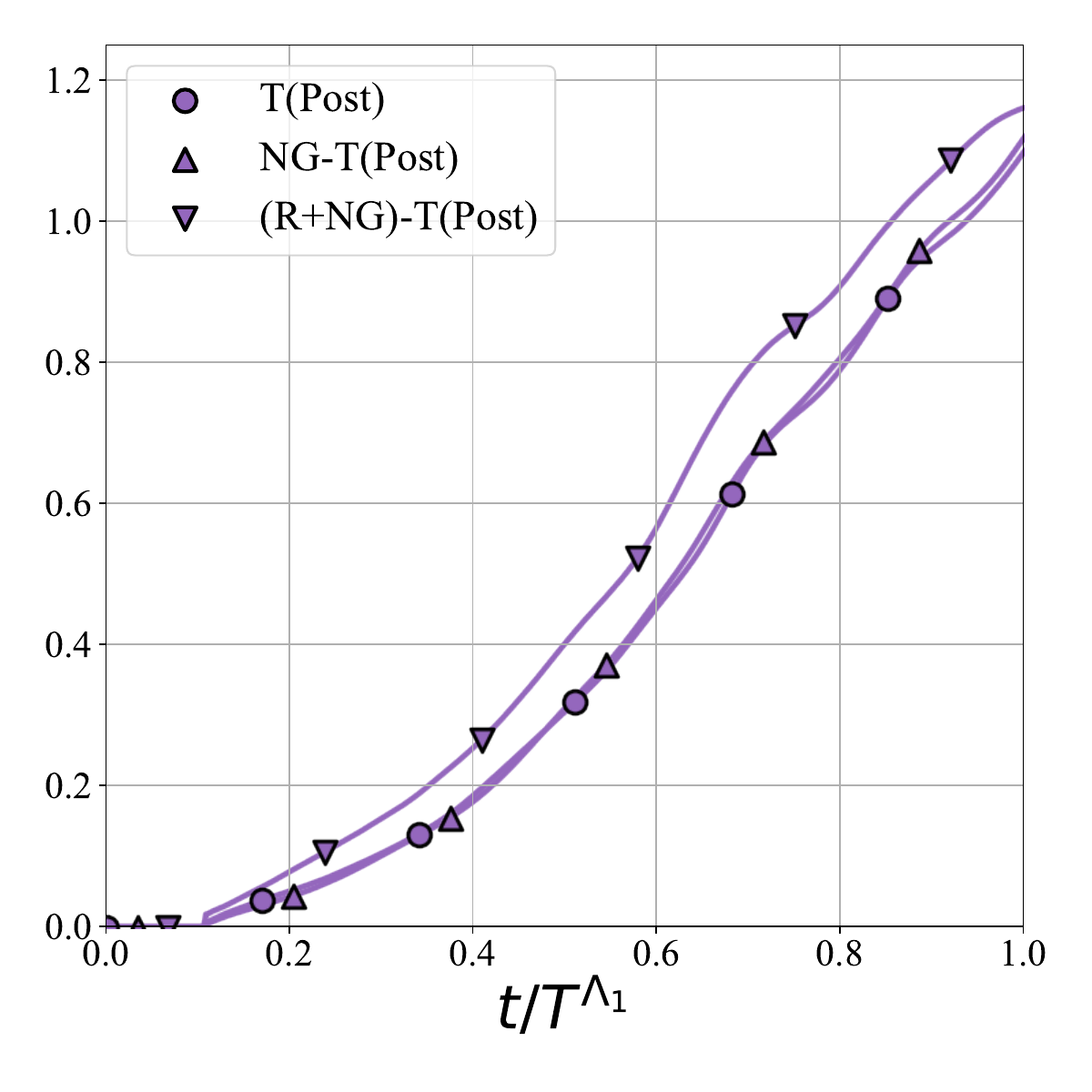}
        \caption{PostLN Transformer}
        \label{fig:l96_nrmse_post}
    \end{subfigure}%
    
    \begin{subfigure}{0.33\textwidth}
        \centering
        \includegraphics[width=\linewidth]{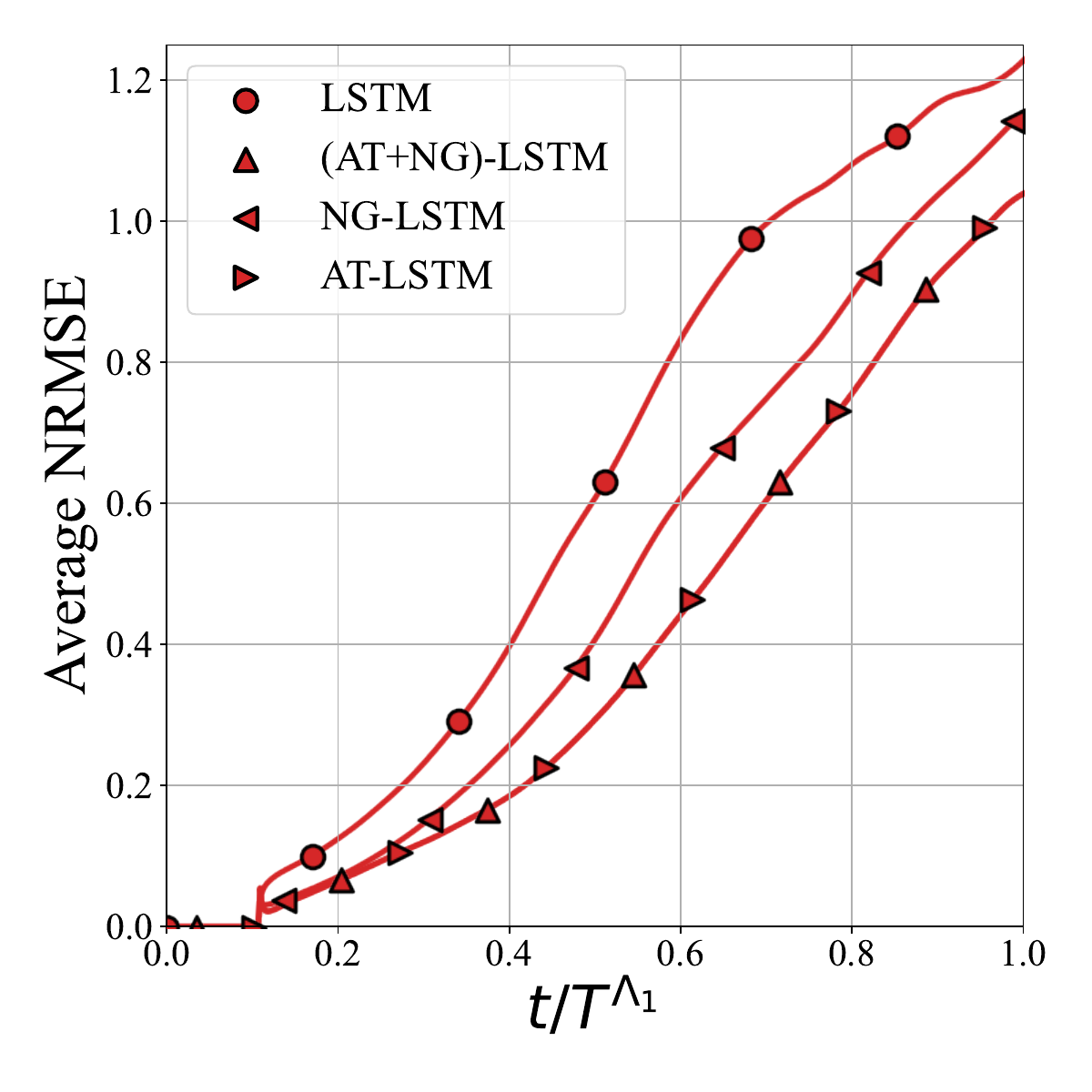}
        \caption{LSTM}
        \label{fig:l96_nrmse_lstm}
    \end{subfigure}%
    \begin{subfigure}{0.33\textwidth}
        \centering
        \includegraphics[width=\linewidth]{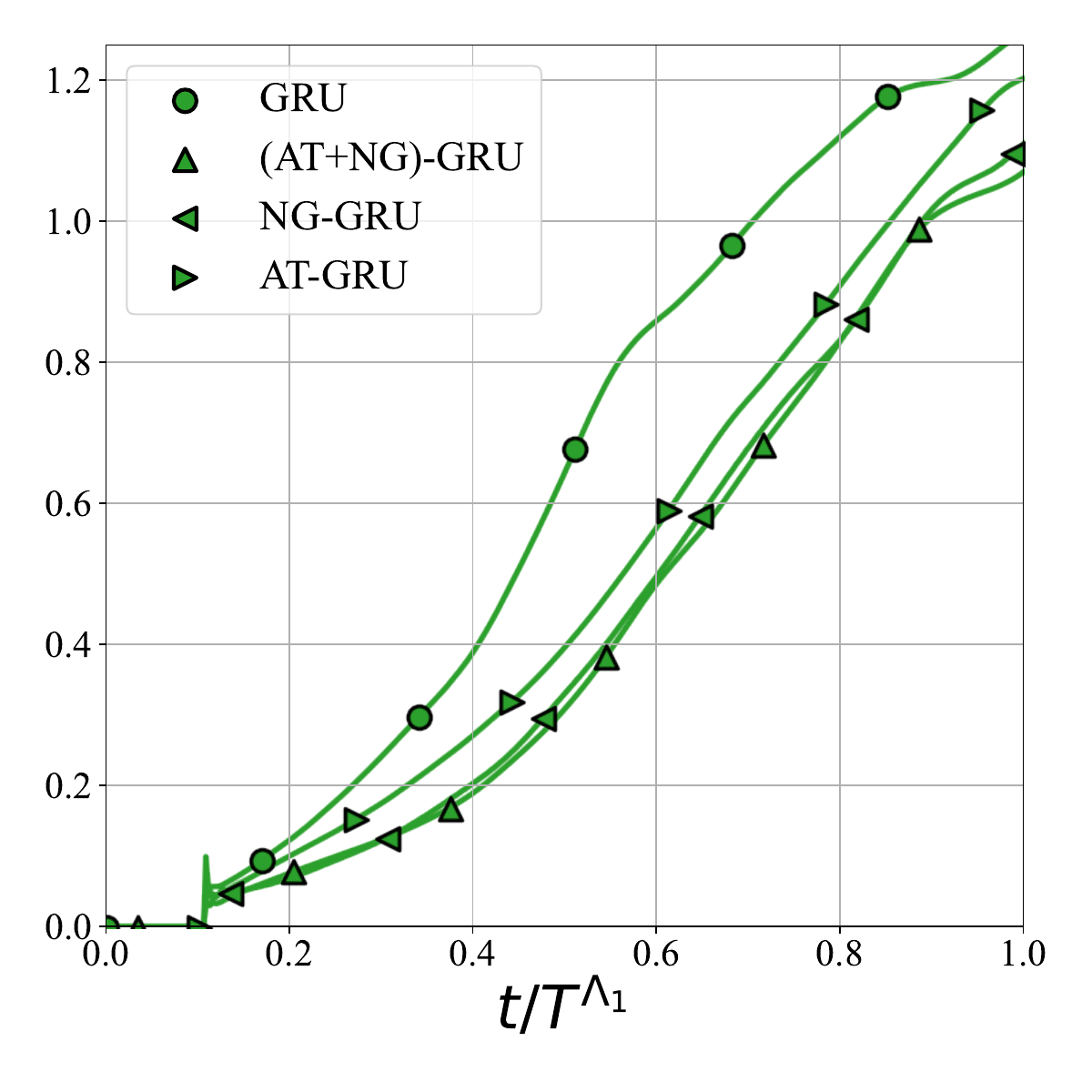}
        \caption{GRU}
        \label{fig:l96_nrmse_gru}
    \end{subfigure}%
    \begin{subfigure}{0.33\textwidth}
        \centering
        \includegraphics[width=\linewidth]{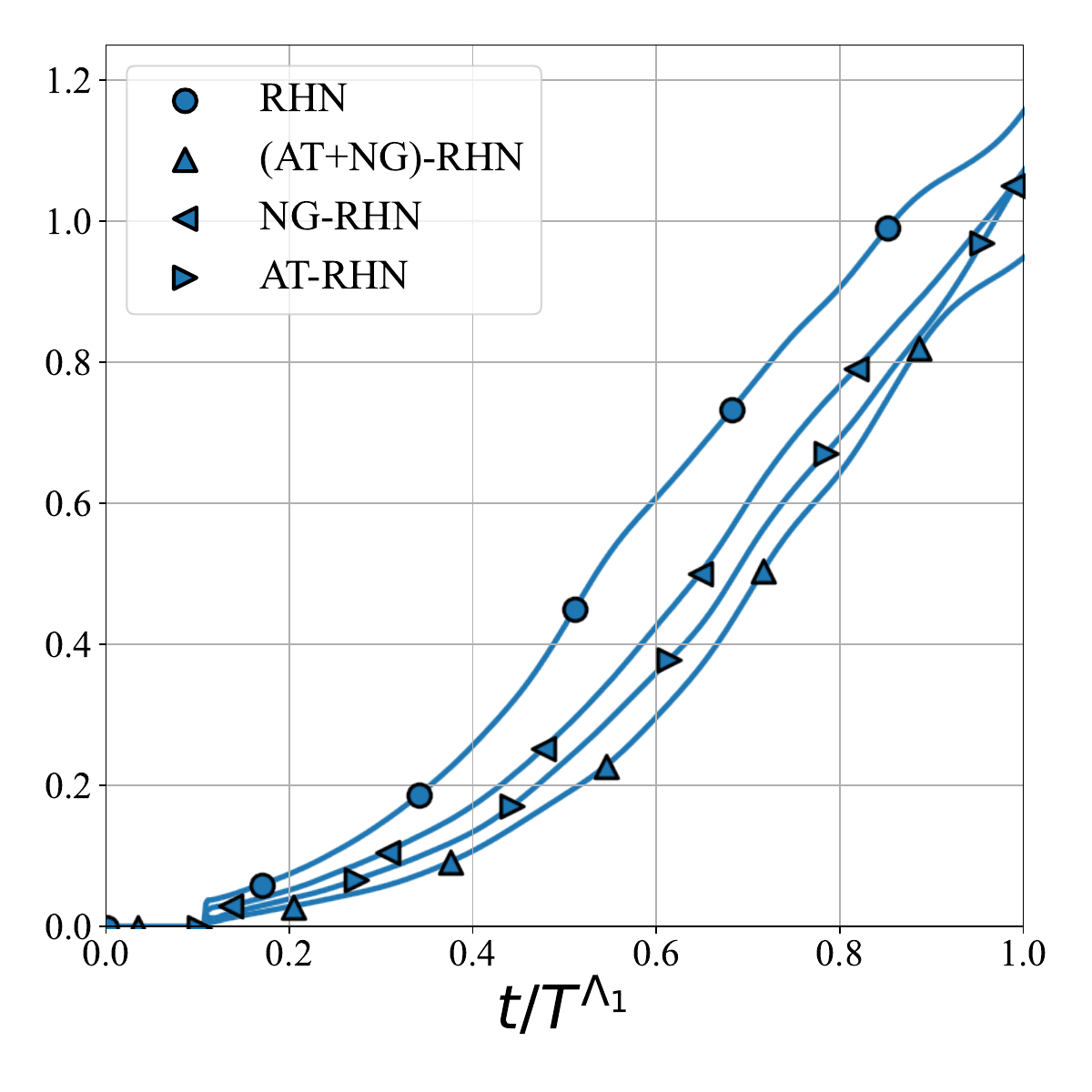}
        \caption{RHN}
        \label{fig:l96_nrmse_rhn}
    \end{subfigure}%
    \caption{
        The evolution of average NRMSE error for top models of each architecture on Multiscale Lorenz-96 with $F = 10$. 
        NRMSE is averaged with respect to the initial conditions in the test split.
        Though many models are able to achieve comparable performance, the RHN exhibits the least error. 
    }
    \label{fig:ml96_f10_nrmse}
\end{figure*}

\begin{figure}[htb!]
    \centering
    Multiscale Lorenz-96 (F=20)\\
    \includegraphics[width=0.5\textwidth]{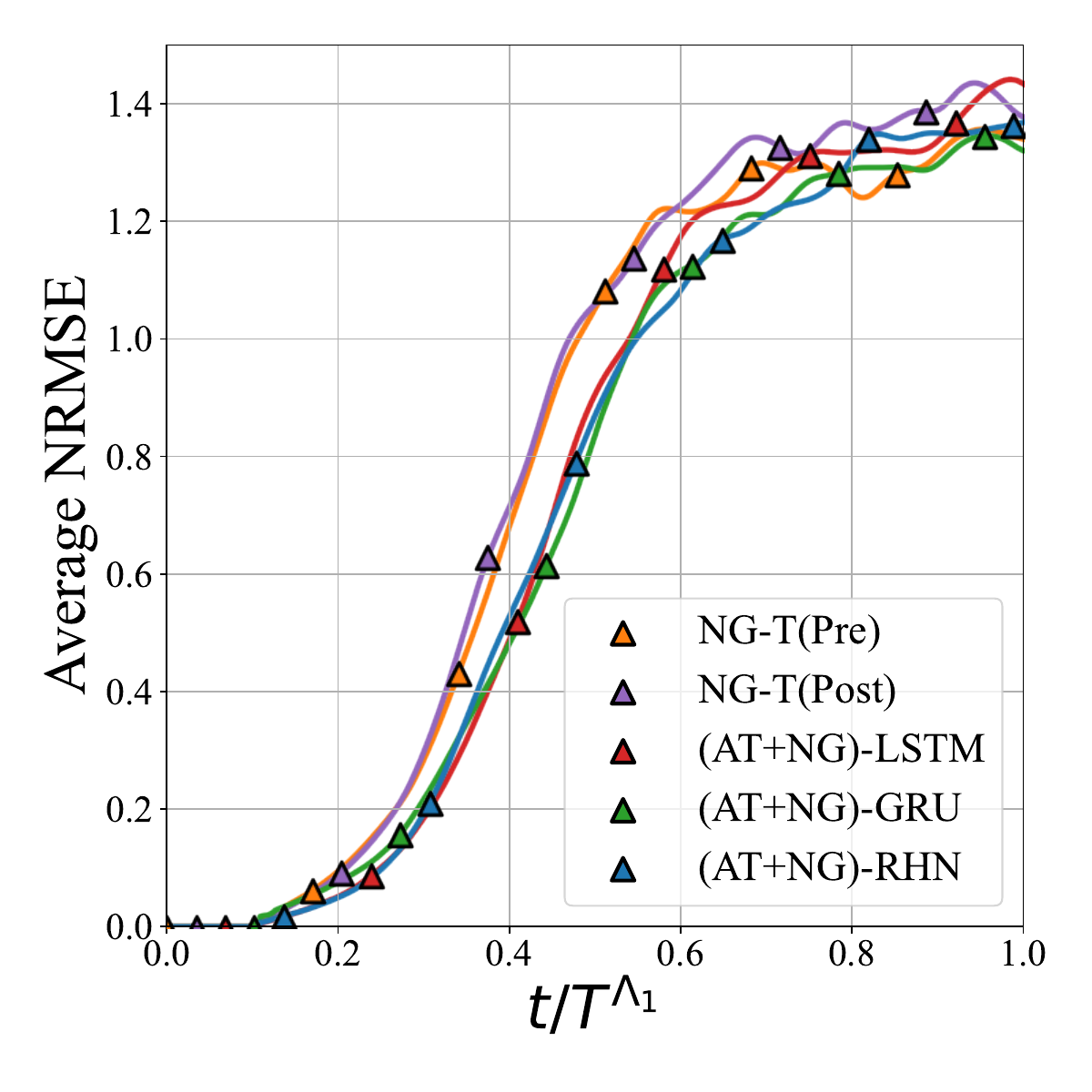}
    \caption{
        Evolution of the average NRMSE error for top models of each architecture on Multiscale Lorenz-96 with $F = 20$. 
        For the more difficult $F=20$ case, the RNNs appear to perform slightly better than the Transformer models.
    }
    \label{fig:ml96_f20_nrmse}
\end{figure}

\begin{figure*}[htb!]
    \centering
    \includegraphics[width=\textwidth]{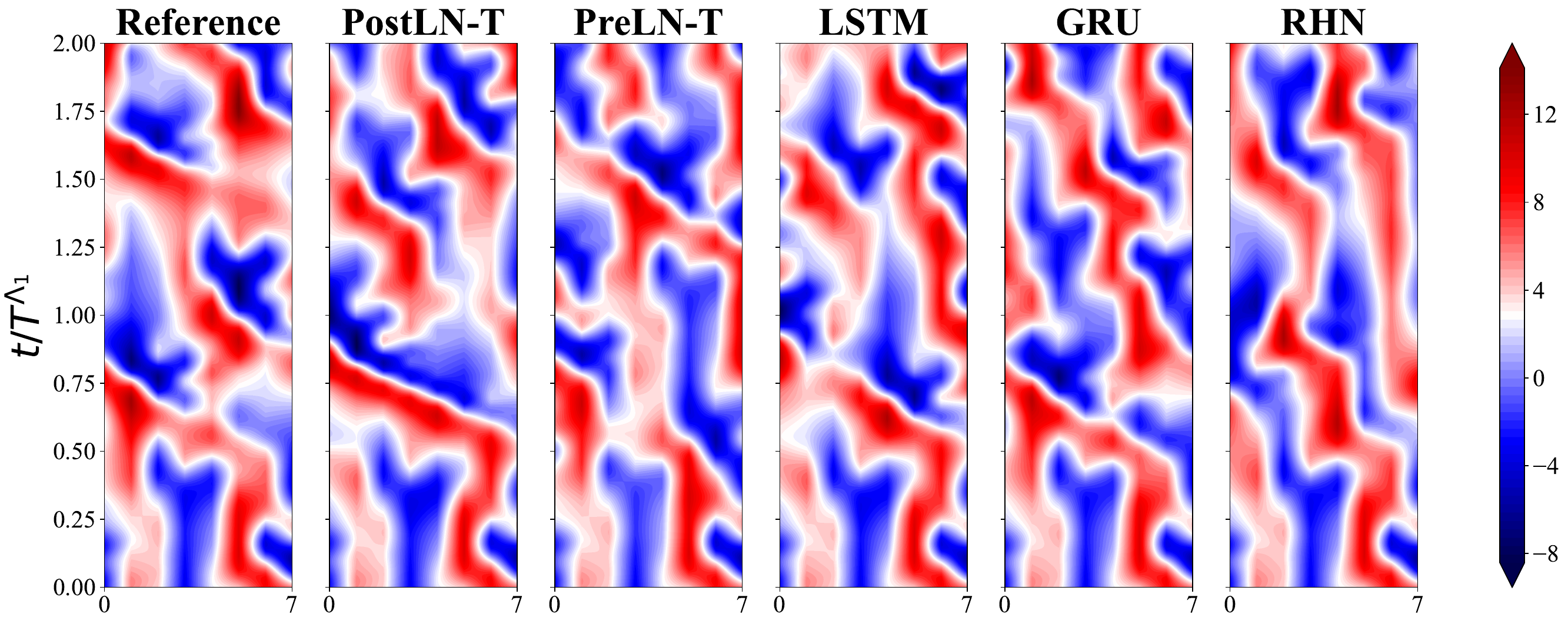}
    \includegraphics[width=\textwidth]{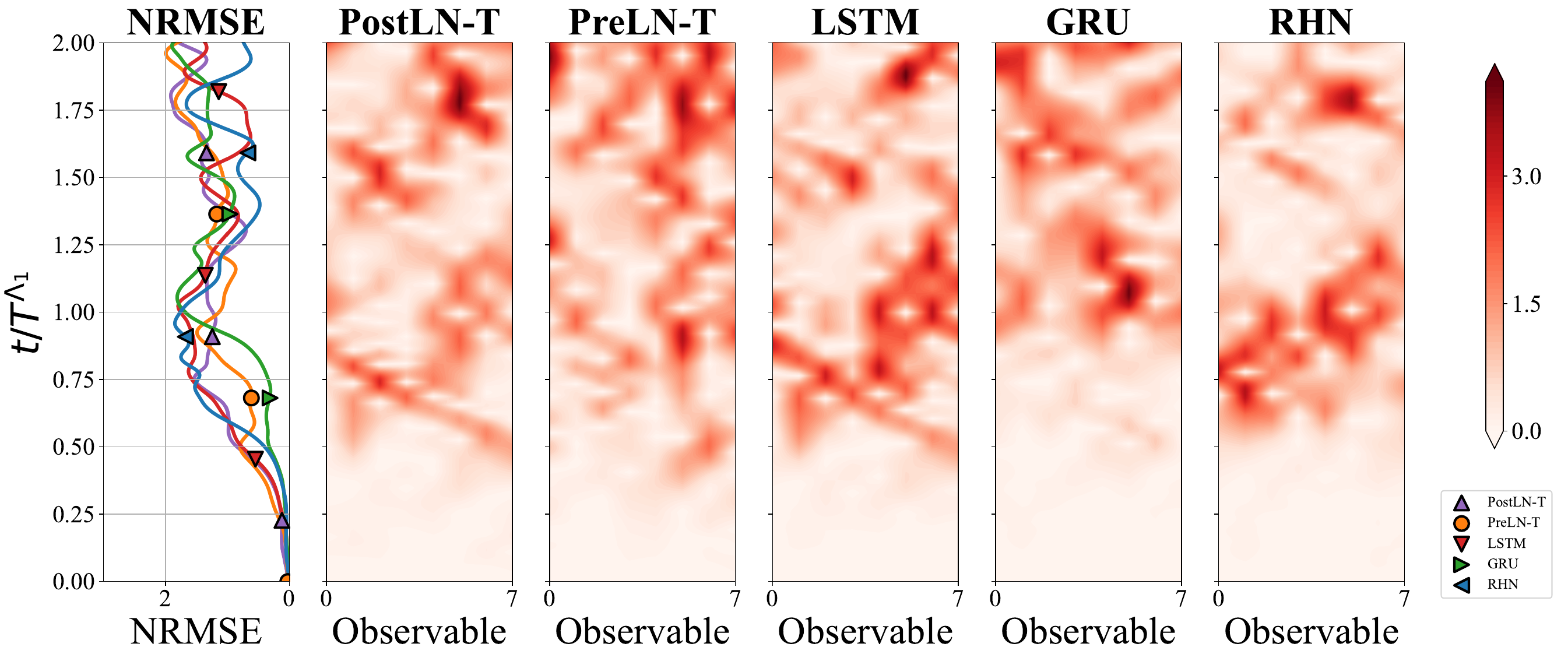}
    \caption{
        Two Lyapunov times of iterative predictions
        for a trajectory sampled randomly from the test split of the dataset for Multiscale Lorenz-96 with $F=10$.
        Errors are plotted in terms of normalized root squared error (NRSE).
    }
    \label{fig:ml96_f10_qual}
\end{figure*}

\begin{figure*}[htb!]
    \centering
    \includegraphics[width=\textwidth]{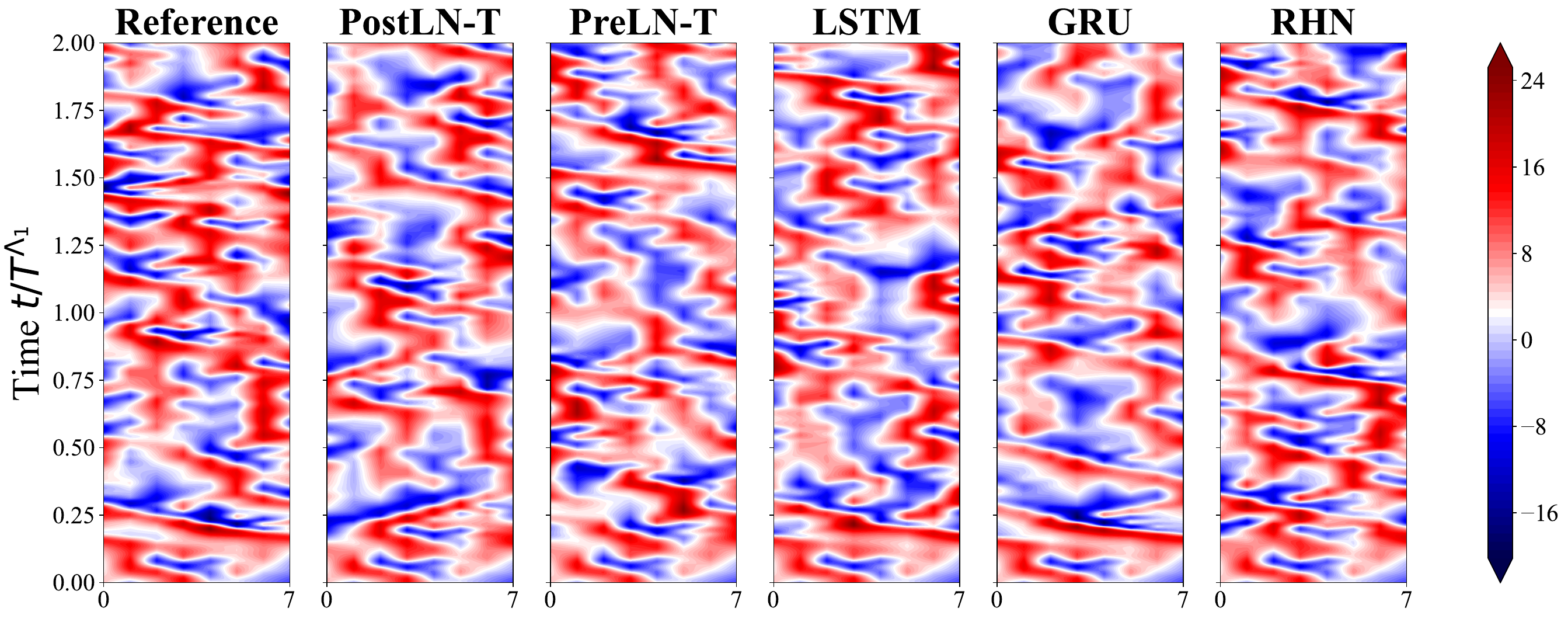}
    \includegraphics[width=\textwidth]{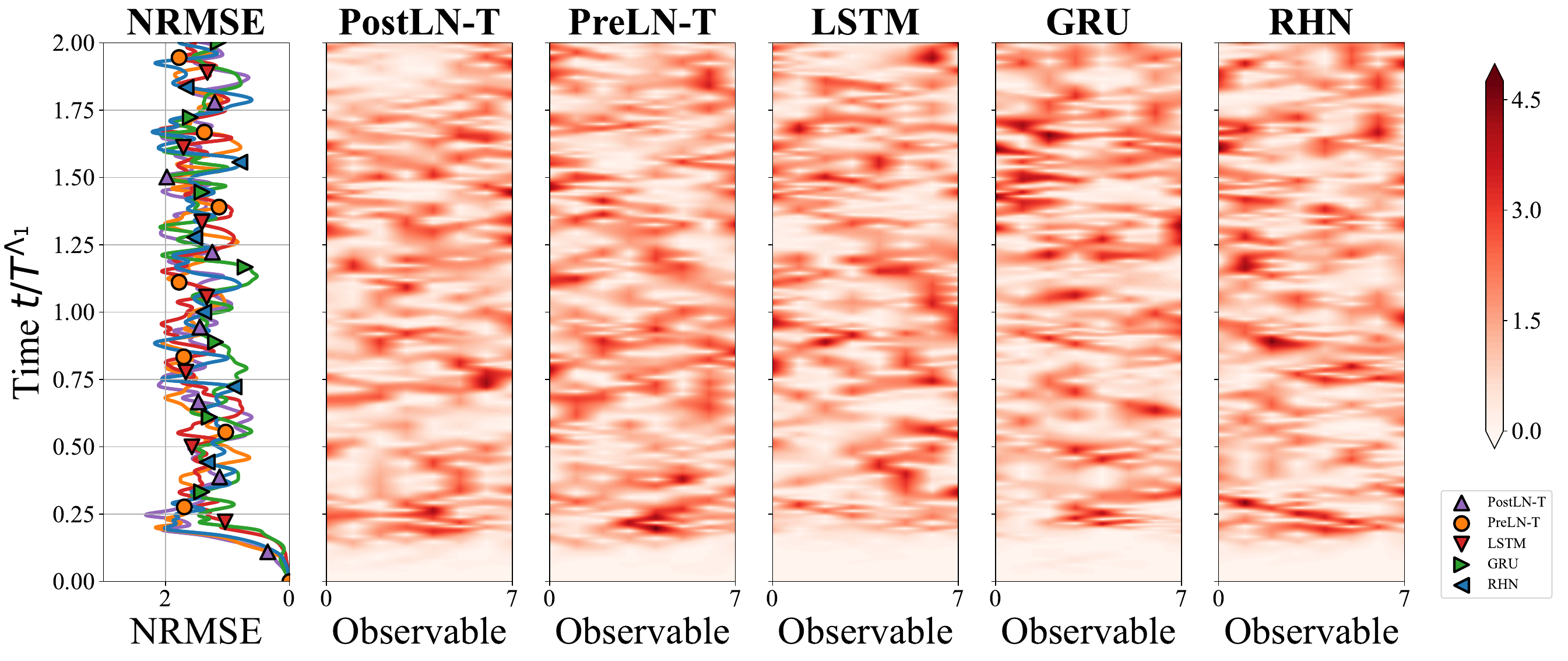}
    \caption{
        Two Lyapunov times of iterative predictions
        for a trajectory sampled randomly from the test split of the dataset for Multiscale Lorenz-96 with $F=20$.
        Errors are plotted in terms of normalized root squared error (NRSE).
    }
    \label{fig:ml96_f20_qual}
\end{figure*}

For each architectural class, we identify the top-performing model based on the highest validation VPT (averaged over 100 initial conditions) and plot the evolution of its average test NRMSE in~\Cref{fig:ml96_f10_nrmse,fig:ml96_f20_nrmse}.
\Cref{fig:ml96_f10_nrmse} exhibits the best model within each architecture and for each architectural variation, for the lower forcing regime $F=10$.
Considering~\Cref{fig:l96_nrmse_pre,fig:l96_nrmse_rhn}, we observe that all standard architectures, without the modifications considered in this work, show the worst forecasting performance and accumulate error most rapidly.
The best-performing networks of each class are hybrids, e.g., RNNs equipped with attention.  
The best model is the RHN, almost reaching a full Lyapunov time before exceeding $\epsilon=1$ as shown in~\Cref{fig:l96_nrmse_all,fig:l96_nrmse_rhn}. 
\Cref{fig:ml96_f20_nrmse} shows top-performing models on $F=20$, where we observe similar performance between RNN cells and Transformer normalization types (pre and post), with RNNs outperforming Transformers in this more chaotic regime. 
As illustrative examples, we plot iterative predictions for two full Lyapunov times for random test samples in~\Cref{fig:ml96_f10_qual,fig:ml96_f20_qual}.

\begin{figure*}[htb!]
    \centering
    Multiscale Lorenz-96 (F=10)
    \begin{subfigure}{0.33\textwidth}
        \centering
        \includegraphics[width=.99\linewidth]{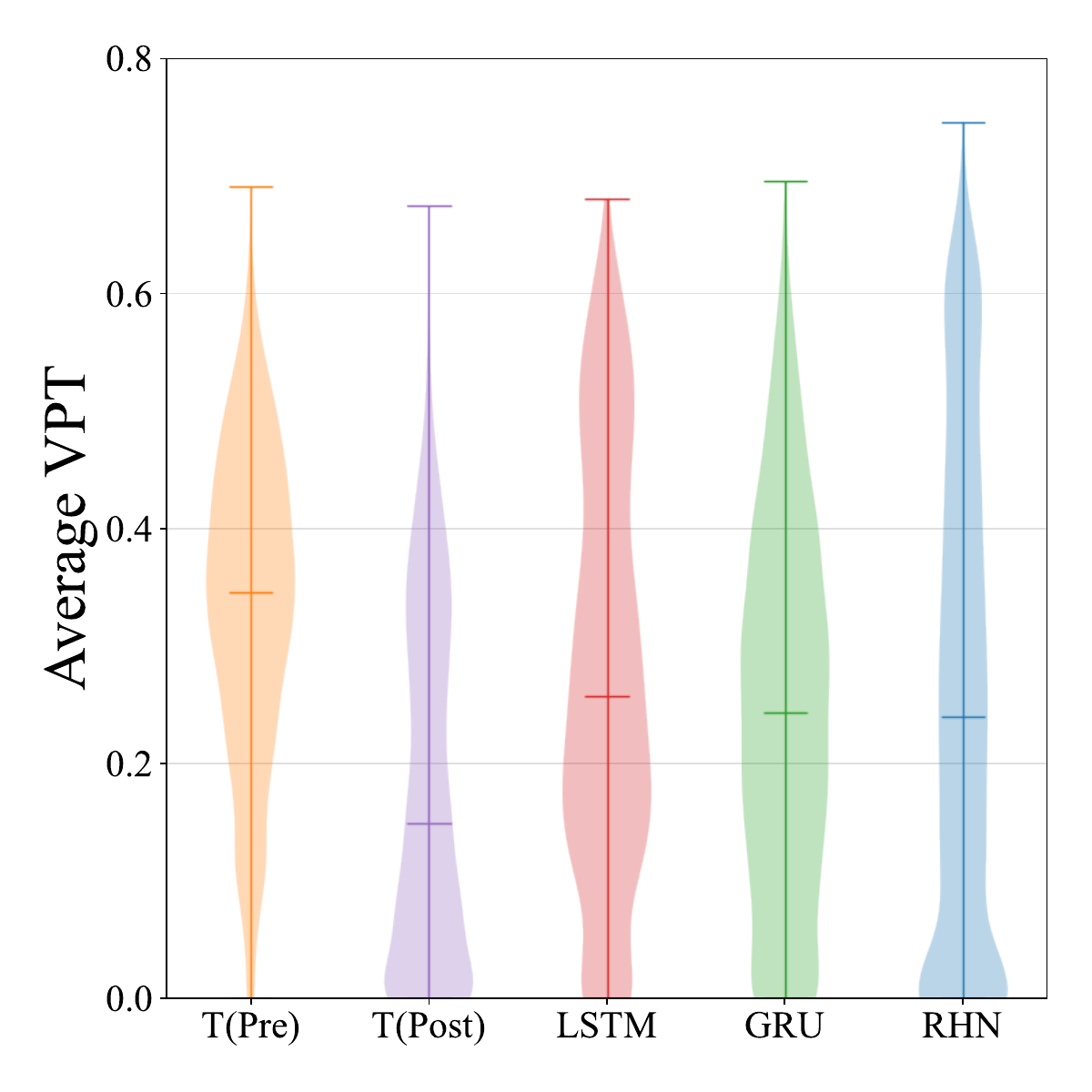}
        \caption{Best Models}
        \label{fig:l96_vpt_all}
    \end{subfigure}%
    \begin{subfigure}{0.33\textwidth}
        \centering
        \includegraphics[width=.99\linewidth]{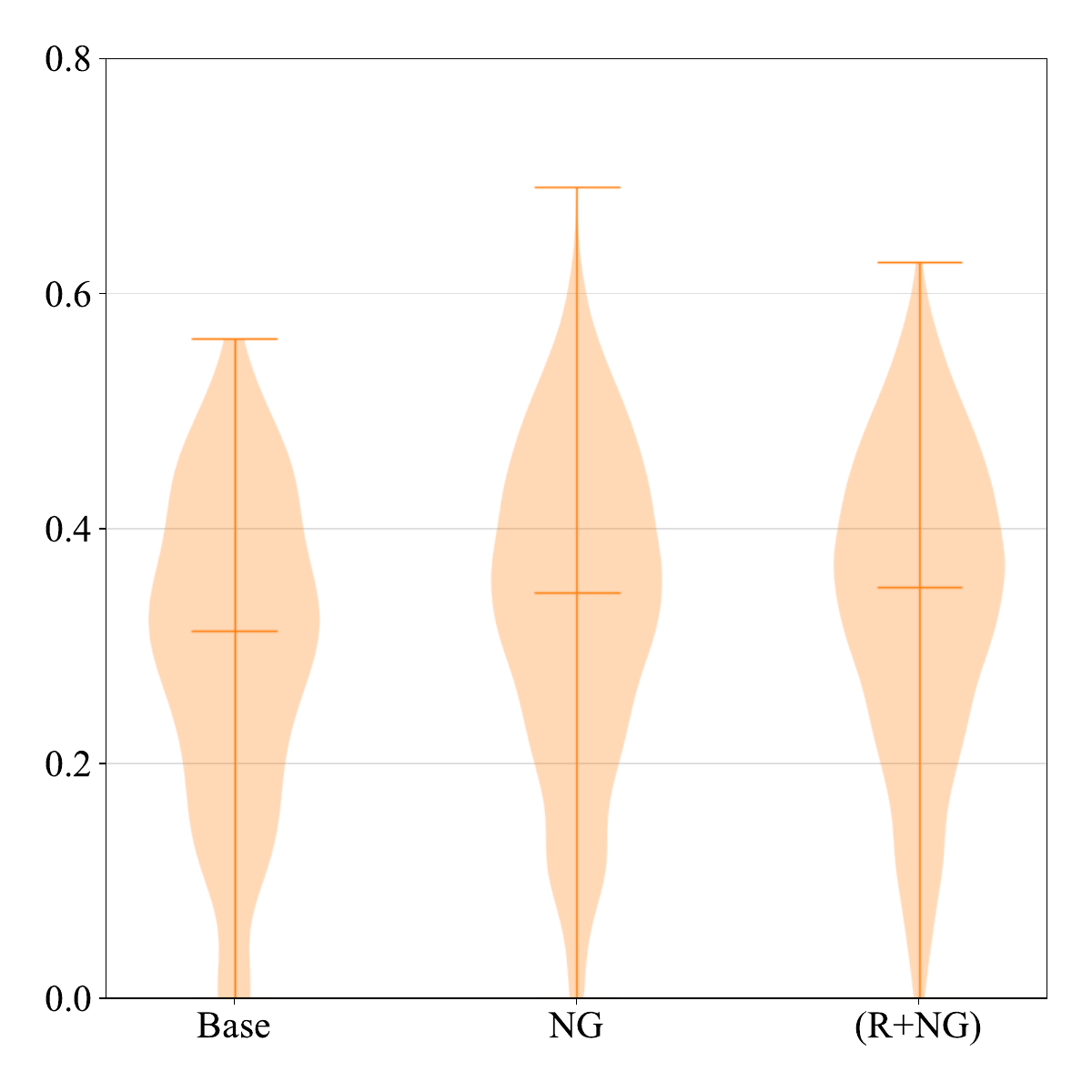}
        \caption{PreLN Transformer}
        \label{fig:l96_vpt_pre}
    \end{subfigure}%
    \begin{subfigure}{0.33\textwidth}
        \centering
        \includegraphics[width=.99\linewidth]{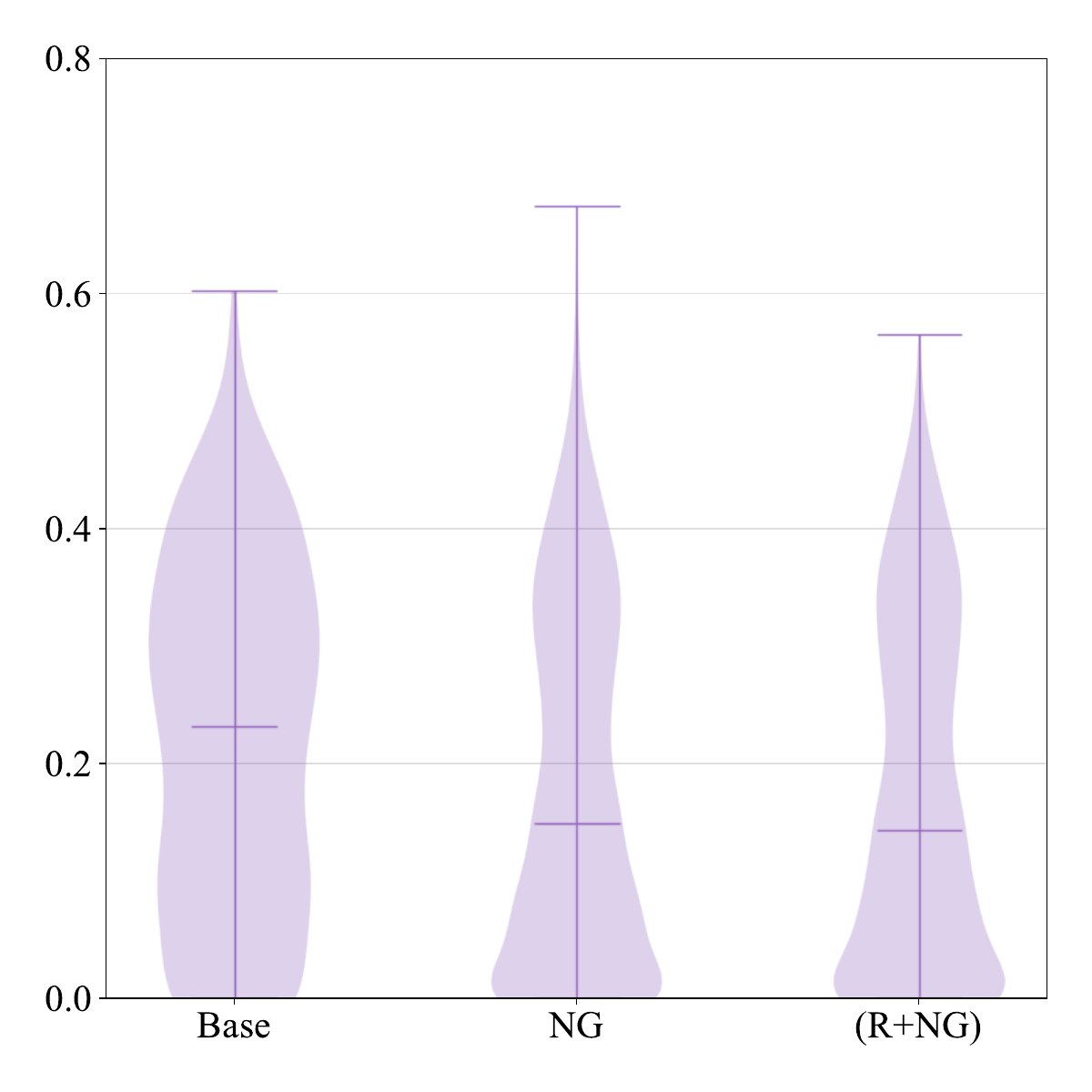}
        \caption{PostLN Transformer}
        \label{fig:l96_vpt_post}
    \end{subfigure}%
    
    \begin{subfigure}{0.33\textwidth}
        \centering
        \includegraphics[width=.99\linewidth]{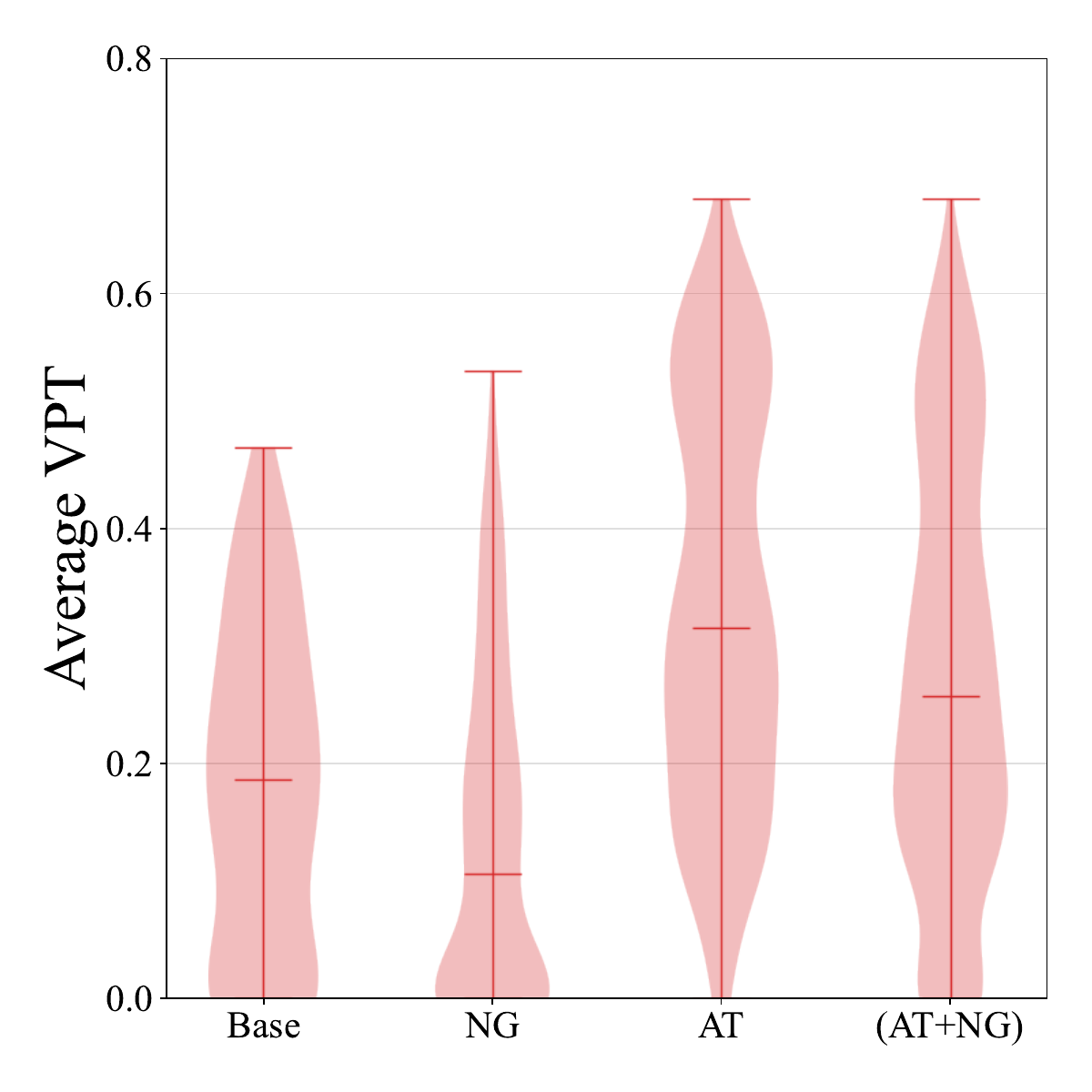}
        \caption{LSTM}
        \label{fig:l96_vpt_lstm}
    \end{subfigure}%
    \begin{subfigure}{0.33\textwidth}
        \centering
        \includegraphics[width=.99\linewidth]{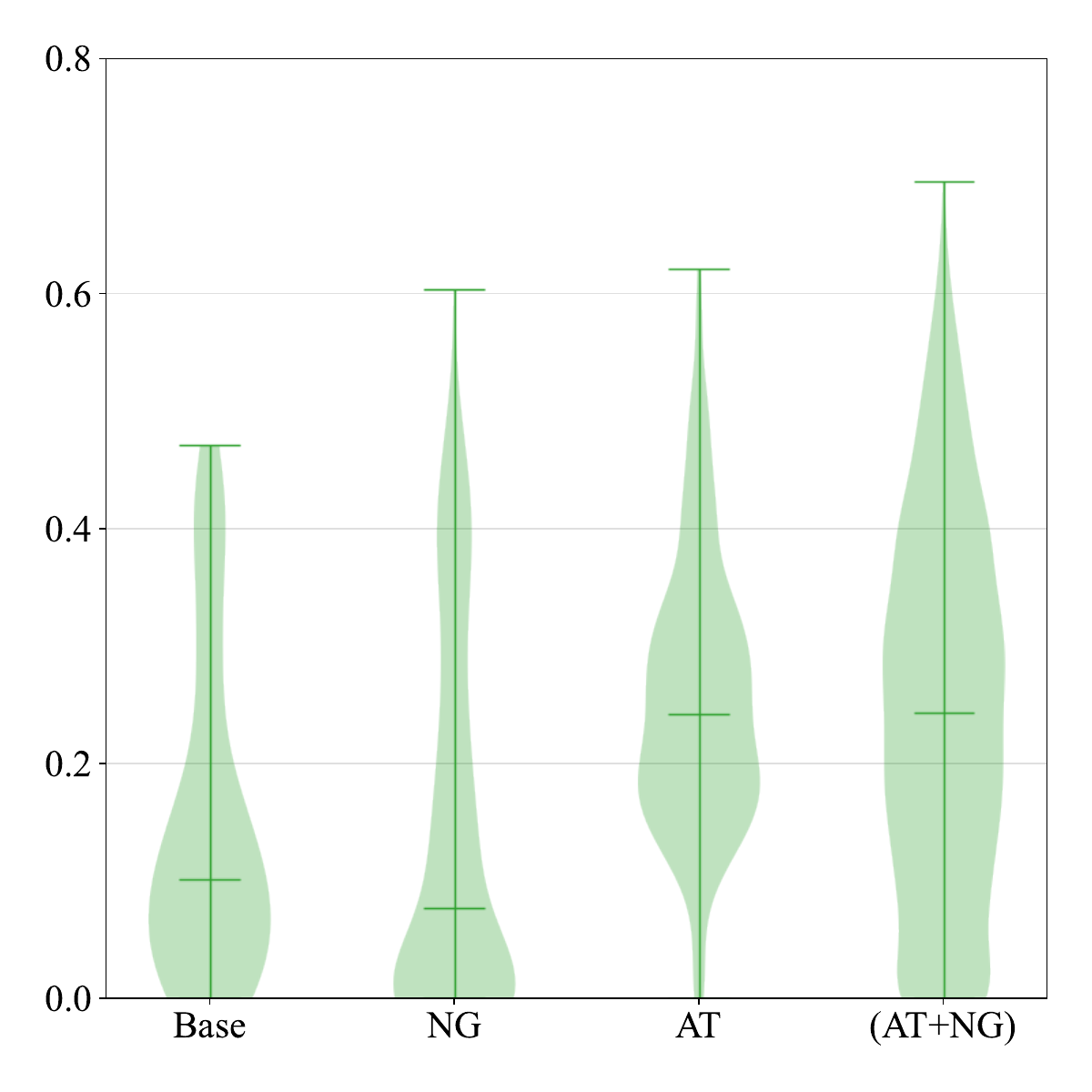}
        \caption{GRU}
        \label{fig:l96_vpt_gru}
    \end{subfigure}%
    \begin{subfigure}{0.33\textwidth}
        \centering
        \includegraphics[width=.99\linewidth]{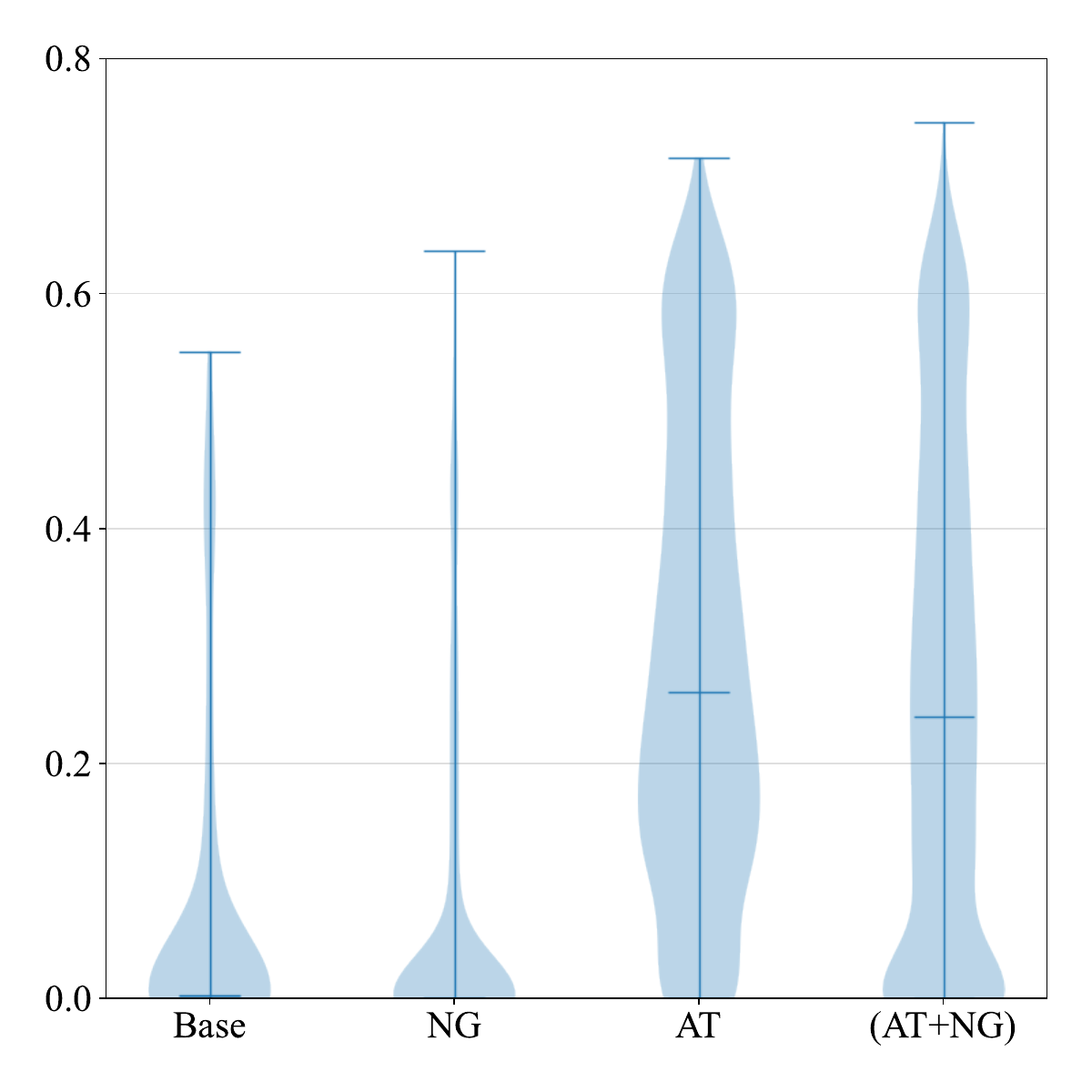}
        \caption{RHN}
        \label{fig:l96_vpt_rhn}
    \end{subfigure}%
    \caption{
        Violin plots showing the smoothed kernel density estimate of average VPT performance over all models trained in this study for forcing regimes $F=10$ in Multiscale Lorenz-96.
        The upper-left plot displays the distribution of NG Transformers and the (AT+NG) RNNs for cross-architectural comparison.
    }
    \label{fig:ml96_vpt_f10}
\end{figure*}

\begin{figure}[htb!]
    \centering
    Multiscale Lorenz-96 (F=10)

    \includegraphics[width=0.5\textwidth]{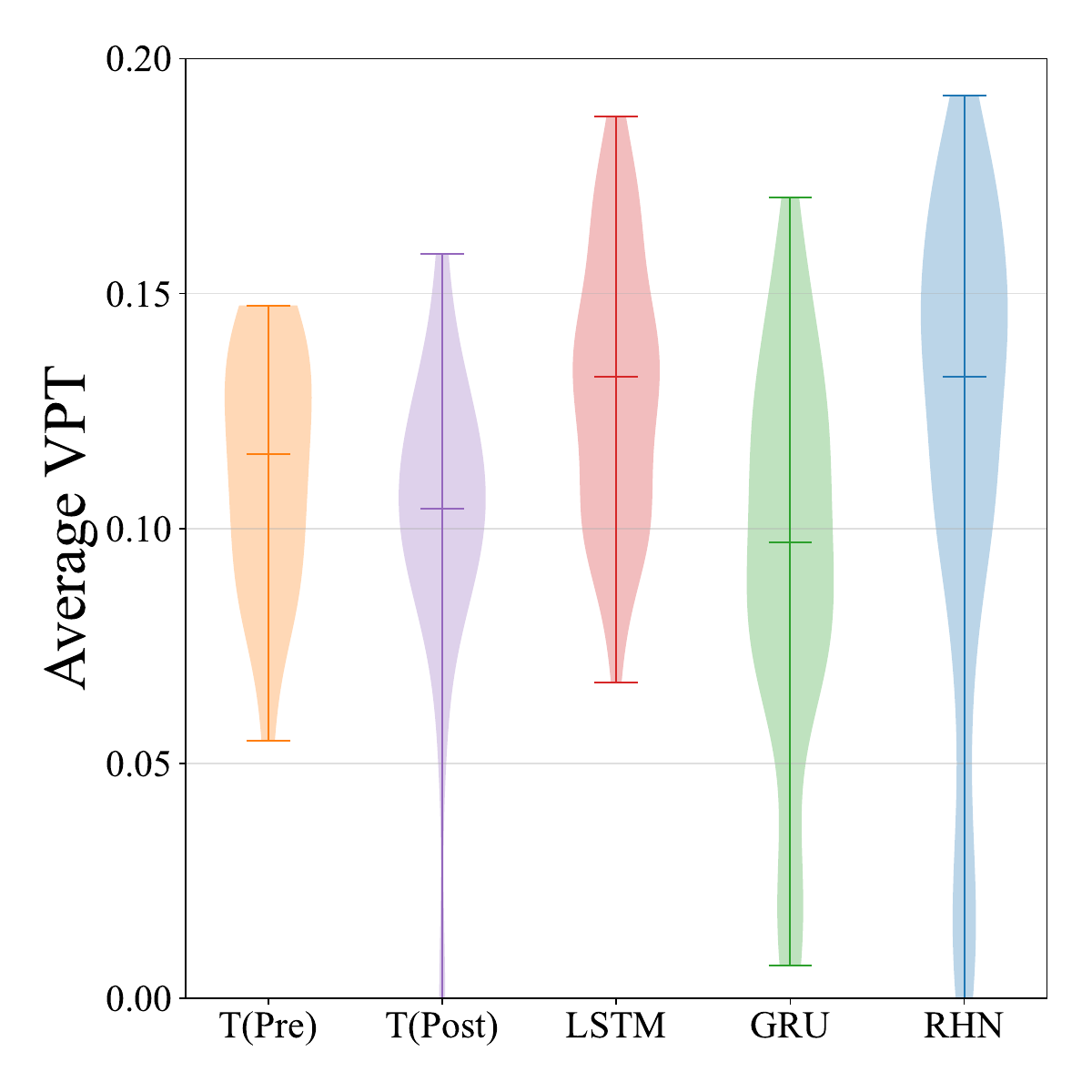}
    \caption{
        Violin plots showing the smoothed kernel density estimate of average VPT performance over all models trained in this study for  $F=20$ in Multiscale Lorenz-96.
    }
    \label{fig:ml96_vpt_f20}
\end{figure}

To illustrate model sensitivity to hyperparameter selection, we provide violin plots of the smoothed kernel density estimate of all tested models in~\Cref{fig:ml96_vpt_f10,fig:ml96_vpt_f20}.
Horizontal lines denote the median and extrema of each empirical distribution. 
Note that VPT plots for $F=20$ contain fewer models than for $F=10$ due to the reduced search space. 
The lower end of distributions corresponds to models that fail to learn the system and achieve a test VPT of 0, exceeding $\epsilon=0.5$ at the first predictive step.
The Transformer(Pre) and Transformer(Post) plots in~\Cref{fig:l96_vpt_pre,fig:l96_vpt_post,fig:ml96_vpt_f20} demonstrate that the PreLN variant is more robust with respect to NG and R+NG than the PostLN variant, as indicated by the higher mean VPTs for PreLN variants and the significant amount of PostLN variants achieving a VPT of 0.
These observations agree with past studies in NLP that indicate PostLN Transformers are more susceptible to divergence during training~\cite{wang2022deepnet,wang2022foundation}. 
In RNNs, we find that all baseline models obtain a high incidence rate of failing models as shown in the left-most violin in~\Cref{fig:l96_vpt_lstm,fig:l96_vpt_rhn}.
While gating can help increase performance, it is highly dependent on the type of gate considered. 
Many gated RNNs still fail to learn the system as a result of ill-suited gating mechanisms, such as the additive or learned rate gate, which is consistent with past RNN studies that argue gating is critical for stability in recurrent cells~\cite{greff2016lstm, tallec2018can}.
The most significant amelioration of these RNN failures occurs when introducing attention, as observed in the significant shifting of probability mass from 0 VPT to above 0.2 VPT in both the AT and AT+NG variants. 

\begin{table}[htb!]
    \centering
    \begin{tabular}{|l|l|l|}
    \hline
    \diagbox{Model}{Scenario} & $F = 10$ & $F = 20$\\
    \hline
    \hline
    LSTM & 0.44 & 0.12 \\
    AT-LSTM & 0.64 (45.9\%) & 0.17 (39.2\%) \\
    NG-LSTM & 0.53 (18.9\%) & 0.12 (0.0\%) \\
    (AT+NG)-LSTM & 0.64 (45.9\%) & 0.17 (39.2\%) \\
    \hline 
    GRU & 0.44 & 0.11 \\
    AT-GRU & 0.59 (33.4\%) & 0.16 (44.2\%) \\
    NG-GRU & 0.60 (36.7\%) & 0.13 (14.9\%) \\
    (AT+NG)-GRU & 0.63 (43.0\%) & 0.17 (53.9\%)  \\
    \hline 
    RHN & 0.51 & 0.15 \\
    AT-RHN & 0.65 (28.4\%) & \textbf{0.19} (24.9\%) \\
    NG-RHN & 0.64 (25.7\%) & 0.15 (0.0\%)  \\
    (AT+NG)-RHN & \textbf{0.73} (44.4\%)  & 0.17 (15.2\%)\\
    \hline 
    Transformer(Post) & 0.60 & 0.13 \\
    NG-Transformer(Post) & 0.61 (1.8\%)  & 0.14 (10.1\%) \\
    (R+NG)-Transformer(Post) & 0.54 (-9.7\%) & - \\
    \hline 
    Transformer(Pre) & 0.56 & 0.11 \\
    NG-Transformer(Pre) & 0.66 (17.0\%) & 0.15 (29.3\%) \\
    (R+NG)-Transformer(Pre) & 0.57 (1.5\%) & - \\
    \hline
    \end{tabular}
    \caption{
        Average test VPT over 100 initial conditions of top models, stratified by model class and mechanism.
        The best in each column is highlighted in bold. 
        Modified architectures are denoted with prefixes: Neural Gating (NG), Attention (AT), and Recurrence (R). 
        Relative performance changes with respect to baseline models are displayed in parentheses.
    }
    \label{tab:ml96}
\end{table}

We display quantitative results for top-performing models in~\Cref{tab:ml96}. 
To properly ascribe performance gains to the neural mechanisms studied here, we include separate rows in~\Cref{tab:ml96} highlighting neural gating (NG), attention (AT), and recurrence (R), when considered. 
As shown in~\Cref{tab:ml96}, we find that neural gating and neural attention tend to increase performance in RNNs over baseline variants, and that they complement each other.
For $F=10$, RNNs are improved by more than 40\% over their baseline performance when considering neural attention and gating.
The results depicted in~\Cref{tab:ml96} suggest that neural gating is beneficial for Transformers, adding more evidence in support of the neural attention and gating combination.
In contrast with NG, we find that recurrence is detrimental to Transformer performance and training stability.
Because of this, we do not consider it further in this work. 

\begin{table}[htb!]
    \centering
    \begin{tabular}{|l|cc|}
        \hline
        \textbf{Model} & $F=10$ & $F=20$ \\ 
        \hline
        \hline
        LSTM & 1.13E-03 & 3.56E-03 \\
        NG-LSTM & 1.41E-03 & 1.43E-03 \\
        AT-LSTM & 2.84E-03 & 1.24E-03 \\
        (AT+NG)-LSTM & 1.41E-03 & 1.24E-03 \\
        \hline
        GRU & 3.87E-03 & 4.27E-03 \\
        NG-GRU & \textbf{9.28E-04} & 2.56E-03 \\
        AT-GRU & 1.88E-03 & 2.07E-03 \\
        (AT+NG)-GRU & 1.18E-03 & 2.56E-03 \\
        \hline
        RHN & 1.82E-03 & 1.40E-03 \\
        NG-RHN & 1.30E-03 & 1.73E-03 \\
        AT-RHN & 2.57E-03 & 1.95E-03 \\
        (AT+NG)-RHN & 1.24E-03 & \textbf{9.85E-04} \\
        \hline
        Transformer(Post) & 9.55E-04 & 2.48E-03 \\
        NG-Transformer(Post) & 2.02E-03 & 2.04E-03 \\
        (R+NG)-Transformer(Post) & 2.20E-03 & - \\
        \hline
        Transformer(Pre) & 4.78E-03 & 5.79E-03 \\
        NG-Transformer(Pre) & 3.31E-03 & 2.57E-03 \\
        (R+NG)-Transformer(Pre) & 3.31E-03 & - \\
        \hline
    \end{tabular}
    \caption{  
        MSE of the power spectrum of model predictions versus the ground-truth data for the Multiscale Lorenz-96 system with forcing $F=10$ and $F=20$. The lowest error is highlighted in bold. We observe that recurrent models produce marginally better power spectra than Transformer models.
    }
    \label{tab:ml96_psd}
\end{table} 

\begin{figure}[htb!]
    \centering
    \begin{subfigure}{0.5\textwidth}
        \centering
        \includegraphics[width=.99\linewidth]{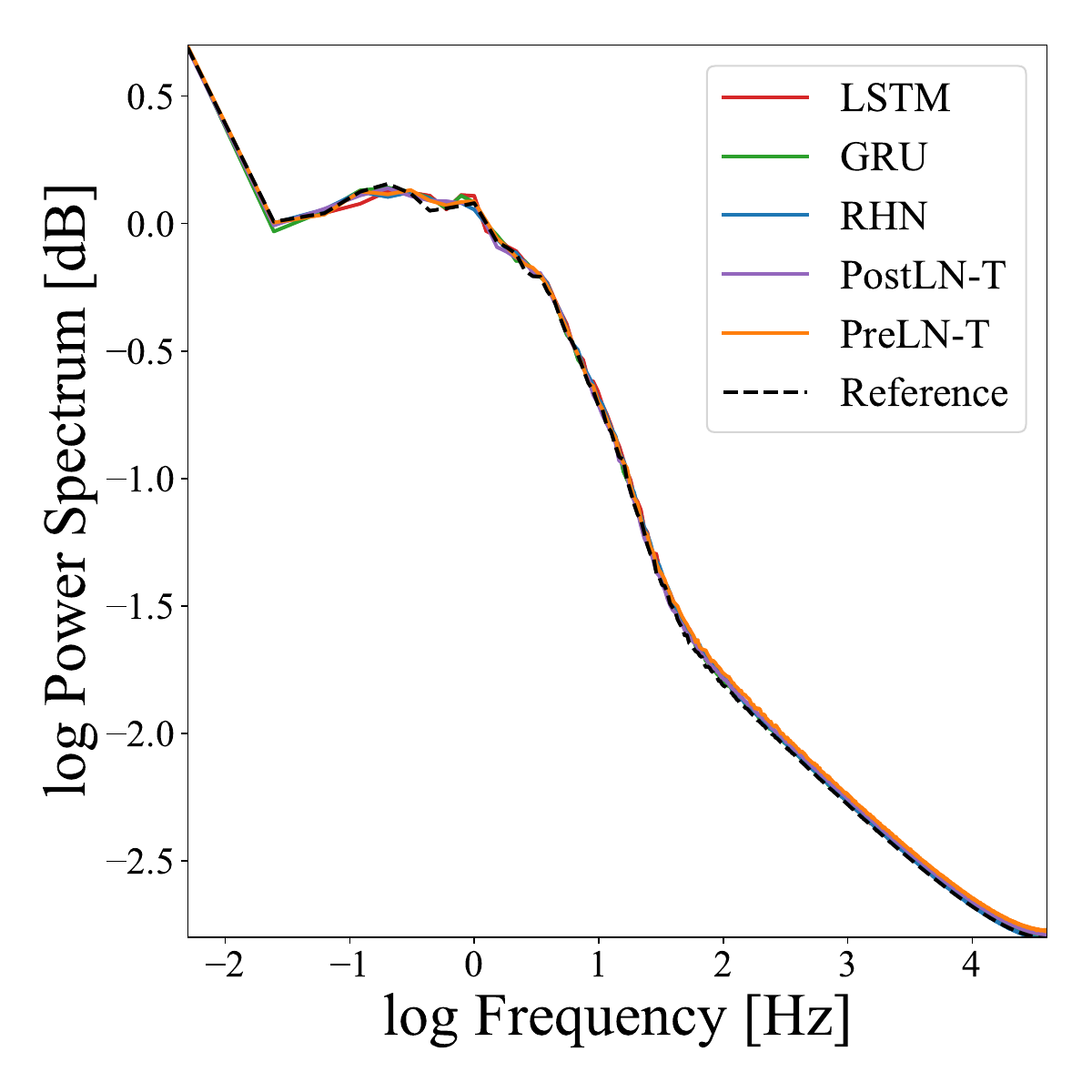}
        \caption{F=10}
        \label{fig:l96f10_psd}
    \end{subfigure}%
    \begin{subfigure}{0.5\textwidth}
        \centering
        \includegraphics[width=.99\linewidth]{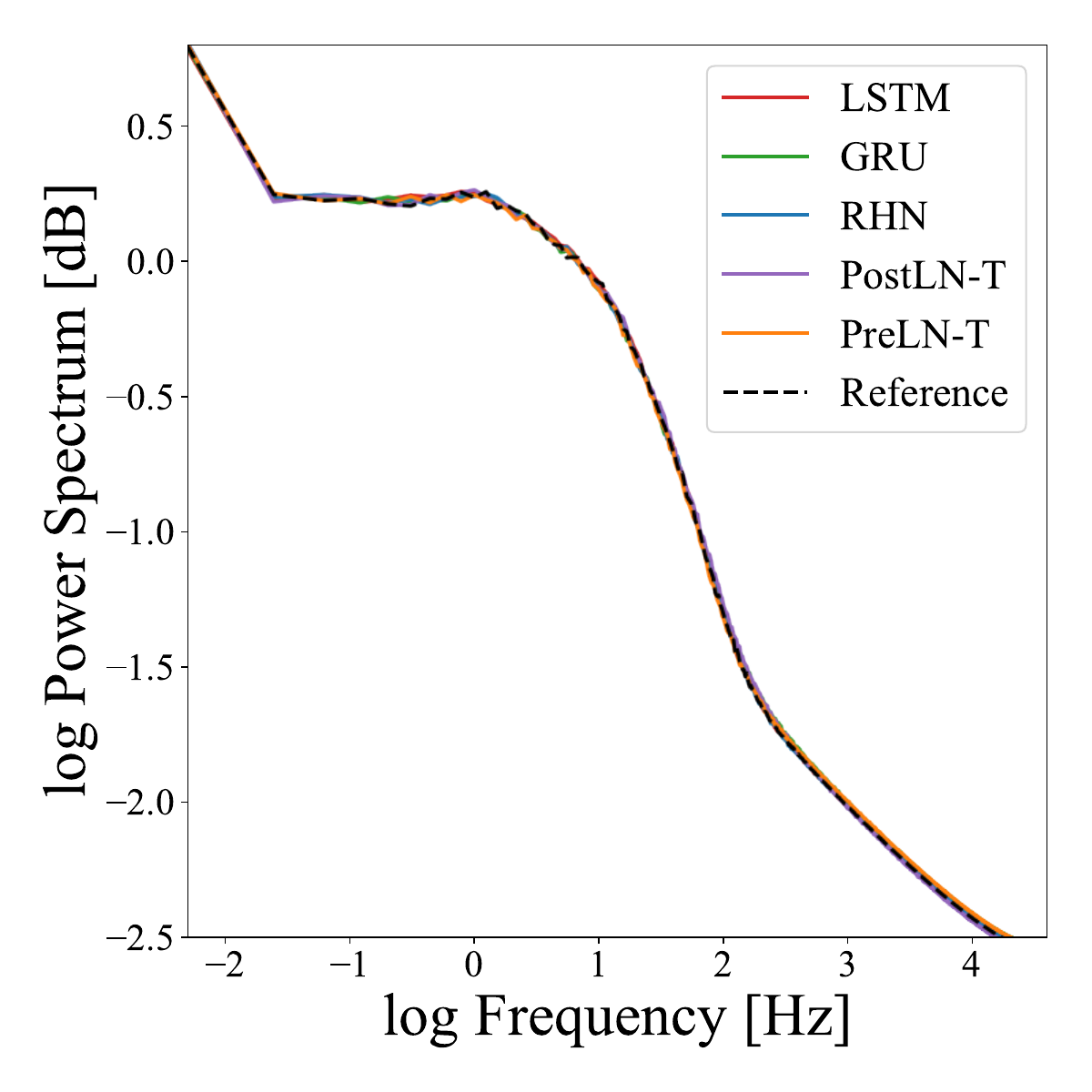}
        \caption{F=20}
        \label{fig:l96f20_psd}
    \end{subfigure}%
    \caption{
        Predicted power spectra for top models selected by lowest deviation from the true power spectra (measured on the validation split).
        We observe that models are able to identify weights that allow faithful reproductions of the power spectra of the true underlying system.
    }
    \label{fig:ml96_psd}
\end{figure}

Beyond the validity of short-term temporal prediction, we assess whether learned models make stable predictions and reproduce the long-term statistics of dynamical evolution.
The top model of each class produces non-divergent predictions, even for forecasts that span 5 Lyapunov times.
These results contrast with previous studies in forecasting chaotic dynamical systems \cite{vlachas2020backpropagation}, where models like Unitary RNNs and Reservoir Computers (RCs) were observed to diverge in settings with partial observability.
To quantify the long-term behavior of models, we compare the power spectra of model predictions and ground-truth testing data in~\Cref{fig:ml96_psd}.
In both forcing regimes and for all models studied, we observe high agreement between power spectra.
Again, this indicates that learned models are stable and faithfully capture long-term aspects of the Lorenz-96 system.
\Cref{tab:ml96_psd} compares the mean square error (MSE) of predicted spectra and shows that RNNs reproduce long-term statistics slightly better.

\begin{table}[htb!]
    \centering
    \begin{tabular}{|l|r|}
        \hline 
        \textbf{Model} & \textbf{Training Time [s]} \\
        \hline 
        \hline
        LSTM & 2838$\pm$1855 \\
        NG-LSTM & 2931$\pm$1546 \\
        AT-LSTM & 7541$\pm$2584 \\
        (AT+NG)-LSTM & 7901$\pm$2374 \\
        \hline
        GRU & 13130$\pm$5454 \\
        NG-GRU & 5590$\pm$5679 \\
        AT-GRU & 8614$\pm$2484 \\
        (AT+NG)-GRU & 8347$\pm$2274 \\
        \hline
        RHN & 3922$\pm$2799 \\
        NG-RHN & 3940$\pm$1710 \\
        AT-RHN & 10001$\pm$2794 \\
        (AT+NG)-RHN & 9554$\pm$3097 \\
        \hline
        Transformer(Post) & 2045$\pm$2413 \\
        NG-Transformer(Post) & 2917$\pm$2655 \\
        (R+NG)-Transformer(Post) & 2306$\pm$1556 \\
        \hline
        Transformer(Pre) & 1663$\pm$755 \\
        NG-Transformer(Pre) & 1906$\pm$1114 \\
        (R+NG)-Transformer(Pre) & 2423$\pm$1196 \\
        \hline
    \end{tabular}
    \caption{
        Mean $\pm$ standard deviation of the training times of the top 100 models for each model class.
        Training times are just for the Multiscale Lorenz-96 data with $F=10$.
        Top models are selected with respect to average VPT performance on the validation split of the dataset.
    }
    \label{tab:ml96_f10_train}
\end{table}

Next, we report the training times of each model class in~\Cref{tab:ml96_f10_train}.
We select the top 100 models for $F=10$ by their average VPT and display the average and standard deviation of their train times.
On average, the various mechanisms increase train time for every model class except the GRU.
As expected, the training times for Transformers are significantly less than RNNs.
We observe that the training time of Transformers is only 25-50\% of RNNs, even after modifying their architectures and increasing their parameter counts.
Thus, we find that although Transformers may not outperform RNNs here, they offer competitive predictive performance for a lower training budget.
Furthermore, the reduced training budget offers the additional benefit of cheaper model evaluation, further increasing the number of experiments one may consider.

\begin{figure*}[htb!]
    \centering

     \begin{subfigure}{0.5\textwidth}
        \centering
        \includegraphics[width=.99\linewidth]{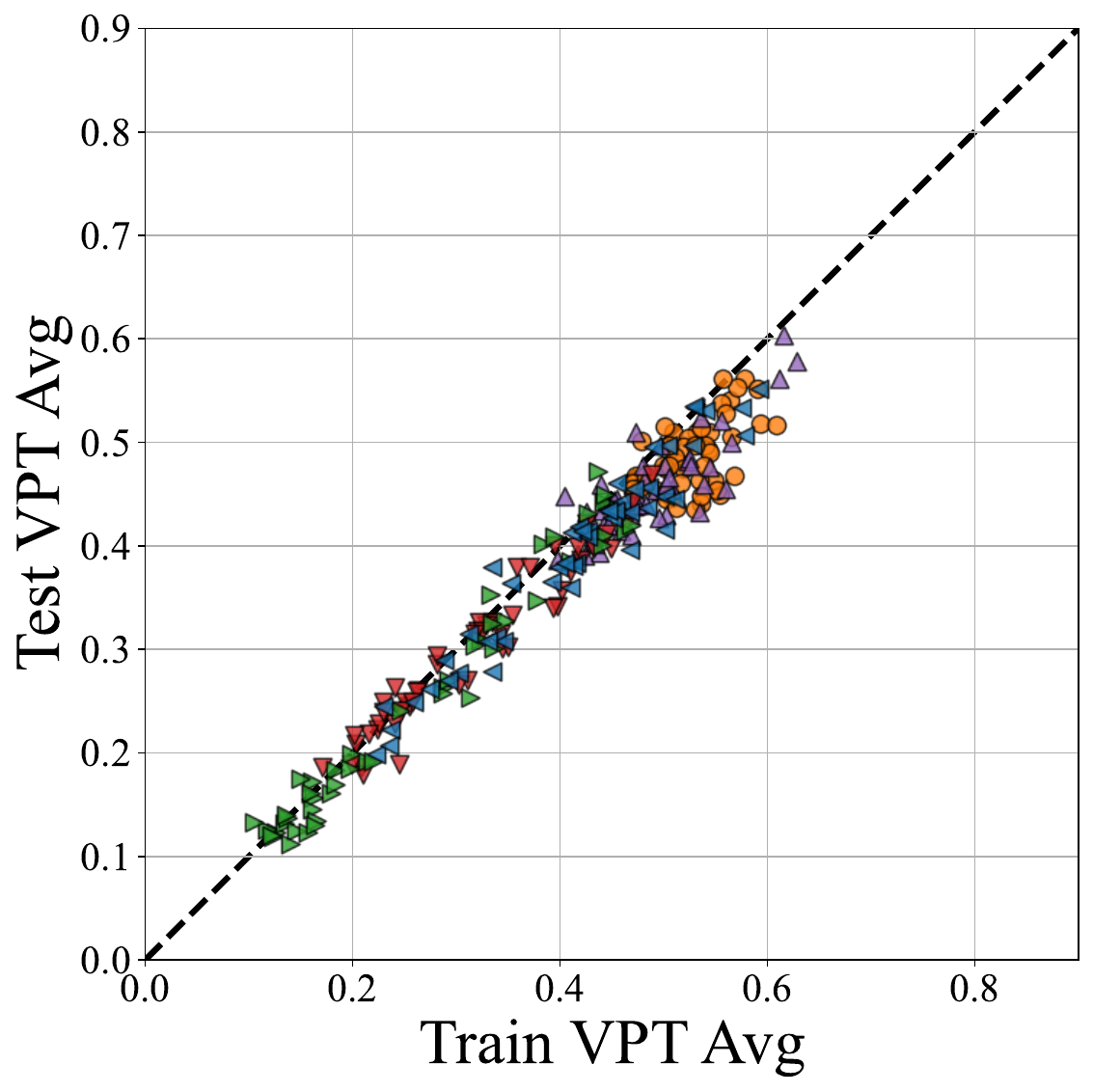}
        \caption{Baseline (F=10)}
        \label{fig:l96f10_overfit_base}
    \end{subfigure}%
    \begin{subfigure}{0.5\textwidth}
        \centering
        \includegraphics[width=.99\linewidth]{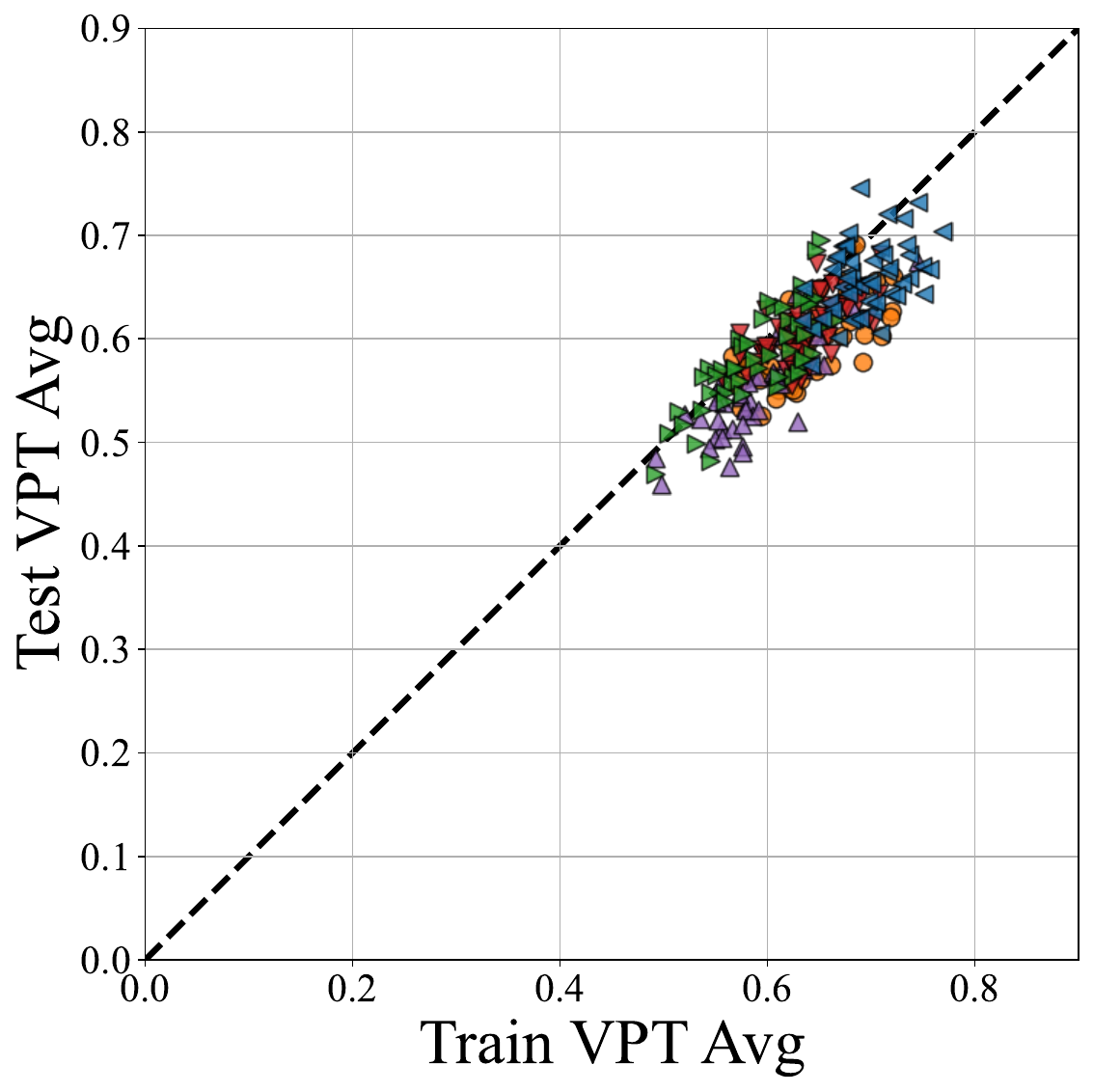}
        \caption{Augmented (F=10)}
        \label{fig:l96f10_overfit_aug}
    \end{subfigure}%

    \begin{subfigure}{0.5\textwidth}
        \centering
        \includegraphics[width=.99\linewidth]{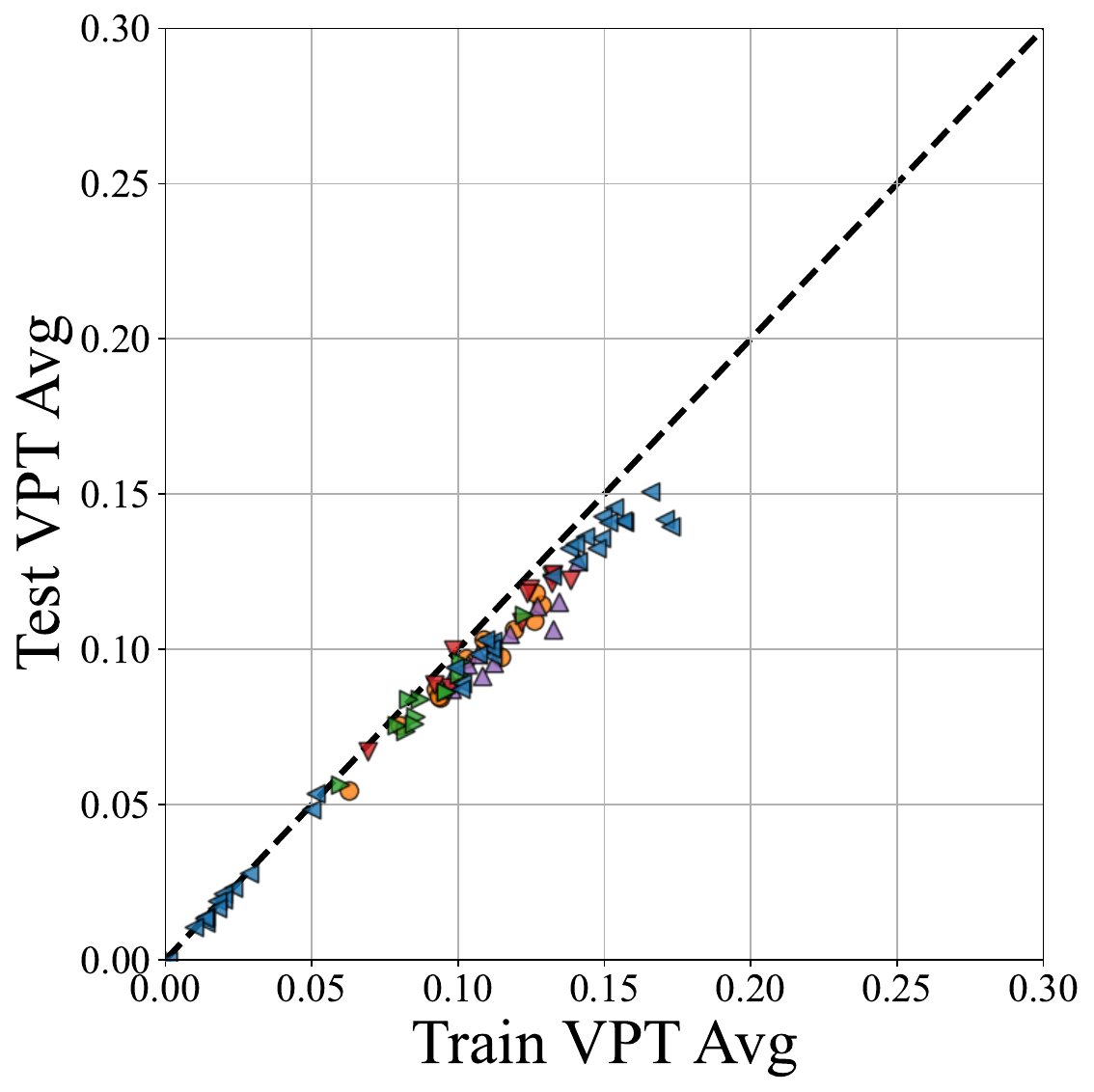}
        \caption{Baseline (F=20)}
        \label{fig:l96f20_overfit_base}
    \end{subfigure}%
    \begin{subfigure}{0.5\textwidth}
        \centering
        \includegraphics[width=.99\linewidth]{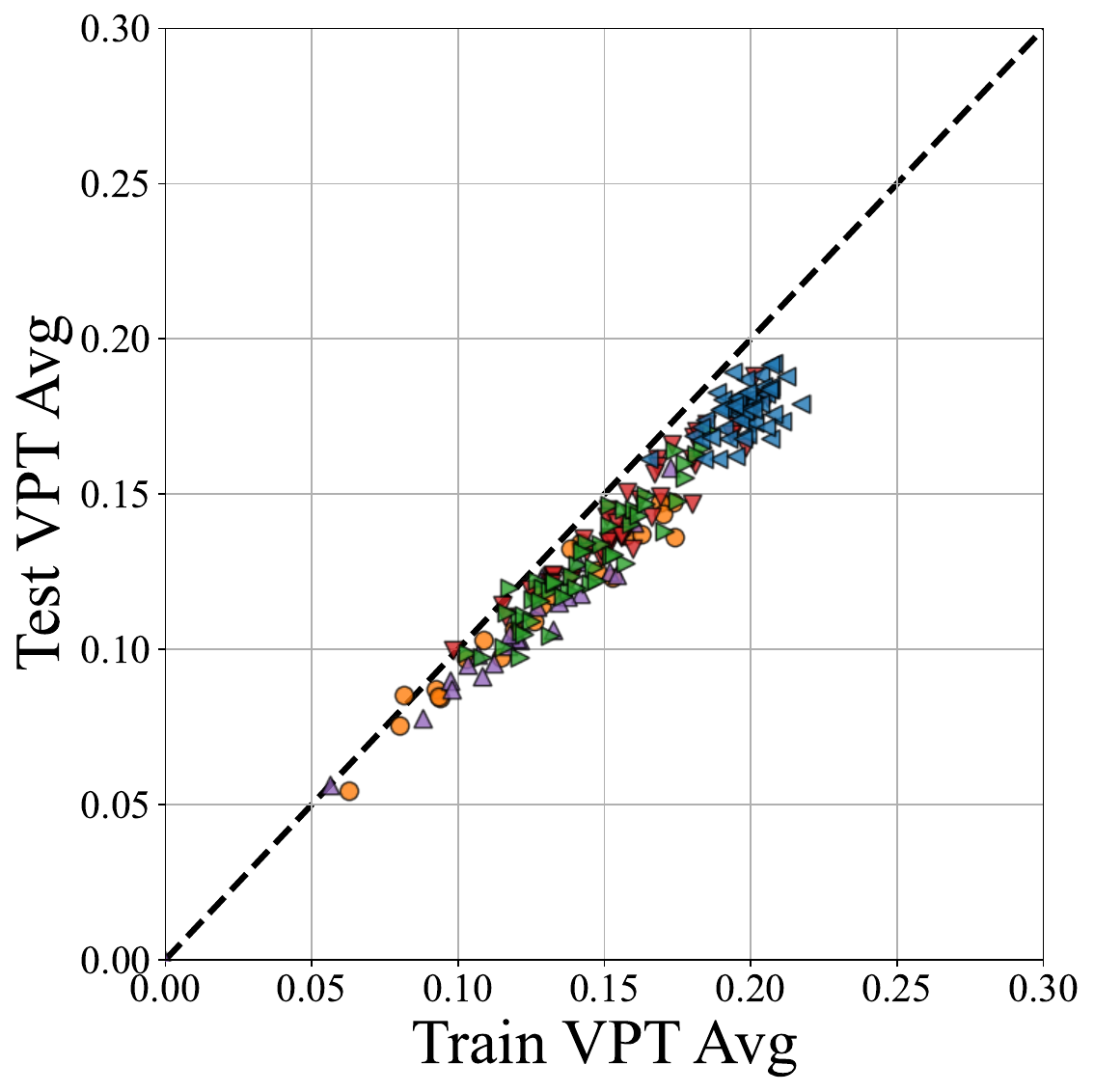}
        \caption{Augmented (F=20)}
        \label{fig:l96f20_overfit_aug}
    \end{subfigure}%
    
    \caption{
    Averge VPT over 100 initial conditions on the train dataset versus the test dataset for Multiscale Lorenz 96 ($F\in \{10,20 \}$).
    For each scatter plot, only the top 50 models (with respect to the average VPT on validation dataset) are shown for each class.
    }
    \label{fig:ml96_overfit}
\end{figure*}

To qualitatively validate the generalization performance of our models, we plot the train versus test VPT in~\Cref{fig:ml96_overfit}.
The dashed line $y=x$ represents the ideal generalization performance, where training and testing performance are equal; models below this line can be said to have overfit to the training set, achieving over-optimistic performance on this dataset split.
We restrict plotting to the top 50 models of each class, selecting models by the highest VPT on the validation data (averaged over 100 initial conditions), and plotting the training and test VPT averages.
From these plots, we observe that the augmentations proposed in this work serve to increase forecasting performance without inducing any significant increase in overfitting to the training set. 
This can be observed in~\Cref{fig:l96f10_overfit_aug,fig:l96f20_overfit_aug} by noting that models move upwards along the $y=x$ line without significantly falling beneat it.
Furthermore, the plots highlight another key difference between Transformers and RNNs: 
While RNN performance is most significantly increased over the baseline performance, Transformers are fairly robust to their hyperparameter selection before and after augmentation.
These results suggest that Transformers are relatively less improved as they are less sensitive to alternative hyperparameter selection. 
In contrast, RNNs (especially the RHN) achieve the top predictive performance, at the cost of significant sensitivity to how its hyperparameters are chosen.

\subsection{Perturbing Neural Mechanisms}

\Cref{tab:ml96} shows the effects of neural mechanism modification.
Here, we summarize the observations generated from these ablations. 

\subsubsection{Neural Gating (NG)}

The effects of Neural Gating (NG) on RNN variants can be observed by comparing baseline and NG rows of~\Cref{tab:ml96}.
For $F=10$, we see increases in average VPT for every RNN cell by varying its NG, spanning a relative increase over the baseline of 18.9\%-36.7\%.
However, for $F=20$, we only observe an increase for the GRU (+14.9\%) by using an uncoupled gate.
Data-independent gates (additive and learned rate) cause both the GRU and RHN to diverge, something which can be observed by the high-degree of 0 VPT models shown in~\Cref{fig:l96_vpt_gru,fig:l96_vpt_rhn} for NG.  
In contrast, the LSTM is more robust to data-independent gates likely due to its additional output gate and cell state. 
Nevertheless, these results suggest that neither data-dependent neural gate (coupled or uncoupled) is necessarily better than the other, in agreement with previous studies~\cite{greff2016lstm} and in support of considering both when tuning allows.

Few prior works consider the residual connections in Transformers as parameterized gates, and predominantly in the context of language modeling \cite{chai2020highway,davis2021catformer}.
Here, we observe that while the PostLN variant receives limited benefit from residual gating, the PreLN variant realizes a 17\% and 29.3\% increase in average VPT for $F=10$ and $F=20$, respectively.
Past works suggest that the PreLN Transformer has smoother loss landscapes and stable gradients during training but may fail to learn to use its residual branches \cite{liu2020understanding, wang2022deepnet, liu2023branchnorm}. 
The PostLN Transformer is known to suffer from instability during training, so more stabilization measures may be needed to better utilize gated connections (e.g., more careful initializations and learning rate schedules \cite{wang2022deepnet}).

\subsubsection{Attention (AT)}

We observe significant performance increases when augmenting RNNs with attention, as shown in~\Cref{tab:ml96}. 
For $F=10$, RHNs see the smallest relative increase (+28.4\%), and LSTMs see the largest (+45.9\%). 
In contrast with NG, we find attention improves performance in the $F=20$ setting. 
The RHN with attention (AT-RHN) is the highest-performing model with a VPT of 0.19, a relative increase of +24.9\% over the baseline RHN. 
Notably, when using attention with relative position bias, RNN performance is marginally worse than using scaled dot-product attention directly. 
Because positional biases result in increased parameters without increased performance, we neglect the further use of positionally-biased attention in RNNs for the remainder of this work.
When combining NG and attention simultaneously in RNNs, we observe that higher performance is possible. 
Both the RHN and GRU perform best with both variations for $F=10$.
Similarly, the GRU performs better with both augmentations for $F=20$.
While not the top-performing RNNs in every category, these results suggest that input-dependent gates augmented with dot-product attention are a strong starting point for training RNNs on chaotic dynamics.

The results exhibited in~\Cref{tab:ml96} suggest that for Transformers, the differences between attention with data-dependent and data-independent biases are negligible. 
Both yield similar distributions, median, and top-performing models. 
PreLN tends towards slight improvements with the data-dependent positional biases, however, given the limited increases in performance and the larger increases in parameter counts and training times, we opt for the cheaper data-independent positional biases for the remainder of experiments.

\subsubsection{Recurrence (R)}

Our experiments with recurrent Transformers were unsuccessful.
For many settings, recurrence harmed predictive performance more than neural gating increased it. 
For the PostLN Transformer, recurrent performance is less than baseline performance, as shown in the bottom rows of~\Cref{tab:ml96}. 
Given that the residual connections in the PostLN Transformer are known to exhibit stability issues, perhaps it is unsurprising that adding recurrent branches further exacerbates performance issues. 
For PreLN Transformers, recurrence slightly improves performance over baseline models but only when combined with neural gating. 
Furthermore, gated recurrent connections have significantly worse predictive performance than solely using gating within the PreLN Transformer. 
Given our inability to identify performant hyperparameterizations of recurrent Transformers for $F=10$, we neglect their application to $F=20$ and do not consider them further in this study. 

\section{Forecasting Dynamics of Kuramoto-Sivashinsky}
\label{sec:KS}

\subsection{Kuramoto-Sivashinsky Model}

The Kuramoto-Sivashinsky (K-S) equation is a fourth-order, nonlinear partial differential equation which has been used to model turbulence in a variety of phenomena.
Initially derived to model reaction-diffusion phase gradients \cite{kuramoto1978diffusion} and instabilities in laminar flame fronts \cite{sivashinsky1977nonlinear}, the K-S equation has become a useful benchmark for data-assimilation methods as an example of one-dimensional spatiotemporal chaos \cite{pathak2017using,pathak2018model,vlachas2020backpropagation,vlachas2022multiscale}.
While higher-dimensionality K-S equations are possible, we restrict our consideration to the one-dimensional K-S system,
\begin{align}
    \label{eq:ks}
    \frac{\partial u}{\partial t} & = -\nu\frac{\partial^4 u}{\partial x^4} - \frac{\partial^2 u}{\partial x^2} - u \frac{\partial u}{\partial x}
\end{align}
on the domain $\Omega = [0, L]$ with periodic boundary conditions $u(0, t) = u(L, t)$.
The dimensionless parameter $L$ has a direct effect on the dimensionality of the attractor, with the dimension scaling linearly for large values of $L$ \cite{manneville1985macroscopic}.

To spatially discretize~\Cref{eq:ks}, we select a grid size $\Delta x$ with $D = L/\Delta x + 1$ as the number of nodes. 
We denote each $u_i = u(i \Delta x)$ as the value of u at node $i \in \{ 0, ..., D - 1\}$.
We select $\nu = 1$, $L = 60$, $D = 128$, and $\delta t = 0.25$, solving this system with a fourth-order method for stiff PDEs first introduced in \cite{kassam2005fourthorder}.
We solve this system up to $T=6 \cdot 10^4$, generating a total of $24 \cdot 10^4$ samples. 
The first $10^4$ samples are treated as transients and discarded, with the remaining $23 \cdot 10^4$ samples equally partitioned into training and testing sets.
We adopt previous estimates of the MLE for K-S with this parameterization, taking $\Lambda_1 \approx 0.08844$ \cite{pathak2018model,vlachas2020backpropagation}.
All hyperparameters included in the search space are listed in~\Cref{app:hyper}.

\subsection{Results on Forecasting Kuramoto-Sivashinsky}

\begin{table}[htb!]
    \centering
    \begin{tabular}{|l|rr|}
    \hline
    & \multicolumn{2}{c|}{K-S} \\ 
    \diagbox[]{Model}{Metric} & VPT & PSD MSE \\
    \hline
    \hline
    LSTM & 47.86 & 2.71E-03 \\
    AT-LSTM & 203.16 (424.5\%) & 2.71E-03\\
    NG-LSTM & 47.77 (-0.2\%) & 1.02E-03 \\
    (AT+NG)-LSTM &  203.16 (424.5\%) & 1.02E-03 \\
    \hline 
    GRU & 40.20 & 3.32E-03 \\
    AT-GRU & 263.96 (656.6\%) & 1.86E-03 \\
    NG-GRU & 98.03 (243.9\%) & 1.07E-03 \\
    (AT+NG)-GRU & 263.96 (656.6\%) & 1.86E-03 \\
    \hline 
    RHN & 44.44 & 2.17E-03 \\
    AT-RHN & 293.65 (660.8\%) & \textbf{6.93E-04} \\
    NG-RHN & 126.13 (283.8\%) & 1.39E-03 \\
    (AT+NG)-RHN & \textbf{313.8 (706.1\%)} & 1.30E-03 \\
    \hline 
    Transformer(Post) & 222.92 & 7.15E-04 \\
    NG-Transformer(Post) & 222.92 (-) & 2.00E-03 \\
    \hline 
    Transformer(Pre) & 193.86 & 1.55E-03\\
    NG-Transformer(Pre) & 208.36 (7.5\%) & 1.55E-03 \\
    \hline
    \end{tabular}
    \caption{
        Test performance of top models averaged over 100 initial conditions, stratified by model class and mechanism.
        The best in each column is highlighted in bold. 
        Modified architectures are denoted with prefixes: Neural Gating (NG) and Attention (AT).
        Relative performance changes with respect to baseline models are displayed in parentheses.
    }
    \label{tab:ks}
\end{table}

\begin{figure*}[htb!]
    \centering
    K-S

    \begin{subfigure}{0.33\textwidth}
        \centering
        \includegraphics[width=.99\linewidth]{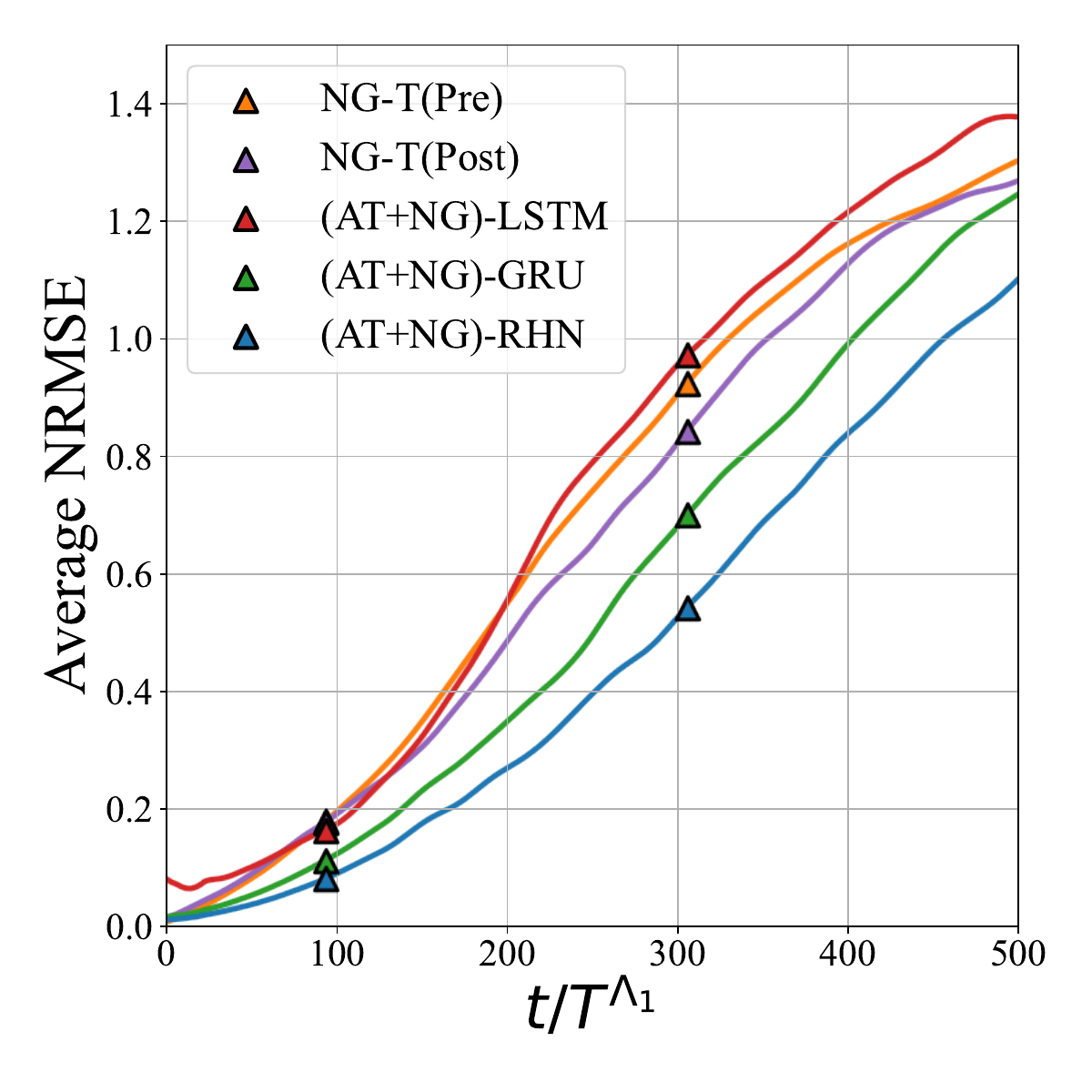}
        \caption{Best Models}
        \label{fig:ks_nrmse_all}
    \end{subfigure}%
    \begin{subfigure}{0.33\textwidth}
        \centering
        \includegraphics[width=.99\linewidth]{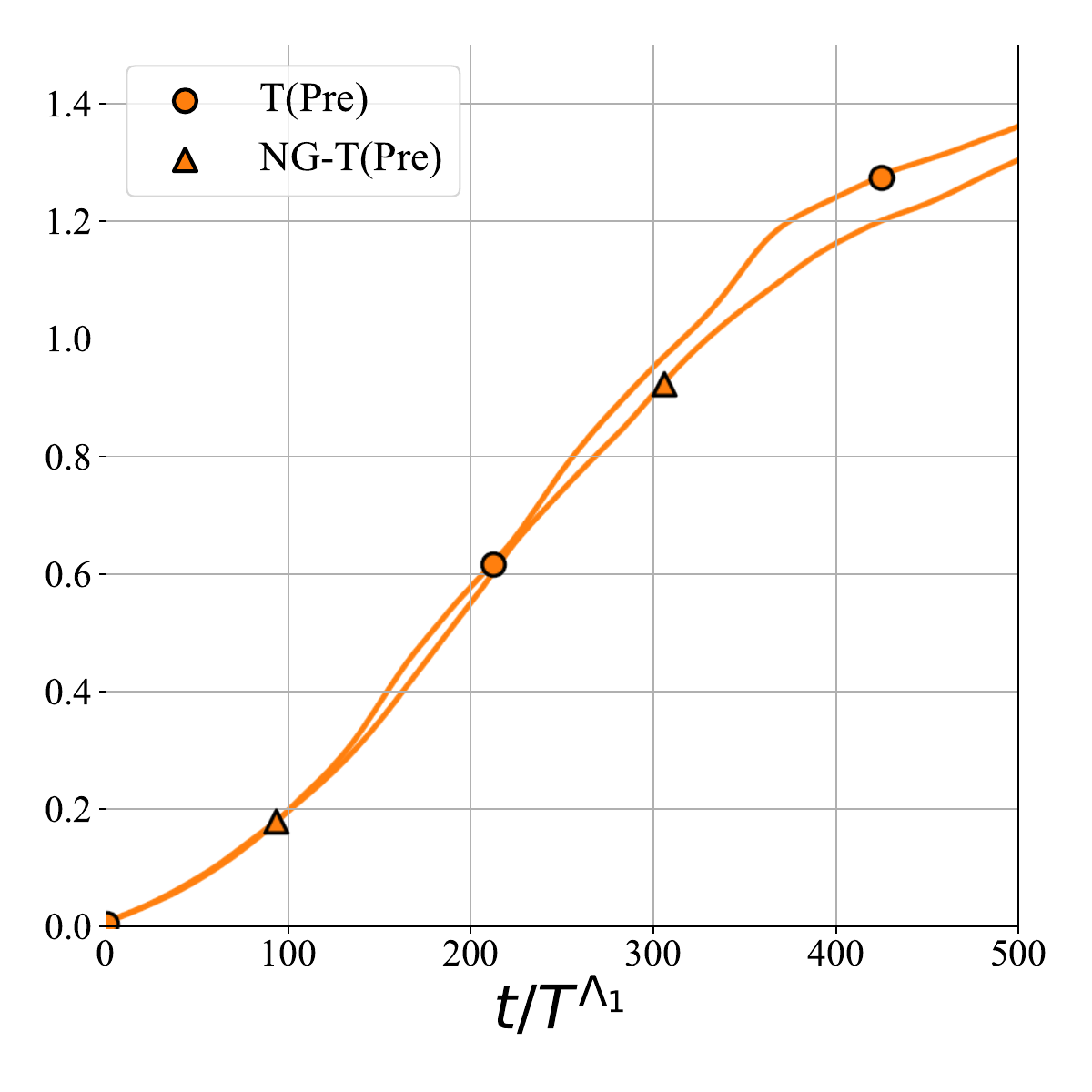}
        \caption{PreLN Transformer}
        \label{fig:ks_nrmse_pre}
    \end{subfigure}%
    \begin{subfigure}{0.33\textwidth}
        \centering
        \includegraphics[width=.99\linewidth]{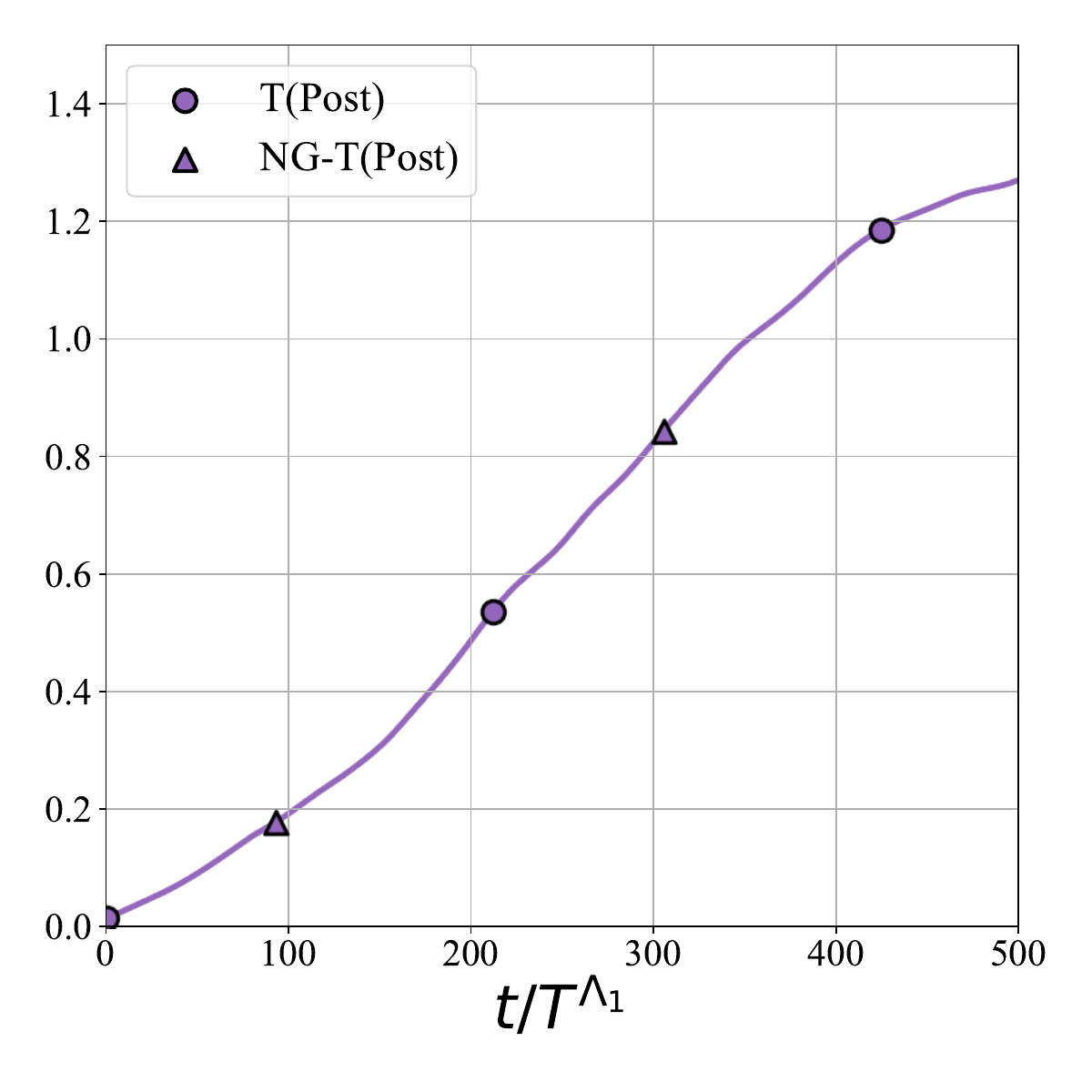}
        \caption{PostLN Transformer}
        \label{fig:ks_nrmse_post}
    \end{subfigure}%
    
    \begin{subfigure}{0.33\textwidth}
        \centering
        \includegraphics[width=.99\linewidth]{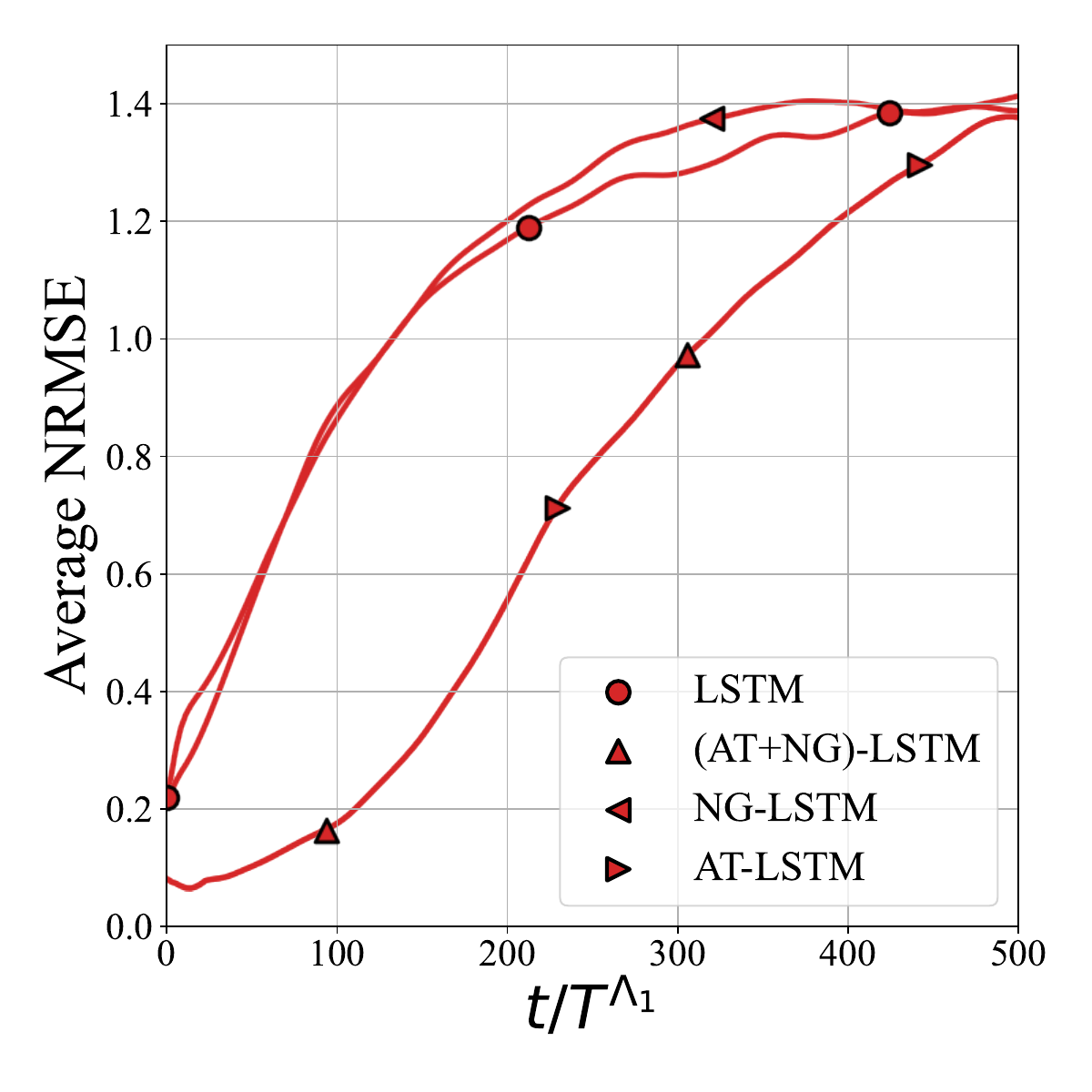}
        \caption{LSTM}
        \label{fig:ks_nrmse_lstm}
    \end{subfigure}%
    \begin{subfigure}{0.33\textwidth}
        \centering
        \includegraphics[width=.99\linewidth]{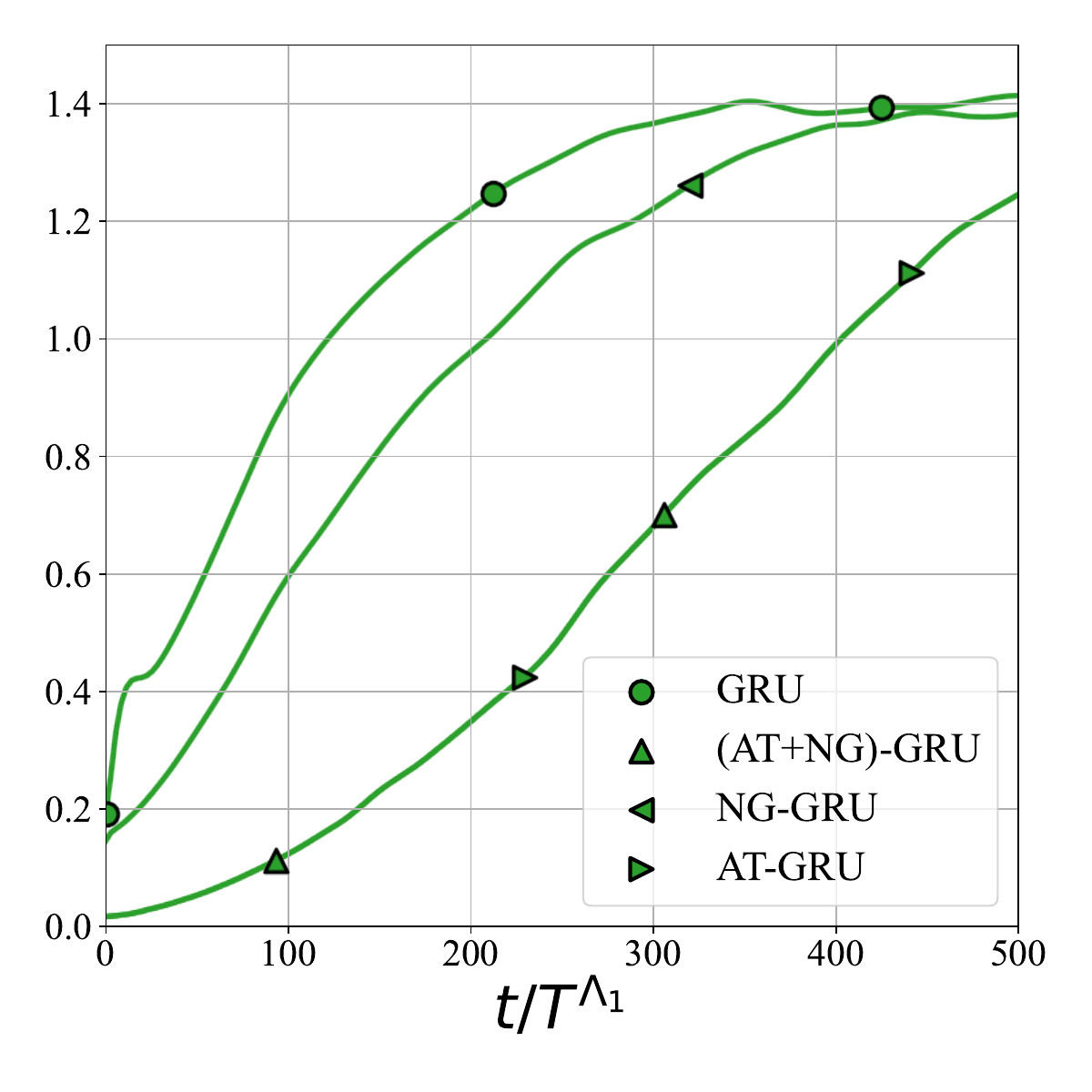}
        \caption{GRU}
        \label{fig:ks_nrmse_gru}
    \end{subfigure}%
    \begin{subfigure}{0.33\textwidth}
        \centering
        \includegraphics[width=.99\linewidth]{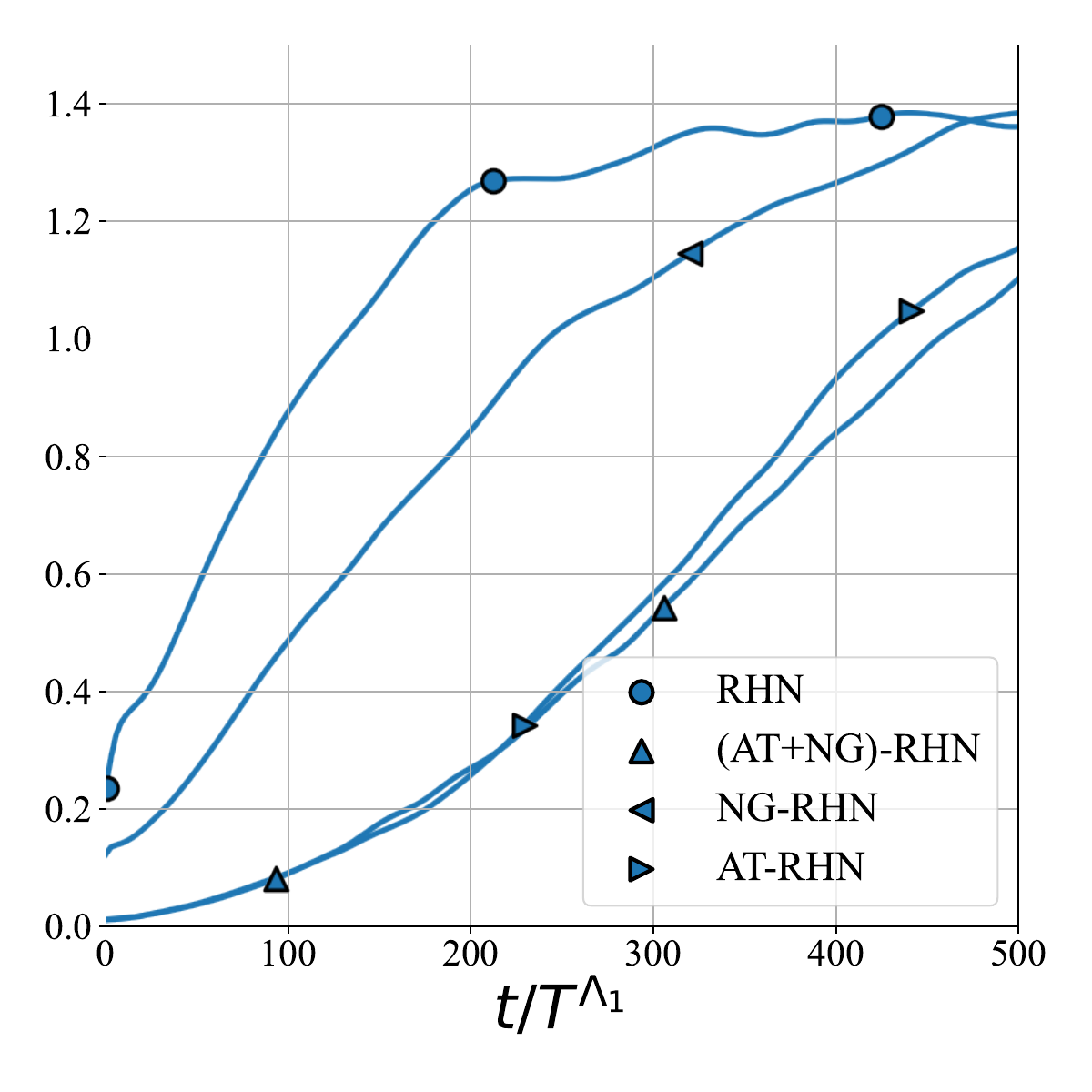}
        \caption{RHN}
        \label{fig:ks_nrmse_rhn}
    \end{subfigure}%
    \caption{
        The evolution of average NRMSE error for top models of each architecture on the K-S Equation. 
        NRMSE is averaged with respect to the initial conditions in the test split.
        Though many models are able to achieve comparable performance, the RHN exhibits the least error. 
    }
    \label{fig:ks_nrmse}
\end{figure*}

\begin{figure*}[htb!]
    \centering
    \includegraphics[width=\textwidth]{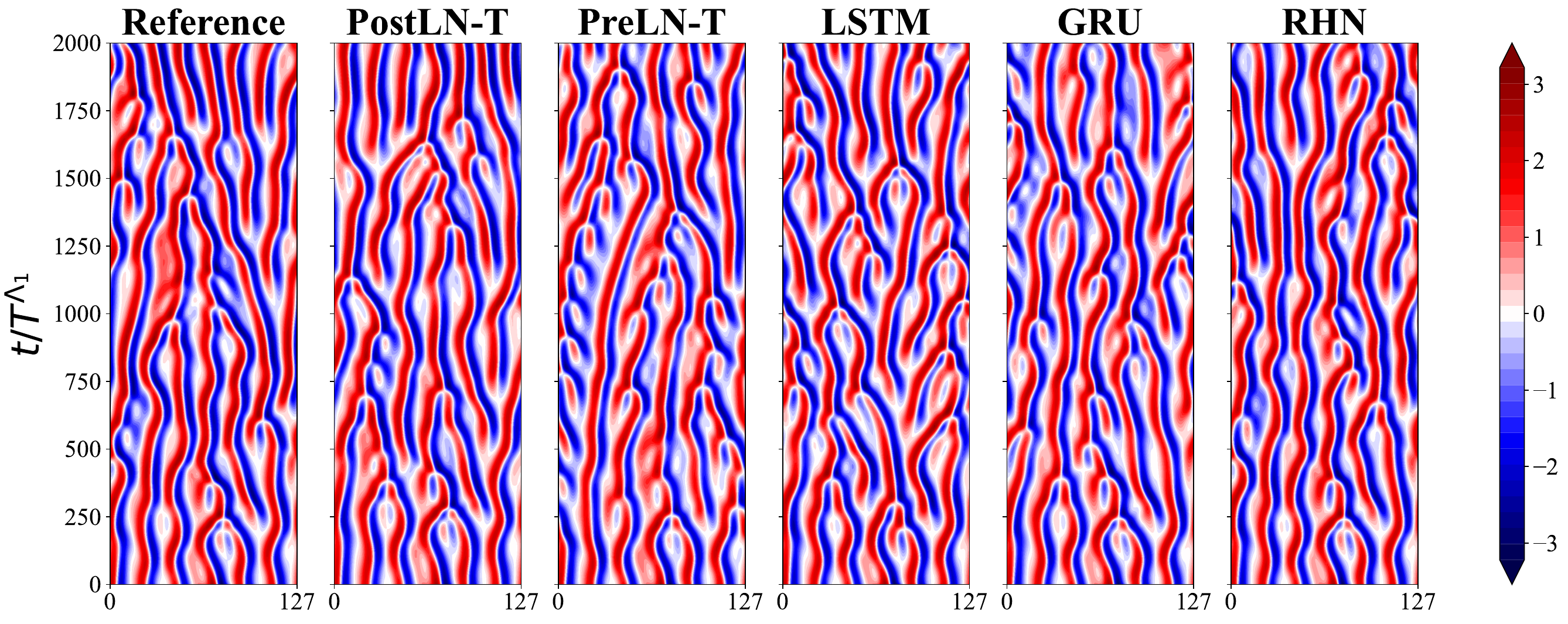}
    \includegraphics[width=\textwidth]{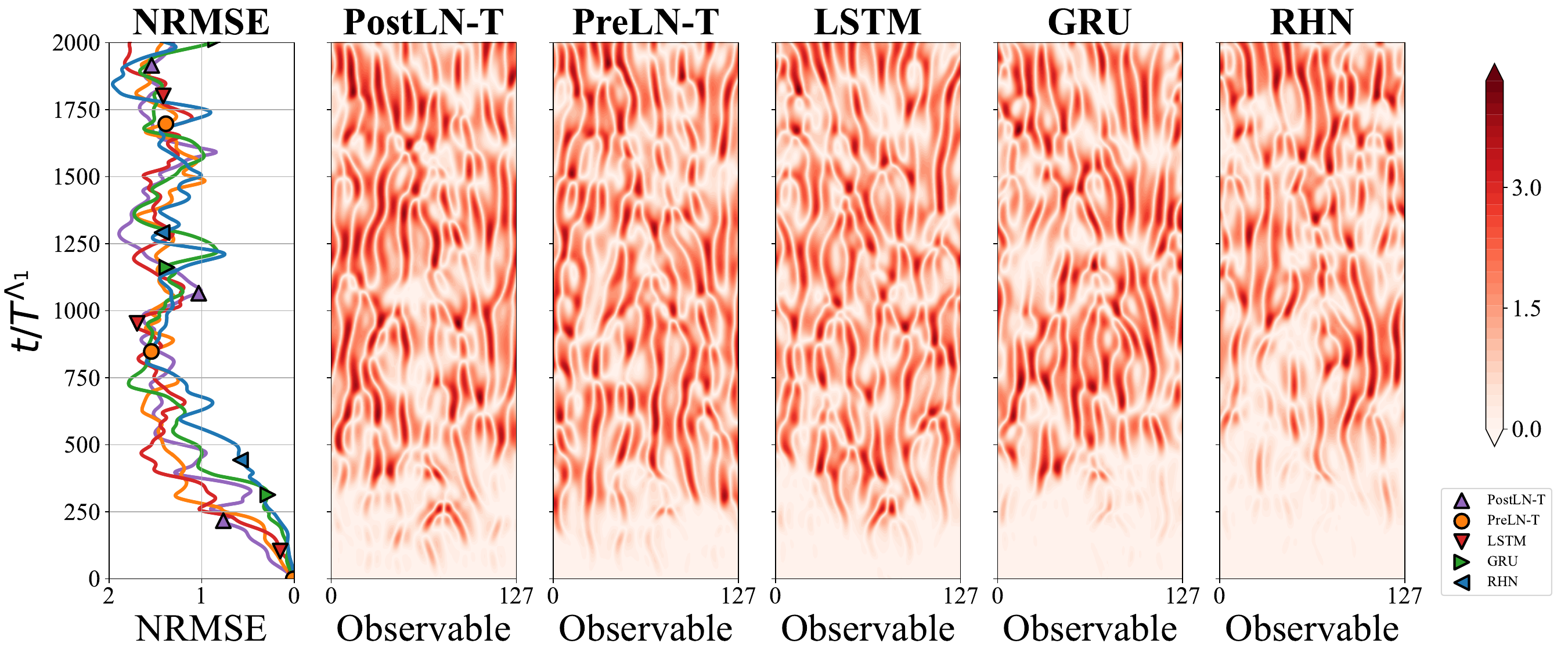}
    \caption{
        2000 Lyapunov times of iterative predictions
        for a trajectory sampled randomly from the test split of the dataset for the K-S system.
        Errors are plotted in terms of normalized root squared error (NRSE).
    }
    \label{fig:ks_qual}
\end{figure*}

For each architectural class, we identify the top-performing model based on the highest validation VPT (averaged over 100 initial conditions) and plot the evolution of its average test NRMSE in~\Cref{fig:ks_nrmse}.
As with the Lorenz-96 system, we display the best model of each architectural class and for each architectural variation in~\Cref{fig:ks_nrmse_all}.
This plot shows that the RHN, when augmented with attention, outperforms all other models quite significantly by maintaining a lower NMRSE for much longer than other architectures. 
In~\Cref{fig:ks_nrmse_pre,fig:ks_nrmse_post}, we observe that both Transformer variants perform similarly to one another, and that NG has a marginal effect on this system.
In the joint plot of \Cref{fig:ks_nrmse_all}, we see that the best-performing RHN and GRU outperform both Transformer variants.
The LSTM is the least performant architecture on the K-S system, although its performance is still significantly improved by the variants explored in this work, as indicated in~\Cref{fig:ks_nrmse_lstm}.
As illustrative examples, we plot iterative prediction plots of the top-performing models for a randomly selected test sample, shown in~\Cref{fig:ks_qual}.

\begin{figure*}[htb!]
    \centering
    K-S

    \begin{subfigure}{0.33\textwidth}
        \centering
        \includegraphics[width=.99\linewidth]{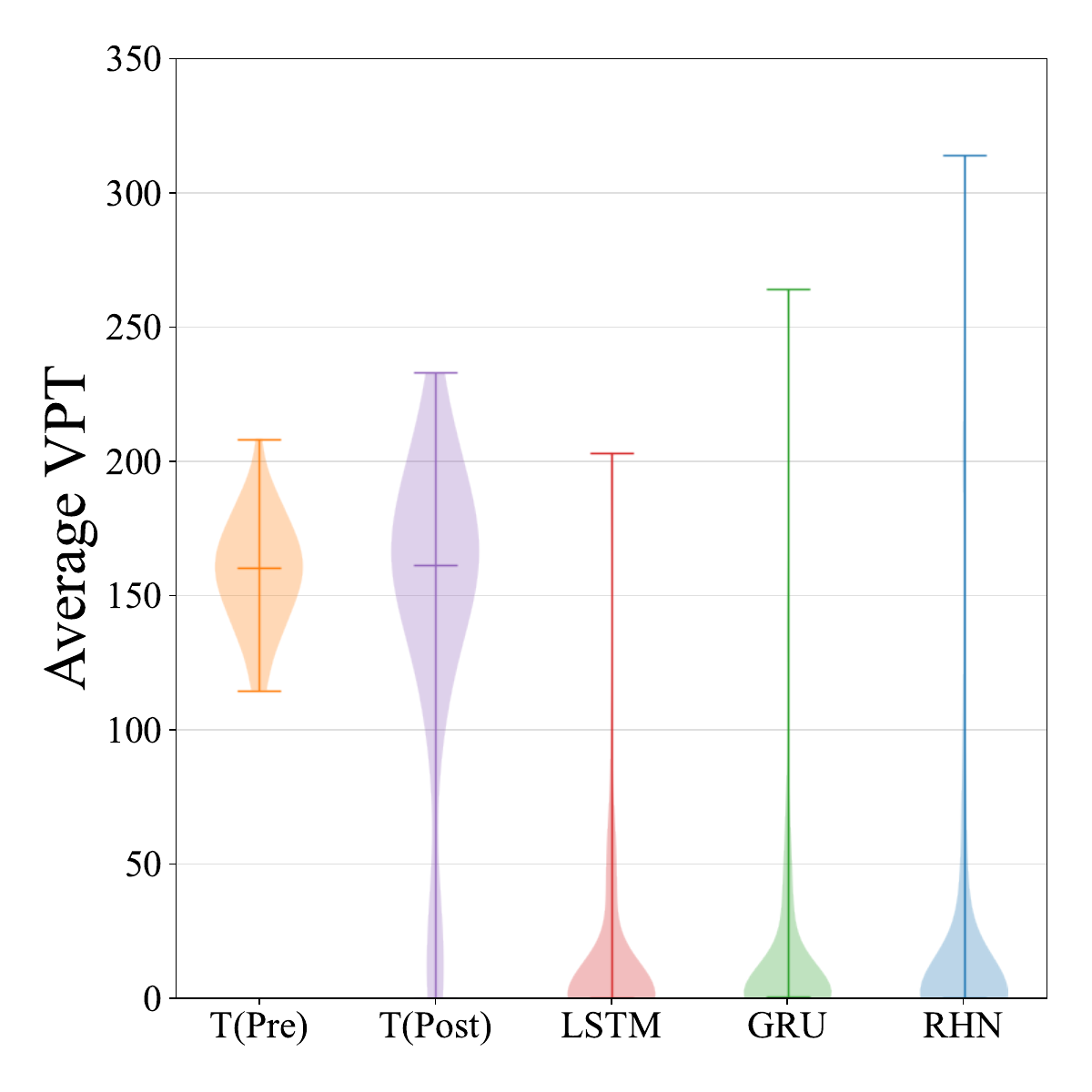}
        \caption{Best Models}
        \label{fig:ks_vpt_all}
    \end{subfigure}%
    \begin{subfigure}{0.33\textwidth}
        \centering
        \includegraphics[width=.99\linewidth]{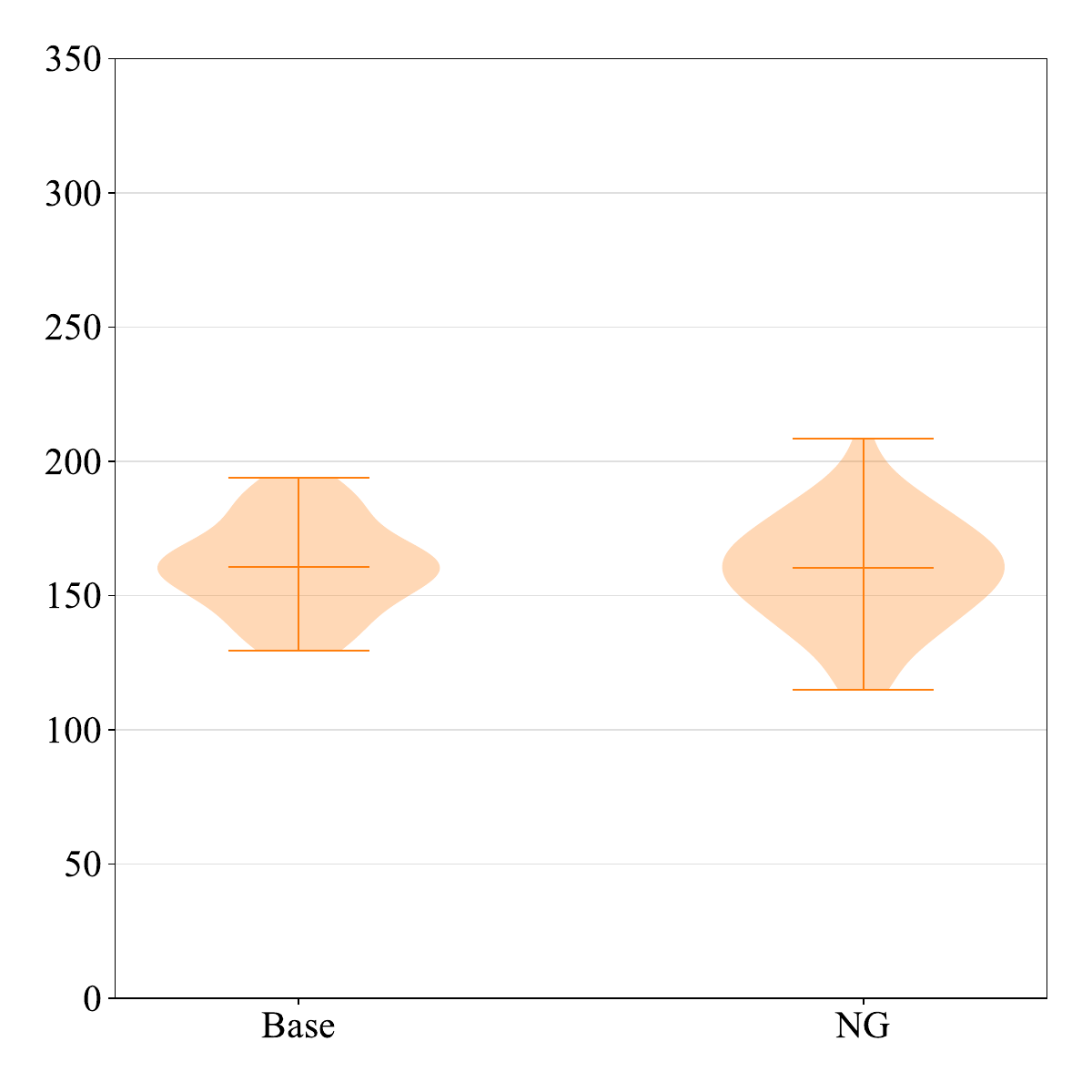}
        \caption{PreLN Transformer}
        \label{fig:ks_vpt_pre}
    \end{subfigure}%
    \begin{subfigure}{0.33\textwidth}
        \centering
        \includegraphics[width=.99\linewidth]{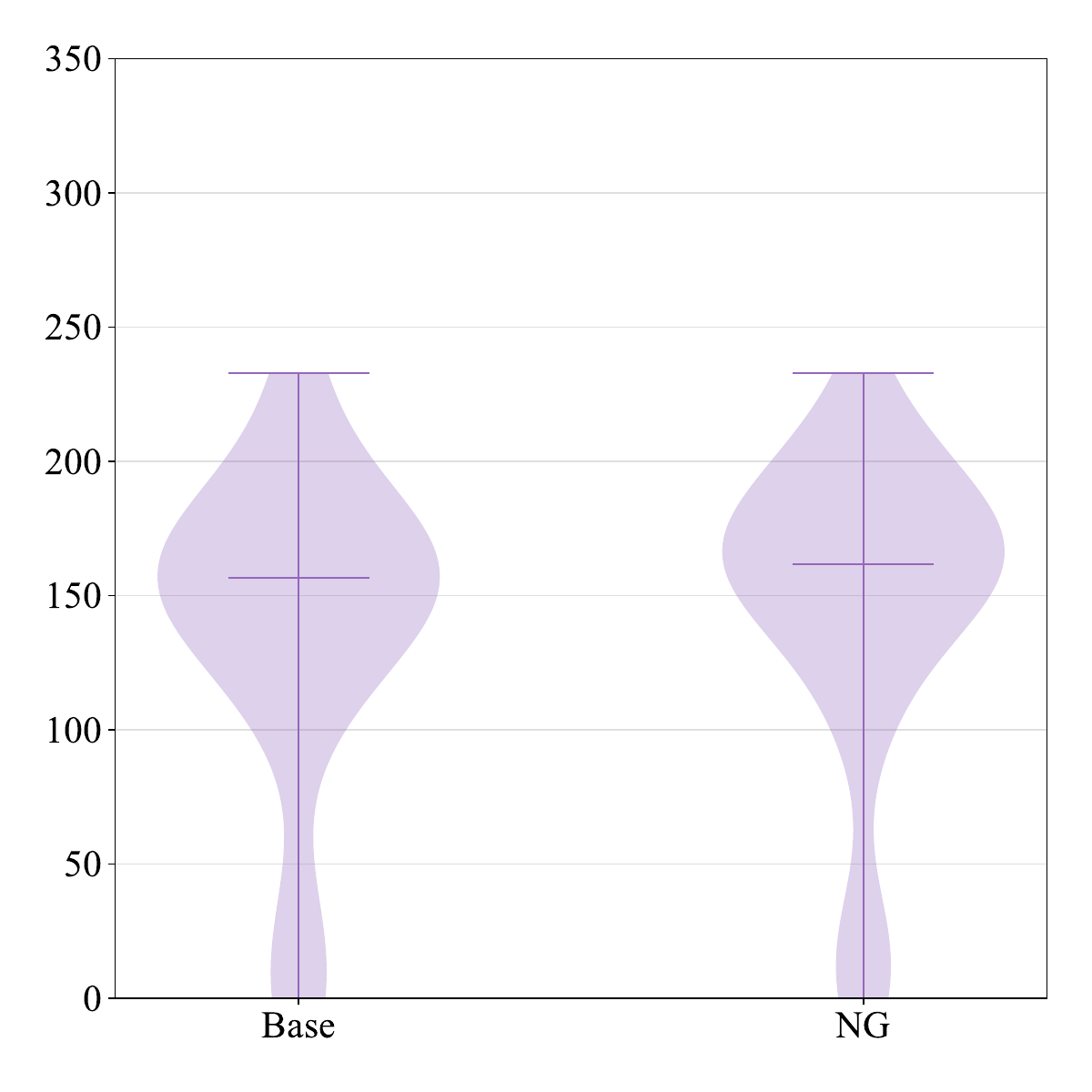}
        \caption{PostLN Transformer}
        \label{fig:ks_vpt_post}
    \end{subfigure}%
    
    \begin{subfigure}{0.33\textwidth}
        \centering
        \includegraphics[width=.99\linewidth]{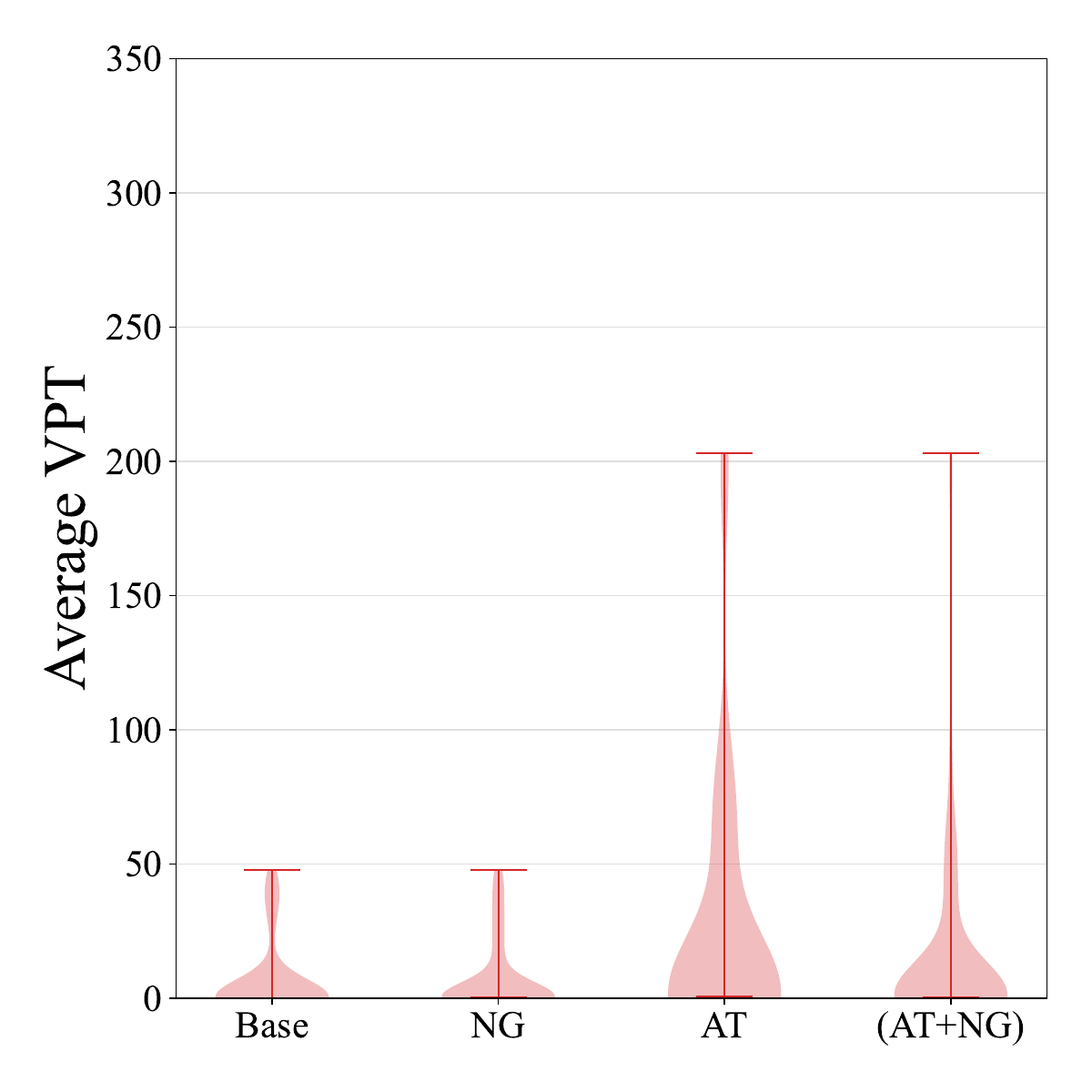}
        \caption{LSTM}
        \label{fig:ks_vpt_lstm}
    \end{subfigure}%
    \begin{subfigure}{0.33\textwidth}
        \centering
        \includegraphics[width=.99\linewidth]{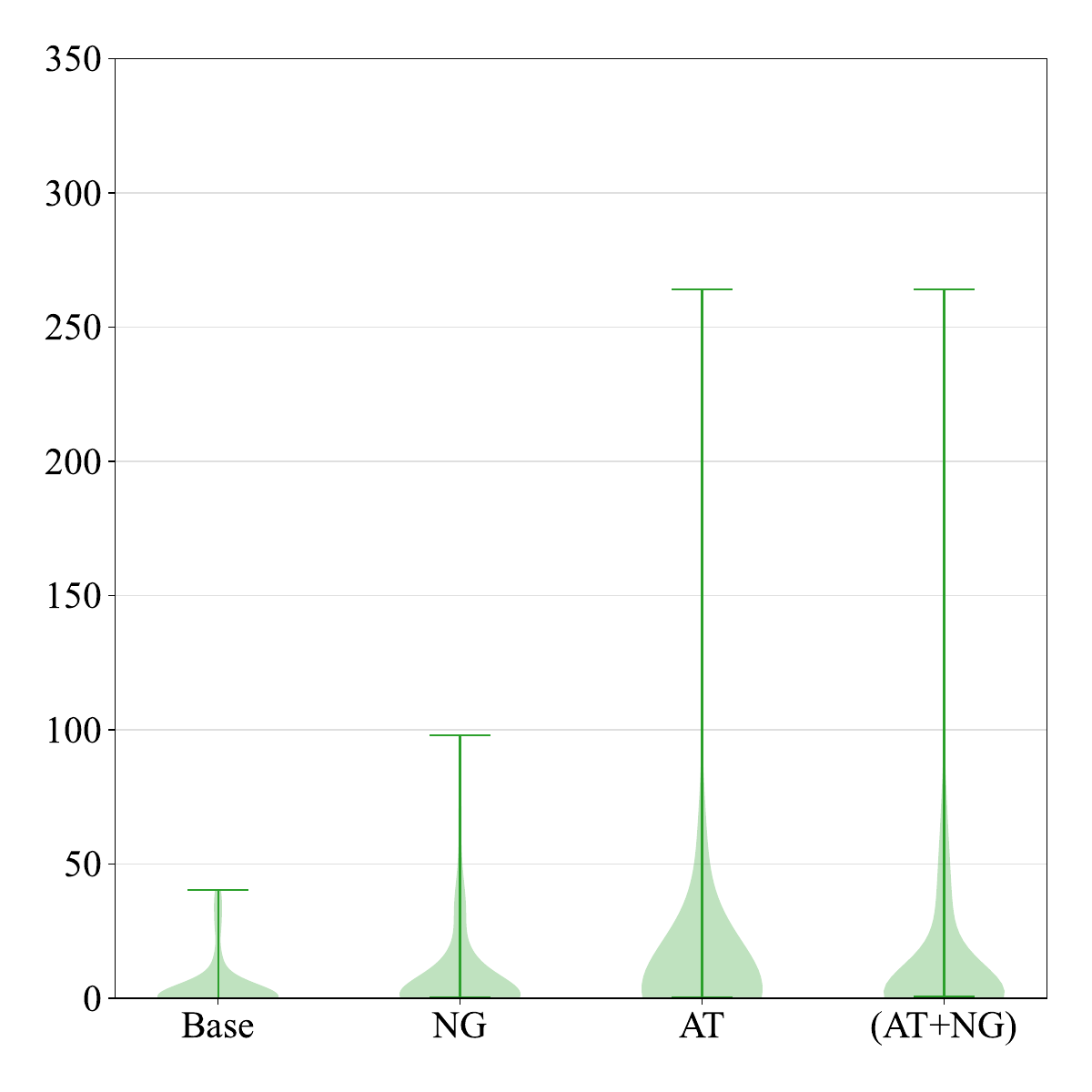}
        \caption{GRU}
        \label{fig:ks_vpt_gru}
    \end{subfigure}%
    \begin{subfigure}{0.33\textwidth}
        \centering
        \includegraphics[width=.99\linewidth]{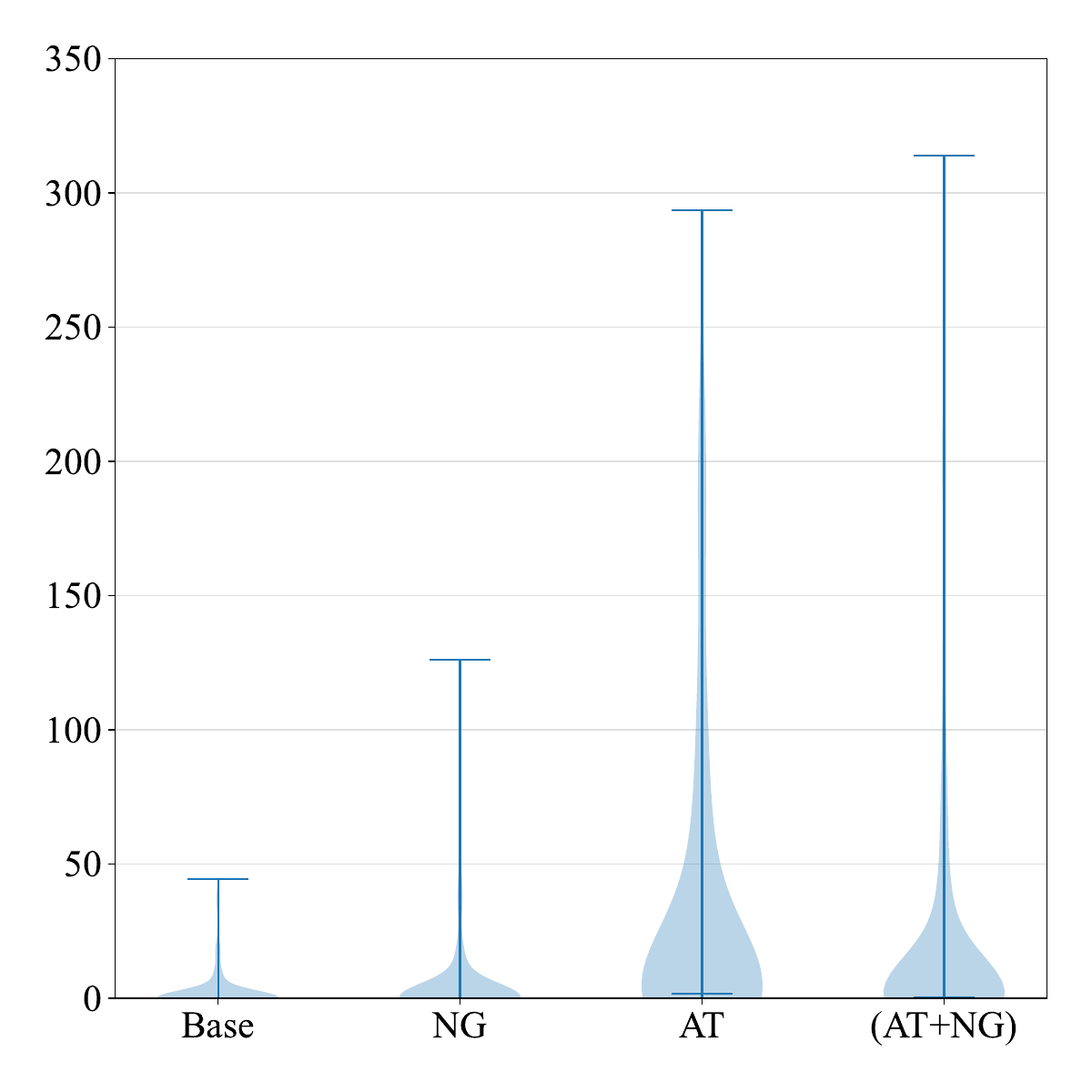}
        \caption{RHN}
        \label{fig:ks_vpt_rhn}
    \end{subfigure}%
    \caption{
        Violin plots showing the smoothed kernel density estimate of average VPT performance over all models trained on the K-S equation.
        The upper-left plot displays the distribution of NG Transformers and the (AT+NG) RNNs for cross-architectural comparison.
    }
    \label{fig:ks_VPT}
\end{figure*}

As a measure of hyperparameter sensitivity, we plot the smoothed kernel density estimate of architecture VPT as violin plots in~\Cref{fig:ks_VPT}.
We observe an even higher degree of hyperparameter sensitivity for RNNs on the K-S system compared to the RNNs trained to model the Multiscale Lorenz-96 system. 
While the best RNN models perform significantly better than all of the Transformer models considered, on average, they perform significantly worse due to the large number of hyperparameters that fail to learn a reasonable model that achieves a VPT greater than 0.
This can be seen by the high-concentration of probability mass in VPT plots for RNNs in~\Cref{fig:ks_vpt_lstm,fig:ks_vpt_rhn}.
In contrast, the Transformer models are more robust over the considered set of hyperparameters.
For example, all considered PreLN Transformers achieve an average test VPT of over 100 even in the worst instance, as is evident in~\Cref{fig:ks_vpt_pre}. 
Notably, the best PostLN Transformer outperforms the PreLN Transformer on the K-S system as well but is not as consistent as the PreLN variant as seen by comparing the violins in~\Cref{fig:ks_vpt_pre,fig:ks_vpt_post} and noting the much broader range of VPT for the PostLN variant.

\begin{figure}[htb!]
    \centering
    K-S
    
    \includegraphics[width=0.5\textwidth]{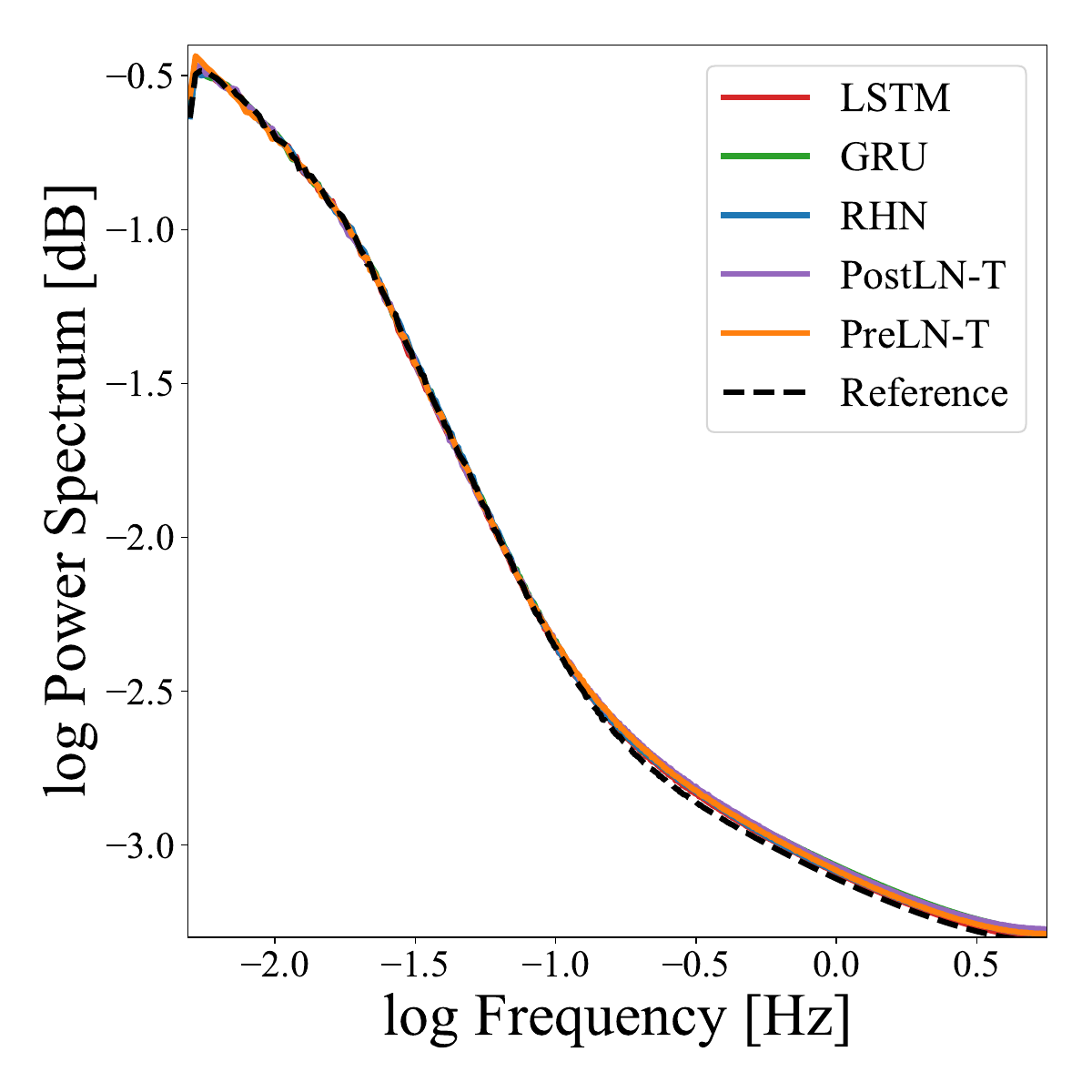}
    \caption{
        Predicted power spectra for top models selected by lowest deviation from the true power spectra (measured on the validation split). We observe that models are able to identify weights that allow faithful reproductions of the power spectra of the true underlying system.
    }
    \label{fig:KS_psd}
\end{figure}

Quantitative results on forecasting the K-S system are shown in~\Cref{tab:ks}.
When considering baseline Transformers and RNNs, we find that Transformers outperform RNNs. 
However, (AT+NG)-RNNs exceed the best Transformers for both the RHN and the GRU cells, with the (AT+NG)-LSTM performing almost equivalently to the Transformer models. 
While attention improves RNNs more than augmenting their gates alone, the effects on forecasting performance are significantly larger for the K-S system than for the partially-observed Lorenz-96 system with the best RNNs exceeding their baseline performance 2-7x over.
In contrast, the effects of varying the gates of Transformer models prove to be minimal for this system.
As a complement to the quantitative values of PSD MSE, we plot the difference in predicted power spectra in~\Cref{fig:KS_psd}.
We observe that all models maintain a close agreement with the ground-truth spectra in all but the highest frequency bands of the spectrum.

\begin{figure}[htb!]
    \centering
    K-S

     \begin{subfigure}{0.5\textwidth}
        \centering
        \includegraphics[width=.99\linewidth]{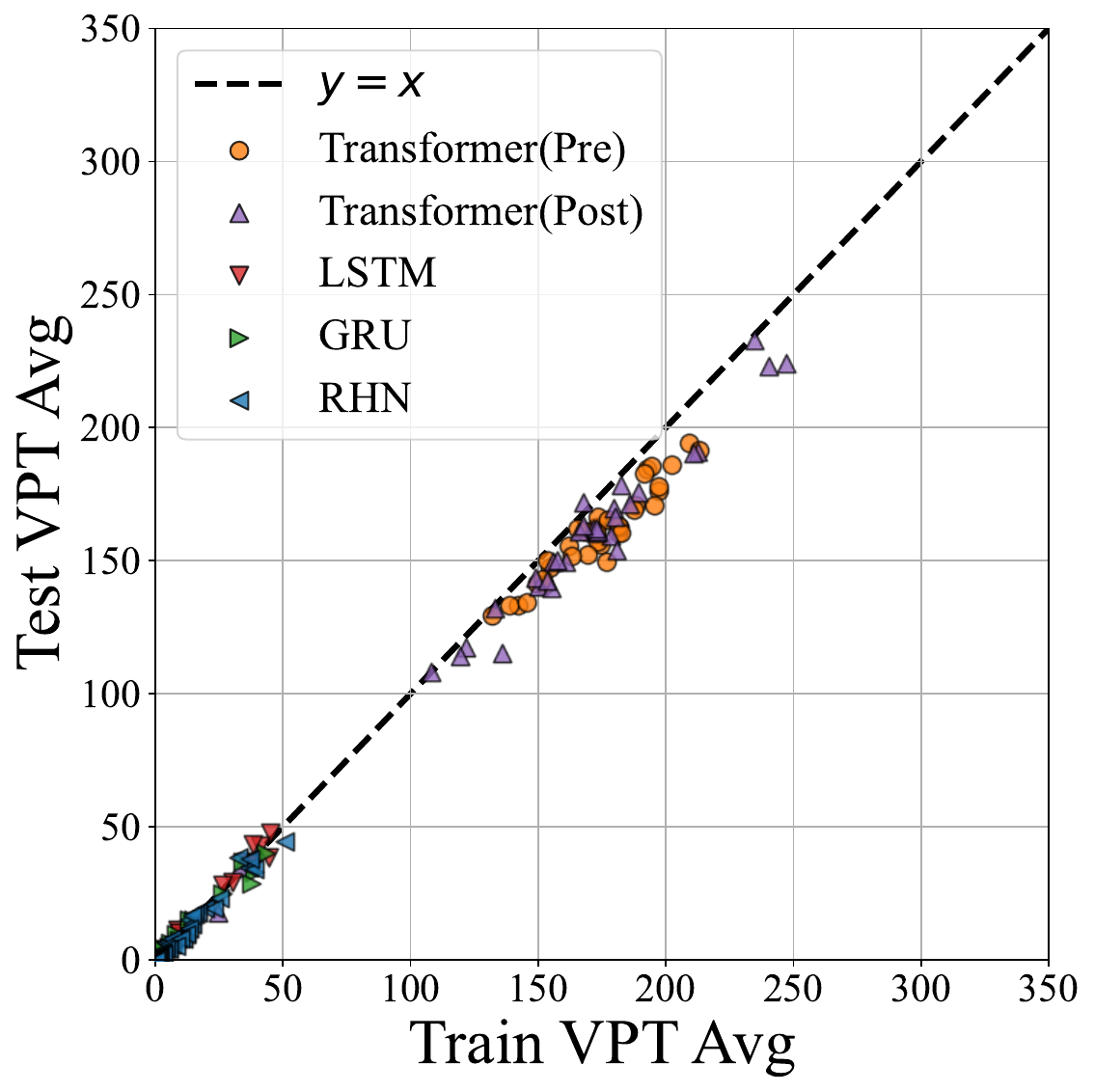}
        \caption{Baseline}
        \label{fig:ks_overfit_base}
    \end{subfigure}%
    \begin{subfigure}{0.5\textwidth}
        \centering
        \includegraphics[width=.99\linewidth]{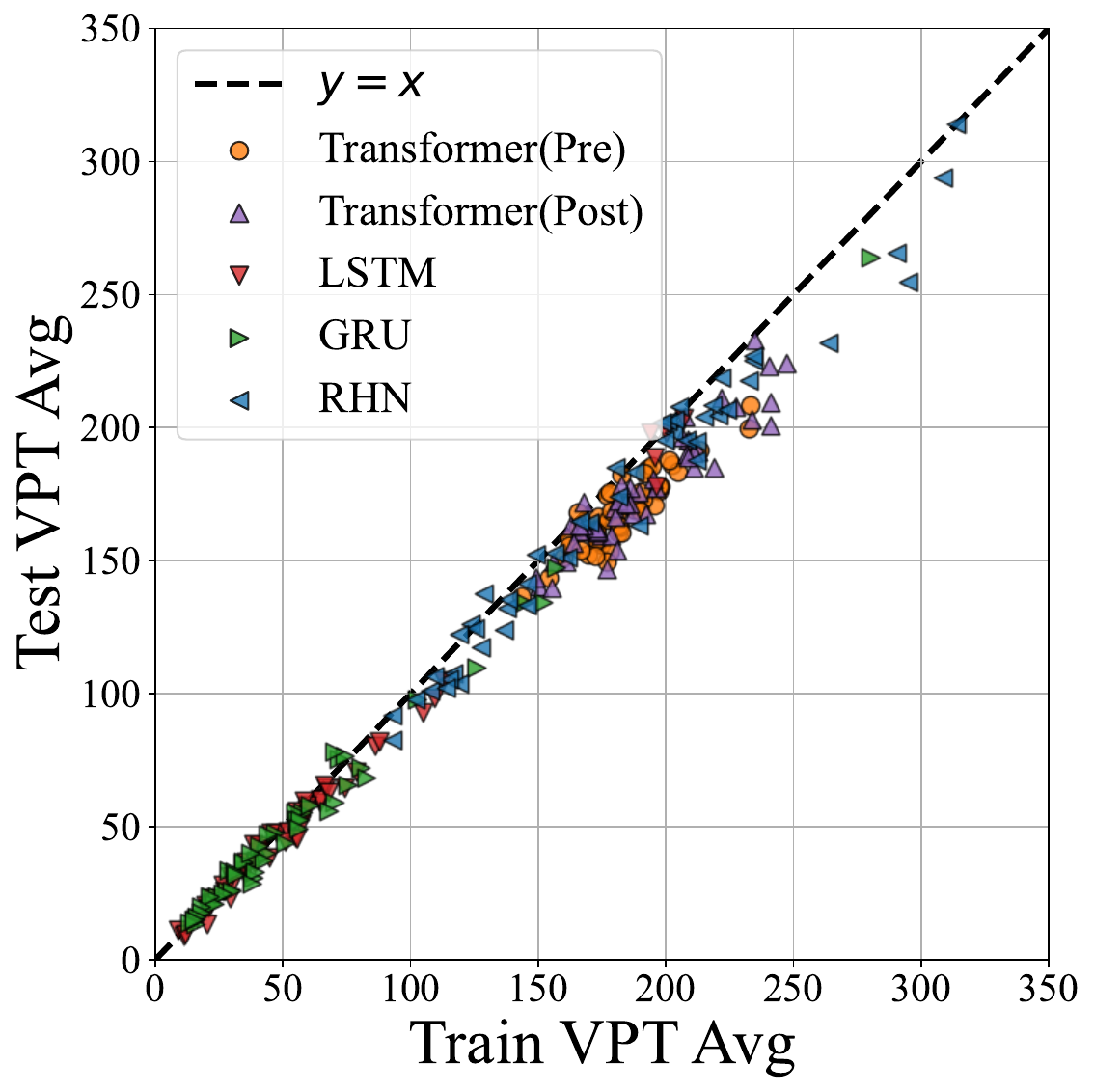}
        \caption{Augmented}
        \label{fig:ks_overfit_aug}
    \end{subfigure}%
    \caption{
    Average VPT over 100 initial conditions on the train versus the test set for the K-S system.
    For each scatter plot, only the top 50 models (with respect to validation average VPT) are shown for each class.
    }
    \label{fig:KS_overfit}
\end{figure}

Finally, as a qualitative display of the generalization performance of our models, we plot the train versus test VPT in~\Cref{fig:KS_overfit}.
As before, the $y=x$ line indicates ideal generalization where train and test performance are equivalent.
This figure yields similar trends as~\Cref{fig:ml96_overfit} for Multiscale Lorenz-96:
Transformers are more robust to the hyperparameter selection (though with less of an increase in performance when considering modified variants) and RNNs are extremely sensitive to their selected hyperparameters.
We observe limited overfitting by noting that most models fall fairly close to the $y=x$ line.

\section{Real-World Time-Series Data}
\label{sec:realworld}

There is broad interest in applying deep learning methods to long-term temporal forecasting for real-world phenomena like electricity grid demand, traffic patterns, weather trends, and industrial efficiency \cite{lai2018modeling,li2019enhancing,liu2021pyraformer,wu2021autoformer,zhou2021informer}. 
Unfortunately, despite the purported success of deep models on long-term forecasting, recent work clearly shows that many of the complex, deep learning methods fail to outperform a linear, autoregressive baseline \cite{zeng2022transformers}. 
The failure to overcome such a simple baseline speaks to the difficulty of this class of problems and shows that it is a largely unsolved problem.

Our approach varies from prior applications of deep learning to real-world forecasting. 
Instead of using encoder-decoder, sequence-to-sequence formalism, and targeting long-term horizons, we focus on achieving valid, short-term predictions of real-world trends and only consider decoder-only sequence models. 
Our intention is not to attain state-of-the-art forecasting performance with these datasets. 
Instead, we seek to understand how the modified architectures considered here affect predictive performance for real-world data using standard benchmark datasets. 

\subsection{Real-World Datasets}

We evaluate our models using seven standard real-world datasets: The four subsets of Electricity Transformer Temperature (ETT) \cite{zhou2021informer}, Traffic, Electricity, and Weather. 
We obtain the latter three datasets from the Autoformer repository \cite{wu2021autoformer}. 
All datasets are multivariate datasets. 
We include basic statistics about their length, dimensionality, and resolution in~\Cref{tab:real_data_desc}.

\begin{table}[htb!]
    \centering
    \begin{tabular}{|c|rrr|}
        \hline 
        \textbf{Dataset} & $N$ & $d$ & $\Delta t$ \\ 
        \hline 
        \hline 
        ETTh1 & 17420 & 7 & 1 hour \\ 
        ETTh2 & 17420 & 7 & 1 hour \\ 
        ETTm1 & 69680 & 7 & 15 min \\ 
        ETTm2 & 69680 & 7 & 15 min\\ 
        Electricity & 26304 & 321 & 1 hour \\
        Traffic & 17544 & 862 & 1 hour \\ 
        Weather & 52696 & 21 & 10 min \\
        \hline 
    \end{tabular}
    \caption{
        Descriptive statistics on considered real-world datasets. 
        All datasets have $N$ timesteps, an observable vector of $D$ dimensions, and a temporal resolution of $\Delta t$.
    }
    \label{tab:real_data_desc}
\end{table}

We split each dataset into a train/validation/test partition chronologically at a ratio of 6:2:2 for Electricity and ETT and 7:1:2 for Traffic and Weather. 
Because real-world datasets do not have a characteristic Lyapunov time, we lack an exact notion of VPT. 
For this reason, we scale predictions such that 1 VPT = 1 day for all datasets.
\Cref{app:hyper} describes all hyperparameters considered. 

\subsection{Results on Forecasting Real-World Datasets}

We summarize quantitative results for all datasets in~\Cref{tab:real_dat,tab:real_dat_ett}, showing the average test VPT (averaging over initial conditions in the test split).
While the results lack a clear indication of the supremacy of one architecture over all others for all datasets considered, we do identify several actionable trends. 

\begin{table}[htb!]
    \centering
    \begin{tabular}{|l|l|l|l|}
    \hline
    \diagbox{Model}{Scenario} & Electricity & Traffic & Weather \\
    \hline
    \hline
    LSTM & 0.13 & 0.21 & 0.14 \\
    AT-LSTM & 0.19 (39\%) & 0.21 (-2\%) & 0.13 (-7\%) \\
    GV-LSTM & 0.13 (-) & 0.20 (-3\%) & 0.14 (6\%)\\
    (AT+GV)-LSTM & 0.13 (-) & 0.20 (-3\%) & 0.13 (-7\%)\\
    \hline 
    GRU & 0.14 & 0.19 & 0.14 \\
    AT-GRU & 0.23 (64\%) & 0.18 (-3\%) & 0.12 (-10\%) \\
    GV-GRU & 0.14 (-) & \textbf{0.23} (18\%) & 0.14 (-) \\
    (AT+GV)-GRU & 0.14 (-)& \textbf{0.23} (18\%) & \textbf{0.21} (56\%) \\
    \hline
    RHN & 0.14 & 0.20 & 0.14 \\
    AT-RHN & 0.16 (13\%) & 0.22 (12\%) & 0.17 (20\%) \\
    GV-RHN & 0.14 (-) & 0.18 (-8\%) & 0.12 (-17\%) \\
    (AT+GV)-RHN & 0.14 (-) & 0.18 (-8\%) & 0.18 (23\%) \\
    \hline
    Transformer(Post) & 0.19 & 0.22 & 0.15 \\
    GV-Transformer(Post) & 0.24 (26\%) & 0.22 (-) & 0.15 (-) \\
    \hline
    Transformer(Pre) & \textbf{0.27}& 0.22 & 0.13\\
    GV-Transformer(Pre) & \textbf{0.27} (-) & 0.22 (-) & 0.17 (33\%) \\
    \hline
    \end{tabular}
    \caption{
        Average Valid Prediction Time (VPT) of top models, stratified by model class and mechanism. 
        All metrics are computed from 100 initial conditions sampled from the testing data. 
        VPT is scaled such that 1 VPT = 1 day for all datasets. 
    }
    \label{tab:real_dat}
\end{table}

\begin{table*}[htb!]
    \centering
    \begin{tabular}{|l|l|l|l|l|}
    \hline
    \diagbox{Model}{Scenario}&  ETTh1 &  ETTh2 & ETTm1 & ETTm2 \\
    \hline
    \hline
    LSTM & 0.03 & 0.22 & 0.00 & 0.34 \\
    AT-LSTM & 0.03 (-19\%) & 0.34 (54\%) & 0.02 (407\%) & 0.48 (43\%) \\
    GV-LSTM & 0.03 (0\%) & 0.18 (-17\%) & 0.00 (0\%) & 0.40 (20\%) \\
    (AT+GV)-LSTM & 0.03 (-19\%) & 0.34 (54\%) & 0.02 (407\%) & 0.48 (43\%) \\
    \hline
    GRU & 0.04 & 0.21 & 0.01 & 0.43 \\
    AT-GRU & 0.04 (-1\%) & 0.22 (5\%) & 0.03 (139\%) & 0.67 (57\%) \\
    GV-GRU & 0.04 (0\%) & 0.21 (0\%) & 0.01 (0\%) & 0.43 (0\%) \\
    (AT+GV)-GRU & 0.04 (-1\%) & 0.22 (5\%) & 0.03 (139\%) & 0.67 (57\%) \\
    \hline
    RHN & 0.04 & 0.24 & \textbf{0.09} & 0.41 \\
    AT-RHN & \textbf{0.08} (107\%) & 0.20 (-14\%) & 0.03 (-65\%) &  0.26 (-37\%) \\
    GV-RHN & 0.03 (-6\%) & 0.24 (0\%) & 0.03 (-72\%) & 0.41 (0\%) \\
    (AT+GV)-RHN & \textbf{0.08} (107\%) & 0.20 (-14\%) & 0.04 (-51\%) & 0.47 (15\%) \\
    \hline
    Transformer(Post) & 0.04 & 0.41 & 0.04 & 0.52 \\
    GV-Transformer(Post) & 0.05 (44\%) & 0.45 (10\%) & 0.03 (-18\%) & \textbf{0.68} (31\%) \\
    \hline
    Transformer(Pre) & 0.06 & 0.34 & 0.02 &  0.60  \\
    GV-Transformer(Pre) & \textbf{0.08} (39\%) & \textbf{0.50} (49\%) & 0.03 (45\%)  & 0.28 (-52\%)\\
    \hline
    \end{tabular}
    \caption{
        Average Valid Prediction Time (VPT) of top-performing models, stratified by model class and mechannism. 
        All metrics are computed from 100 initial conditions sampled from the testing data. VPT is scaled such that 1 VPT = 1 day for all datasets. 
        The best in each column is highlighted in bold. 
    }
    \label{tab:real_dat_ett}
\end{table*}

\Cref{tab:real_dat,tab:real_dat_ett} suggest that the PreLN Transformer exhibits robust performance across all datasets. 
We find that the PreLN Transformer outperforms all other models in three out of seven datasets. 
While it is not the best architecture in every column, it typically follows as the second or third best. 
This is particularly compelling on noisy datasets like ETTm1 and ETTm2.
For the latter, the PostLN Transformer attains the highest average VPT, giving a counter-example where the PostLN variant outperforms the PreLN. 
For the remaining datasets where neither Transformer has the best average VPT, the GRU or RHN is the top architecture. 
The GRU works especially well on the Electricity and Traffic datasets, while the RHN excels on the ETT datasets. 
Surprisingly, the LSTM produces lower-quality forecasts for many of these real-world datasets.

We find that Transformers tend to benefit from residual gating for real-world datasets. 
In \Cref{tab:real_dat,tab:real_dat_ett} we observe that for six of seven datasets , neural gating improves or maintains the average VPT of the PreLN Transformer. 
Similarly, neural gating also improves or retains the average VPT of the PostLN Transformer on six of seven datasets. 
Except for the noisiest datasets (ETTm1, ETTm2), residual gating improves the predictive performance of Transformers in the best case and has a neutral effect at worst. 
Since the early-stopping, validation-based selection criteria work well, the cost of residual gating is primarily in training and evaluating different models. 
However, as noted in the previous sections, the cost of training a Transformer model with residual gating is much less than the training time cost of any recurrent model. 
Therefore, searching through additional models may be worth the increase in performance that is possible. 
Another important consideration is the structure of conditional information within neural gates. 
Here, we used a naive concatenation of the latent vectors of the primary and residual branches of the Transformer block. 
In future works, we hope to explore alternate conditioning structures that are simpler, regularized, or otherwise more suited to the forecasting problem.

The results from modifying the neural mechanisms of RNNs offer less clear trends than those observed for Transformers. 
For example, gating as a hyperparameter is helpful or neutral for most cells and datasets. 
The GRU is neutral to the gate choices, whereas gating reduces the RHN's performance without further considering attention. 
Attention is broadly helpful in recurrent cells. 
For all types of recurrent cells, neural attention improves performance on half of the datasets and reduces performance on the other half. 
When combining gating choice and neural attention, the GRU is improved or held constant on seven of eight datasets. 
Similarly, the combination of mechanisms maintains or increases the performance of the RHN in five of seven datasets. 
The LSTM is the only recurrent cell where mechanistic alteration reduces performance more often than improving it for these real-world test cases. 

The fact that the PreLN Transformer architecture is broadly successful across a diverse set of forecasting tasks, whether modified or unmodified, is promising. 
As Transformers are generally faster to train, this indicates the PreLN Transformer is a robust selection for forecasting tasks. 
Furthermore, Transformers increased training speed allows for more mechanistic exploration through accelerated experimentation cycles. 
When possible, it seems advantageous to compare Transformer performance against advanced, recurrent cells like the RHN; however, in resource-constrained settings where this may not be feasible, the Transformer affords a compelling choice. 
Although much prior focus on Transformers for forecasting has centered on sequence-to-sequence encoder-decoder models, these results are a positive demonstration that decoder-only, generative forecasting models are a strong option for chaotic and real-world forecasting.


\section{Conclusion}
\label{sec:conclusion}

In this work, we studied two of the most prominent and successful neural network architectures, i.e., RNNs and Transformers, and identified their key neural mechanisms, namely gating, attention, and recurrence.
We investigated three different RNN cells, GRUs, LSTMs, and RHNs.
By combining the neural mechanisms, we have designed various novel architectures and benchmarked them on forecasting multiscale chaotic dynamics in the Lorenz 96 system, the K-S equation, and various real-world time-series datasets.

Regarding the forecasting task of the multiscale Lorenz 96 dynamics, results indicate that RNNs outperform Transformers in both forcing regimes considered in this work.
On the K-S system, GRUs and RHNs outperform transformers.
More importantly, we find that neural gating and neural attention increase the performance of RNNs, while recurrence is detrimental to Transformer performance.
This is an important observation, as it directly implies that multiple other works demonstrating state-of-the-art results on these and similar physical dynamical systems, like~\cite{vlachas2020backpropagation,chattopadhyay2020data}, are missing a significant performance boost by not optimizing the neural gating and attention mechanisms.
The RHN cell with attention and neural gating exhibits the highest accuracy on both the KS system and multiscale Lorenz 96.
To our knowledge, this is the very first work to introduce a novel variant of an RNN, namely RHN with attention and neural gating, that outperforms all other variants in spatiotemporal chaotic dynamics.
The increased accuracy of RNNs comes at the cost of the increased training time compared to transformers.
The study also assesses the stability of the model's predictions and their ability to reproduce long-term statistics, finding that the top model of each class produces non-divergent predictions and faithfully captures long-term aspects of the system.

Moreover, we compared the performance of the neural architectures for time series forecasting on eight real-world datasets. 
The PreLN Transformer exhibited robust performance across all datasets, outperforming all other models in four out of eight datasets. 
The study found that Transformers tend to benefit from residual gating for real-world datasets, and the PreLN Transformer architecture is a robust selection for forecasting tasks. 
The study suggests that decoder-only, generative forecasting models like Transformers are a strong option for chaotic and real-world forecasting.

In summary, this work underscores the necessity of pinpointing the critical architectural elements that contribute to a neural model's efficacy.
By treating these elements as hyperparameters, and by crafting hybrid architectures, we can conduct ablation studies to assess their individual impact on specific tasks, and identify a best-performing architecture.
For spatiotemporal forecasting, Recurrent Highway Networks enhanced with neural gating and attention mechanisms stand out as the top-performing architecture, despite being largely overlooked in existing research.


\bibliographystyle{elsarticle-num} 
\bibliography{citations}

\appendix

\section{Hyperparameter Search Space}
\label{app:hyper}

\subsection{Multiscale Lorenz-96}

The hyperparameters considered for the Multiscale Lorenz-96 system are shown in~\Cref{tab:ml96_transformer_params,tab:ml96_rnn_params} for Transformers and RNNs, respectively.
As the lower forcing $F=10$ conducts a large sweep which informs a smaller space for $F=20$, we underline which subset of the search space is used in the high forcing regime.
Additionally, all models trained in this study share the following hyperparameters: An initial learning rate of $\alpha = 10^{-2}$, a batch size of $B=64$, no weight decay, and a learning rate schedule with $R=5$ learning rounds with patience of $P=10$ epochs and a learning rate decay of $\gamma = 0.1$.
For all models, we also consider 3 random initializations for seeds $\{42, 117, 12345\}$.

\begin{table}[htb!]
    \centering
    \begin{tabular}{|l|l|l|}
        \hline 
        \textbf{Parameter} & \textbf{Description} & \textbf{Values} \\ 
        \hline 
        \hline 
        $g$ & Activation & relu \\ 
        $L$ & Layers  & 1, \underline{2}, \underline{4} \\ 
        $d_h$ & Latent dimension & \underline{64}, \underline{256}, 512 \\ 
        $H$ & Attention heads & 1, \underline{4} \\ 
        $p_e$ & Input dropout & 0 \\ 
        $p_{f}$ & MLP dropout & 0.1 \\ 
        $p_{a}$ & Attention dropout & 0.1 \\ 
        $\mathrm{LN}$& LayerNorm order & \underline{pre}, \underline{post} \\ 
        $S$ & Sequence length & 8, \underline{16} \\ 
        $S'$ & Prediction length & 1, \underline{8} \\ 
        & Gate type & \underline{A}, L, C, \underline{D} \\ 
        $N$ & Recurrent States & \underline{0}, 4, 8 \\ 
        \hline 
    \end{tabular}
    \caption{
        Hyperparameter search space for Transformer models on the Multiscale Lorenz-96 system. For rows with multiple values, the parameters considered for the reduced search space for $F=20$ are highlighted by underline.
    }
    \label{tab:ml96_transformer_params}
\end{table} 

\begin{table}[htb!]
    \centering
    \begin{tabular}{|l|l|l|}
        \hline 
        \textbf{Parameter} & \textbf{Description} & \textbf{Values} \\ 
        \hline 
        \hline 
        $g$ & Activation & tanh \\ 
        & Cell Type & LSTM, GRU, RHN \\
        $L$ & Layers & \underline{1}, \underline{2}, \underline{4} \\ 
        $d_h$ & Latent dimension & \underline{64}, \underline{256}, 512 \\ 
        $d_L^\dagger$ & Transition depth & \underline{1}, \underline{2}, \underline{4} \\
        $S$ & Sequence length & \underline{8}, \underline{16} \\ 
        $S'$ & Prediction length & \underline{1}, 8 \\ 
        & Gate type & A, L, \underline{C}, \underline{D} \\ 
        $H$ & Attention heads & 1, \underline{4} \\ 
        $p_{a}$ & Attention dropout & 0.1 \\ 
        & Relative pos. bias & \underline{None}, I, D \\
        \hline 
    \end{tabular}
    \caption{
        Hyperparameter search space for RNN models on the Multiscale Lorenz-96 system. For rows with multiple values, the parameters considered for the reduced search space for $F=20$ are highlighted by underline. Parameters marked with $^\dagger$ are only for the RHN. When augmenting RNNs with attention, we use self-attention and cross-attention to the previous layer and input for $F=10$ and we use self-attention and cross-attention to the input for $F=20$.
    }
    \label{tab:ml96_rnn_params}
\end{table} 

\subsection{Kuramoto-Sivashinsky}

The hyperparameters considered for the K-S system are shown in~\Cref{tab:ks_transformer_params,tab:ks_rnn_params} for Transformers and RNNs, respectively.
All models trained in this study share the following hyperparameters: An initial learning rate of $\alpha = 10^{-2}$, a batch size of $B=64$, no weight decay, and a learning rate schedule with $R=5$ learning rounds with patience of $P=10$ epochs and a learning rate decay of $\gamma = 0.1$.
For all models, we also consider 3 random initializations for seeds $\{42, 117, 12345\}$.

\begin{table}[htb!]
    \centering
    \begin{tabular}{|l|l|l|}
        \hline 
        \textbf{Parameter} & \textbf{Description} & \textbf{Values} \\ 
        \hline 
        \hline 
        $g$ & Activation & relu \\ 
        $L$ & Layers  & 2, 4 \\ 
        $d_h$ & Latent dimension & 128, 256, 512 \\ 
        $H$ & Attention heads & 4 \\ 
        $p_e$ & Input dropout & 0 \\ 
        $p_{f}$ & MLP dropout & 0.1 \\ 
        $p_{a}$ & Attention dropout & 0.1 \\ 
        $\mathrm{LN}$& LayerNorm order & pre, post \\ 
        $S$ & Sequence length & 16 \\ 
        $S'$ & Prediction length & 8 \\ 
        & Gate type & A, D \\ 
        $N$ & Recurrent States & 0\\ 
        \hline 
    \end{tabular}
    \caption{
        Hyperparameter search space for Transformer models on the K-S system.
    }
    \label{tab:ks_transformer_params}
\end{table} 

\begin{table}[htb!]
    \centering
    \begin{tabular}{|l|l|l|}
        \hline 
        \textbf{Parameter} & \textbf{Description} & \textbf{Values} \\ 
        \hline 
        \hline 
        $g$ & Activation & tanh \\ 
        & Cell Type & LSTM, GRU, RHN \\
        $L$ & Layers & 1, 2, 4 \\ 
        $d_h$ & Latent dimension & 128, 256, 512 \\ 
        $d_L^\dagger$ & Transition depth & 1, 2, 4 \\
        $S$ & Sequence length & 8, 16 \\ 
        $S'$ & Prediction length & 1 \\ 
        & Gate type & C, D\\ 
        $H$ & Attention heads & 4 \\ 
        $p_{a}$ & Attention dropout & 0.1 \\ 
        & Relative pos. bias & None \\
        \hline 
    \end{tabular}
    \caption{
        Hyperparameter search space for RNN models on the K-S system. Parameters marked with $^\dagger$ are only for the RHN. When augmenting RNNs with attention, we use self-attention and cross-attention to the input.
    }
    \label{tab:ks_rnn_params}
\end{table} 

\subsection{Real World Datasets}

The hyperparameters considered for the real-world datasets are shown in~\Cref{tab:real_transformer_params,tab:real_rnn_params} for Transformers and RNNs, respectively. 
All models trained in this study share the following hyperparameters: A prediction length of $S' = 1$, an initial learning rate of $\alpha = 10^{-2}$, a batch size of $B=64$, weight decay $\omega=10^{-6}$ for RNNs and $\omega = 10^{-4}$ for Transformers (per \cite{zhuang2020adabelief}), and a learning rate schedule with $R=3$ learning rounds with patience of $P=5$ epochs and a learning rate decay of $\gamma = 0.1$.
For all models, we also consider 3 random initializations for seeds $\{42, 117, 12345\}$.
Sequence lengths are searched over at a dataset level based on the temporal resolution of each dataset: 
\begin{itemize}
    \item Electricity (hours): 12, 24, 48, 96
    \item ETTh1 \& ETTh2 (hours): 4, 8, 12, 16, 20
    \item ETTm1 \& ETTm2 (hours): 1.5, 3, 6, 12
    \item Traffic (hours): 6, 12, 24, 48, 96
    \item Weather (hours): 1, 2, 3, 6, 12
\end{itemize}

\begin{table}[htb!]
    \centering
    \begin{tabular}{|l|l|l|}
        \hline 
        \textbf{Parameter} & \textbf{Description} & \textbf{Values} \\ 
        \hline 
        \hline 
        $g$ & Activation & gelu \\ 
        $L$ & Layers  & 1, 2, 4  \\ 
        $d_h$ & Latent dimension & 16/128 \\ 
        $H$ & Attention heads & 4/16 \\ 
        $p_e$ & Input dropout & 0.2 \\ 
        $p_{f}$ & MLP dropout & 0.2 \\ 
        $p_{a}$ & Attention dropout & 0.2 \\ 
        $\mathrm{LN}$& LayerNorm order & pre, post \\ 
        & Gate type & A, L, C, D \\ 
        & Relative pos. bias & I \\
        \hline 
    \end{tabular}
    \caption{
        Hyperparameter search space for Transformer models on real-world datasets. Parameters split by a `/' have small and large variations so that small datasets (ETTh1, ETTh2) may have better-sized models applied to them.
    }
    \label{tab:real_transformer_params}
\end{table} 

\begin{table}[htb!]
    \centering
    \begin{tabular}{|l|l|l|}
        \hline 
        \textbf{Parameter} & \textbf{Description} & \textbf{Values} \\ 
        \hline 
        \hline 
        $g$ & Activation & tanh \\ 
        & Cell Type & LSTM, GRU, RHN \\
        $L$ & Layers & 1, 2, 4 \\ 
        $d_h$ & Latent dimension & 16/128 \\ 
        $d_L^\dagger$ & Transition depth & 1, 2, 4 \\
        & Gate type & C, D \\ 
        $H$ & Attention heads & 4/16 \\ 
        $p_{a}$ & Attention dropout & 0.2 \\ 
        $p_{e}$ & Input dropout & 0.2 \\ 
        & Relative pos. bias & None \\
        \hline 
    \end{tabular}
    \caption{
        Hyperparameter search space for RNN models on real-world datasets. Parameters marked with $^\dagger$ are only for the RHN. Parameters split by a `/' have small and large variations so that small datasets (ETTh1, ETTh2, ILI) may have better-sized models applied to them. When augmenting RNNs with attention, we use just cross-attention to the input.
    }
    \label{tab:real_rnn_params}
\end{table} 

\end{document}